%% file: main.tex
\documentclass[10pt,twocolumn,letterpaper]{article}

\usepackage[pagenumbers]{cvpr} 

\usepackage{times}
\usepackage{epsfig}
\usepackage{graphicx}
\usepackage{amsmath}
\usepackage{amssymb}
\usepackage{pifont}

\usepackage{gensymb}
\usepackage{mathtools}
\usepackage{booktabs}
\usepackage{multirow}
\usepackage{float}
\usepackage{amsfonts}
\usepackage{comment}
\usepackage{url}
\usepackage[bottom]{footmisc}
\usepackage{wrapfig}
\usepackage[dvipsnames]{xcolor}
\usepackage{dblfloatfix}
\newcommand{\parsection}[1]{\vspace{1mm}\noindent\textbf{#1:}~}

\usepackage{xspace}
\makeatletter
\DeclareRobustCommand\onedot{\futurelet\@let@token\@onedot}
\def\@onedot{\ifx\@let@token.\else.\null\fi\xspace}
\def\eg{\emph{e.g}\onedot} 
\def\ie{\emph{i.e}\onedot}

\def\etal{\emph{et al}\onedot}

\newcommand{\reals}{\mathbb{R}}

\newcommand{\cmark}{\ding{51}}%
\newcommand{\xmark}{\ding{55}}%

\newcommand{\ours}{SPARF\xspace}

\definecolor{yellow}{rgb}{1,1, 0.6}
\definecolor{lightyellow}{rgb}{1,1, 0.8}
\definecolor{orange}{rgb}{1, 0.8, 0.6}
\definecolor{red}{rgb}{1, 0.6, 0.6}

\definecolor{darkyellow}{rgb}{0.8, 0.8, 0.5}
\definecolor{darkred}{rgb}{0.7, 0.3, 0.3}
\definecolor{darkgreen}{rgb}{0.3, 0.7, 0.3}
\definecolor{blue}{rgb}{0, 0, 1.0}
\definecolor{green}{rgb}{0, 1.0, 0}
\definecolor{pink}{rgb}{1, 0.4, 0.7}

\usepackage[pagebackref,breaklinks,colorlinks]{hyperref}

\usepackage[capitalize]{cleveref}
\crefname{section}{Sec.}{Secs.}
\Crefname{section}{Section}{Sections}
\Crefname{table}{Table}{Tables}
\crefname{table}{Tab.}{Tabs.}

\def\confName{CVPR}
\def\confYear{2023}

\begin{document}

\title{\vspace{-10mm} SPARF: Neural Radiance Fields from Sparse and Noisy Poses
}

\author{Prune Truong$^{1, 2*}$\qquad
Marie-Julie Rakotosaona$^2$ \qquad Fabian Manhardt$^2$ \qquad Federico Tombari$^{2, 3}$ \\
$^1$ETH Zurich  \qquad $^2$Google \qquad $^3$Technical University of Munich \\
\small{\texttt{prune.truong@vision.ee.ethz.ch} \qquad \texttt{\{mrakotosaona, fabianmanhardt, tombari\}@google.com}} \\ \\
\textbf{Website}: \mbox{\url{prunetruong.com/sparf.github.io/}}\\
\textbf{Code}: \mbox{\url{github.com/google-research/sparf}}
}

\maketitle
\begin{abstract}
\input{body/00_abstract}
\end{abstract}

\vspace{-2mm}
{\let\thefootnote\relax\footnote{{$^*$This work was conducted during an internship at Google.\vspace{-2mm}}}} 
\vspace{-2mm}


\input{body/01_intro}
\input{body/02_related_work}
\input{body/03_method}
\input{body/04_exp}

\input{body/05_conclusion}

\clearpage
\newpage
\appendix
\begin{center}
	\textbf{\Large Appendix}
\end{center}
\input{supplementary_material}

\clearpage
{\small
\bibliographystyle{ieee_fullname}
\bibliography{egbib}
}

\end{document}

%% file: body/00_abstract.tex
Neural Radiance Field (NeRF) has recently emerged as a powerful representation to synthesize photorealistic novel views. While showing impressive performance, it relies on the availability of  dense input views with highly accurate camera poses, thus limiting its application in real-world scenarios. In this work, we introduce Sparse Pose Adjusting Radiance Field (SPARF), to address the challenge of novel-view synthesis given only few wide-baseline input images (as low as 3) with noisy camera poses. Our approach exploits multi-view geometry constraints in order to jointly learn the NeRF and refine the camera poses. By relying on pixel matches extracted between the input views, our multi-view correspondence objective enforces the optimized scene and camera poses to converge to a global and geometrically accurate solution. Our depth consistency loss further encourages the reconstructed scene to be consistent from any viewpoint. Our approach  sets a new state of the art in the sparse-view regime on multiple challenging datasets. 

%% file: body/01_intro.tex
\section{Introduction}\label{intro}

Novel-view synthesis (NVS) has long been one of the most essential goals in computer vision. It refers to the task of rendering unseen viewpoints of a scene given a particular set of input images. NVS has recently gained tremendous popularity, in part due to the success of Neural Radiance Fields (NeRFs)~\cite{Nerf}. NeRF encodes 3D scenes with  a multi-layer perceptron (MLP) mapping 3D point locations to color and volume density and uses volume rendering to synthesize images.
It has demonstrated remarkable abilities for high-fidelity view synthesis under two conditions: dense input views and highly accurate camera poses. 
 
\begin{figure}[t]
\centering
\vspace{-5mm}
\includegraphics[width=0.42\textwidth]{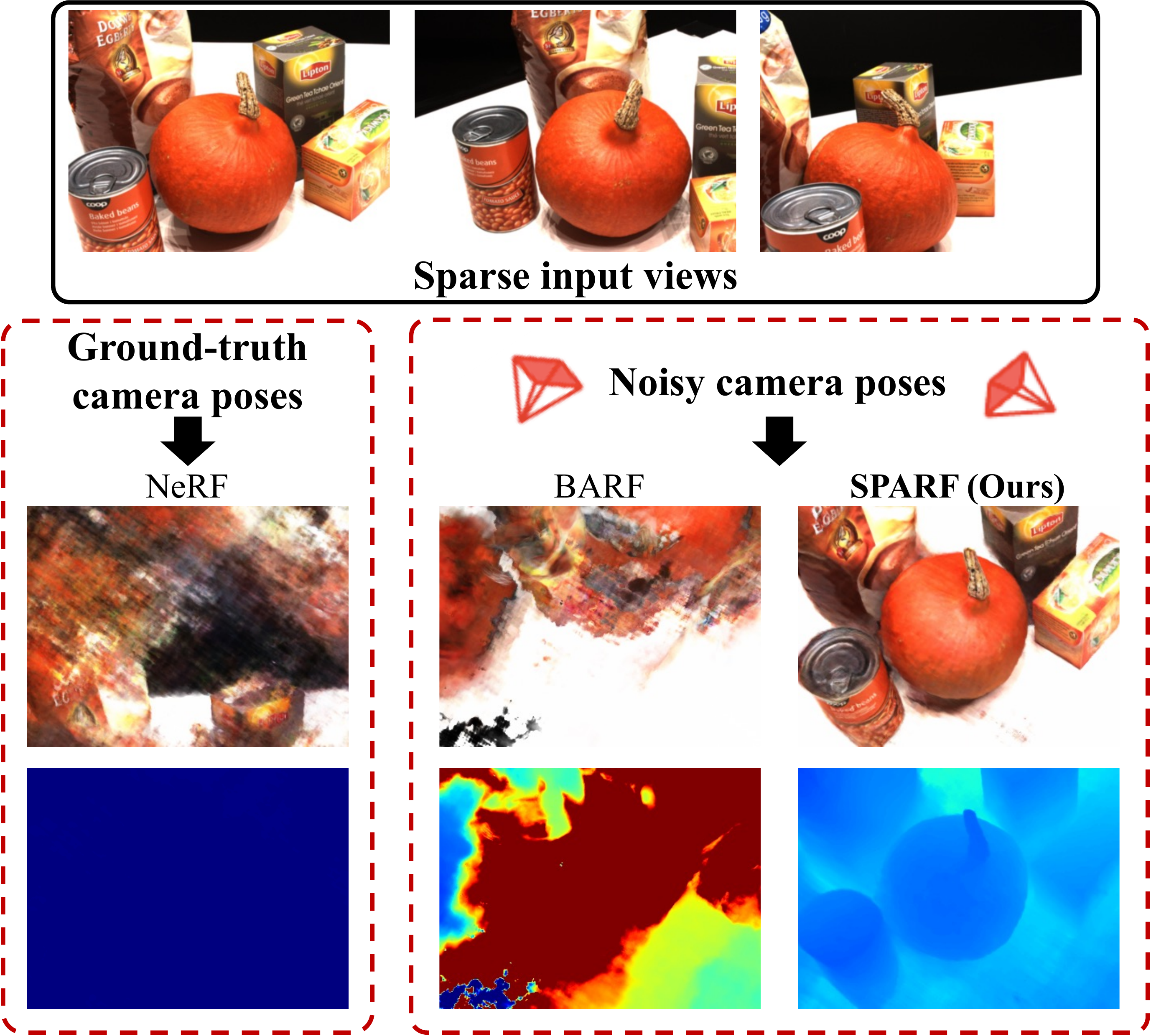}
\vspace{-3mm}
\caption{\textbf{Novel-view rendering from sparse images}. We show the RGB (second row) and depth (last row) renderings from an unseen viewpoint under sparse settings (3 input views only). Even with ground-truth camera poses,  NeRF~\cite{Nerf} overfits to the training images, leading to degenerate geometry (almost constant depth). BARF~\cite{barf}, which can successfully handle noisy poses when dense views are available, struggles in the sparse regime. Our approach \ours instead produces realistic novel-view renderings with accurate geometry, given only 3 input views with noisy poses. 
}
\label{fig:intro}
\vspace{-5mm}
\end{figure}

Both these requirements however severely impede the usability of NeRFs in real-world applications. For instance, in AR/VR or autonomous driving, the input is inevitably much sparser, with only few images of any particular object or region available per scene. In such sparse-view scenario, NeRF rapidly overfits to the input views~\cite{infonerf, Regnerf, DSNerf}, leading to inconsistent reconstructions at best, and degenerate solutions at worst (Fig.~\ref{fig:intro} left).  Moreover, the de-facto standard to estimate per-scene poses is to use an off-the-shelf Structure-from-Motion approach, such as COLMAP~\cite{colmap}. When provided with many input views, COLMAP can generally estimate accurate camera poses. Its performance nevertheless rapidly degrades when reducing the number of views, or increasing the baseline between the images~\cite{relpose}.

Multiple works focus on improving NeRF's performance in the sparse-view setting. One line of research~\cite{pixelnerf, mvsnerf} trains conditional neural field models on large-scale datasets. Alternative approaches instead propose various regularization on color and geometry for per-scene training~\cite{Regnerf, dietnerf, DSNerf, densedepth, infonerf}. Despite showing impressive results in the sparse scenario, all these approaches assume \emph{perfect camera poses} as a pre-requisite. Unfortunately, estimating accurate camera poses for few wide-baseline images is challenging~\cite{relpose} and has spawned its own research direction~\cite{MelekhovYKR17, Chen2021WideBaselineRC, ZhuangC21, EnLJ18, FredrikssonLOK16,0108DHFS19, abs-2104-01085}, hence making this assumption unrealistic.

Recently, multiple approaches attempt to reduce the dependency of NeRFs on highly accurate input camera poses. They rely on per-image training signals, such as a photometric~\cite{barf, sinerf, GARF, GNerF, nerfmm} or silhouette loss~\cite{ners, samurai, neroic}, to jointly optimize the NeRF and the poses. However, in the sparse-view scenario where the 3D space is under-constrained, we observe that it is crucial to explicitly \emph{exploit the relation between} the different training images and their underlying scene geometry, to enforce learning a \emph{global and geometrically accurate solution}. This is not the case of previous works~\cite{ners, samurai, neroic, barf, sinerf, nerfmm}, which hence fail to register the poses in the sparse regime. As shown in Fig.~\ref{fig:intro}, middle for BARF~\cite{barf}, it leads to poor novel-view synthesis quality. 

We propose Sparse Pose Adjusting Radiance Field (\ours), a joint pose-NeRF training strategy. Our approach produces realistic novel-view renderings given only \emph{few wide-baseline input images} (as low as 3) with \emph{noisy camera poses} (see  Fig.~\ref{fig:intro}~right). Crucially, it does not assume any prior on the scene or object shape. 
We introduce novel constraints derived from multi-view geometry~\cite{multiview} to drive and bound the NeRF-pose optimization. 
We first infer pixel correspondences relating the input views with a pre-trained matching model~\cite{pdcnet}. 
These pixel matches are utilized in our multi-view correspondence objective, which minimizes the re-projection error using the depth rendered by the NeRF and the current pose estimates. 
Through the explicit connection between the training views, the loss enforces convergence to a global and geometrically accurate pose/scene solution, consistent across all training views. 
We also propose the depth consistency loss to boost the rendering quality from novel viewpoints. By using the depth rendered from the training views to create pseudo-ground-truth depth for unseen viewing directions, it encourages the reconstructed scene to be consistent \emph{from any viewpoint}.   
We extensively evaluate and compare our approach on the challenging DTU~\cite{dtu}, LLFF~\cite{llff}, and Replica~\cite{replica} datasets, setting a new state of the art on all three benchmarks. 

%% file: body/02_related_work.tex
\section{Related Work}\label{relatedwork}

We review approaches focusing on few-shot novel view rendering as well as joint pose-NeRF refinement. 

\parsection{Sparse input novel-view rendering} 
To circumvent the requirement of dense input views, a line of works~\cite{pixelnerf, mvsnerf, Chibane2021StereoRF, LiuPLWWTZW22, ibrnet, grf} incorporates prior knowledge by pre-training conditional models of radiance fields on large posed multi-view datasets. Despite showing promising results on sparse input images, their generalization to out-of-distribution novel views remains a challenge. Multiple works~\cite{dietnerf, infonerf, Regnerf, DSNerf, densedepth} follow a different direction, focusing on per-scene training for few-shot novel view rendering. DietNeRF~\cite{dietnerf} compares CLIP~\cite{clip} embeddings of rendered and training views. 
InfoNeRF~\cite{infonerf} penalizes the NeRF overfitting to limited input views with a ray entropy regularization.  Similarly, Barron~\etal~\cite{mipnerf360} introduce a distortion loss, which encourages sparsity of the density in each ray. In RegNeRF, Niemeyer~\etal~\cite{Regnerf} propose to regularize the geometry and appearance of rendered patches with a depth smoothness and normalizing flow objectives. Recently, a number of works~\cite{DSNerf, nerfingmvs, densedepth,  MonoSDF} incorporate depth priors to constraint the NeRF optimization. Notably, DS-NeRF~\cite{DSNerf} improves reconstruction accuracy by including additional sparse depth supervision.
Related are also approaches that learn a signed distance function (SDF), aiming  for accurate 3D reconstruction in the sparse-view scenario~\cite{sparseneus, PSNerf}. However, all these works assume perfect poses as a pre-requisite. We instead propose a novel training strategy leading to accurate geometry and novel-view renderings in the sparse regime, \emph{even when facing imperfect input poses}. 

\parsection{Joint NeRF and pose refinement} Several approaches attempt to reduce NeRF's reliance on highly  accurate input camera poses~\cite{barf, nerfmm, GARF, GNerF, sinerf}. BARF~\cite{barf} and NeRF-\/-~\cite{nerfmm} jointly optimize the radiance field and camera parameters of initial noisy poses, relying on the photometric loss as the only training signal. SiNeRF~\cite{sinerf} and GARF~\cite{GARF} propose different activation functions, easing the pose optimization. GNeRF~\cite{GNerF} introduces a sequential training approach including a rough initial pose network that uses GAN-style training, thereby circumventing the need for initial pose estimates. SCNeRF~\cite{SCNeRF} proposes a geometric loss minimizing the ray intersection re-projection error at previously extracted sparse correspondences to optimize over camera extrinsics and intrinsics.   
A number of works~\cite{idr, neroic, samurai} also combine the photometric objective with a silhouette or mask loss, requiring accurate foreground segmentation, and limiting their applicability to objects. Related are also implicit SLAM systems~\cite{imap, niceslam, neuralrgbd}, which progressively optimize over the geometry and camera estimates of an input RGB-D sequence. While previous works assume a dense coverage of the 3D space, Zhang~\etal propose NeRS~\cite{ners}, which tackles the task of single object reconstruction by deforming a unit sphere over time while refining poses of few input views. However, NeRS is restricted to simple objects with a known shape prior. We instead assume access to only few wide-baseline RGB images with noisy pose estimates, without any prior on the scene or object shape.

%% file: body/03_method.tex
\section{Preliminaries}

We first briefly introduce notation, the basics of NeRF representation, and camera operations. 

\parsection{Camera pose} Let $P_{i}^{c2w} = \left[R_i^{c2w}|\mathbf{t}_i^{c2w} \right] \in SE(3)$ be the camera-to-world 
transform of camera $i$, where $R_i^{c2w} \in SO(3)$ and $\mathbf{t}_i^{c2w} \in \reals^{3}$ are the 
rotation and translation, respectively. 
We denote as $K \in \mathbb{R}^{3\times3}$ the intrinsic matrix.
For the rest of the manuscript, we drop the superscript $^{c2w}$. 
As a result, unless otherwise stated, $P = P^{c2w}$ and all 3D quantities are defined in the world coordinate system.  

\parsection{Camera projection} For any vector $\mathbf{x} \in \reals^{l}$ of dimension $l$, $\bar{\mathbf{x}}  \in \reals^{l+1}$ corresponds to its homogeneous representation, \ie $\bar{\mathbf{x}} = \left[\mathbf{x}^T, 1\right]$. We additionally define $\pi$ to be the camera projection operator, which maps a 3D point in the camera coordinate frame $\mathbf{x}^{c} \in \mathbb{R}^3$ to a pixel coordinate $\mathbf{p} \in \mathbb{R}^2$. Likewise, $\pi^{-1}$ is defined to be the backprojection operator, which maps a pixel $\mathbf{p}$ and depth $z$ to a 3D point $\mathbf{x}^{c}$. 
\begin{equation}
    \pi(\mathbf{x}^c) \cong  K \mathbf{x}^c,  \quad \quad \pi^{-1}(\mathbf{p}, z) =  z K^{-1}\bar{\mathbf{p}} \,.
\end{equation}

\vspace{-1mm}
\parsection{Scene representation} We adopt the NeRF~\cite{Nerf} framework to represent the underlying 3D scene and image formation. A neural radiance field is a continuous function that maps a 3D location $\mathbf{x} \in \mathbb{R}^3$ and a unit-norm ray viewing direction $\mathbf{d} \in \mathbb{S}^2$ to an RGB color $\mathbf{c} \in \left[ 0, 1 \right]^3$ and volume density $\sigma \in \reals^+$. It can be formulated as
\begin{equation}
\vspace{-1mm}
\left[\mathbf{c}, \sigma\right] = F_{\theta} \left(\gamma_{\textit{x}}(\mathbf{x}), \gamma_{\textit{d}}({\mathbf{d}})\right) \,.
\label{eq:mlp}
\end{equation}
Here, $F$ is an MLP with parameters  $\theta$, and $\gamma: \reals^{3} \rightarrow \reals^{3+6L}$ is a positional encoding function with $L$ frequency bases. 

\parsection{Volume rendering} Given a camera pose $P_{i}$, each pixel coordinate $\mathbf{p} \in \reals^{2}$ determines a ray in the world coordinate system, whose origin is the camera center of projection $\mathbf{o}_{i} = \mathbf{t}_i$ and whose direction is defined as $\mathbf{d}_{i, \text{p}} = R_i K_i^{-1}\bar{\mathbf{p}}$. We can express a 3D point along the viewing ray associated with $\mathbf{p}$ at depth $t$ as $\mathbf{r}_{i, \text{p}} (t) = \mathbf{o}_i + t \mathbf{d}_{i, \text{p}}$. 
To render the color $\hat{\mathbf{I}}_{i, \text{p}} \in \left[0, 1\right]^3$ at pixel $\mathbf{p}$, we sample $M$ discrete depth values $t_m$ along the ray within the near and far plane $\left[t_n, t_f \right]$, and query the radiance field $F_{\theta}$~\eqref{eq:mlp} at the underlying 3D points. The corresponding predicted color and volume density values $\left\{(\mathbf{c}_m, \sigma_m  ) \right\}_{m=1}^M$ are then composited as,
\begin{align}
\vspace{-1mm}
\hat{\mathbf{I}}_{i, \text{p}} &= \hat{I}(\mathbf{p}; \theta, P_i) = \sum_{m=1}^{M} \alpha_m\mathbf{c}_m \,, \label{eq:volume_rendering} \\
\text{where} \quad \alpha_m &= T_m\left(1 - \exp(-\sigma_m\delta_m)\right) \,, \label{eq:rendering_weight} \\
T_m &= \exp \left( -\sum_{m'=1}^{m} \sigma_{m'}\delta_{m'}\right)\,. \label{eq:transmittance}
\vspace{-1mm}
\end{align}
$T_m$ denotes the accumulated transmittance along the ray from $t_n$ to $t_m$, and $\delta_m = t_{m+1} - t_m$ is the distance between adjacent samples. 
Similarly, the approximate depth of the scene viewed from pixel $\mathbf{p}$ is obtained as,
\begin{equation}
\vspace{-1mm}
\label{eq:rendered-depth}
    \hat{z}_{i, \text{p}} = \hat{z}(\mathbf{p}; \theta, P_i) =  \sum_{m=1}^{M} \alpha_m t_m \,. 
\vspace{-1mm}
\end{equation}
Here, $\hat{I}$ and $\hat{z}$ denote the RGB and depth rendering functions. In practice, NeRF~\cite{Nerf} trains two MLPs, a coarse network $F^c_{\theta}$ and a fine network $F^f_{\theta}$, where the former is used to guide sampling along the ray for the latter, thereby enabling more accurate estimation of~\eqref{eq:volume_rendering}-\eqref{eq:rendered-depth}. 

\parsection{Photometric loss} Given a dataset of $n$ RGB images $\mathcal{I} = \left\{ I_1, I_2, ..., I_n \right\}$ of a scene associated with initial noisy poses $\hat{\mathcal{P}} = \left\{ \hat{P}_1, \hat{P}_2, ..., \hat{P}_n \right\}$, previous approaches~\cite{barf, nerfmm, GARF, sinerf} optimize the radiance field function $F_{\theta}$ along with the camera pose estimates $\hat{\mathcal{P}}$ using a photometric loss as follows, 
\begin{equation}
\label{eq:photo-pose-theta}
    \mathcal{L}_{\text{photo}}(\theta, \hat{\mathcal{P}}) = \frac{1}{n} \sum_{i=1}^n \sum_{\text{p}} \left\| I_i(\mathbf{p}) - \hat{I}(\mathbf{p}; \theta, \hat{P}_i) \right\|_2^2 \,.
\end{equation}
While this works well with dense views, it fails in the sparse regime. We propose an approach to effectively refine the poses and train the neural field for this challenging scenario.

\begin{figure*}[t]
\centering%
\includegraphics[width=0.99\textwidth]{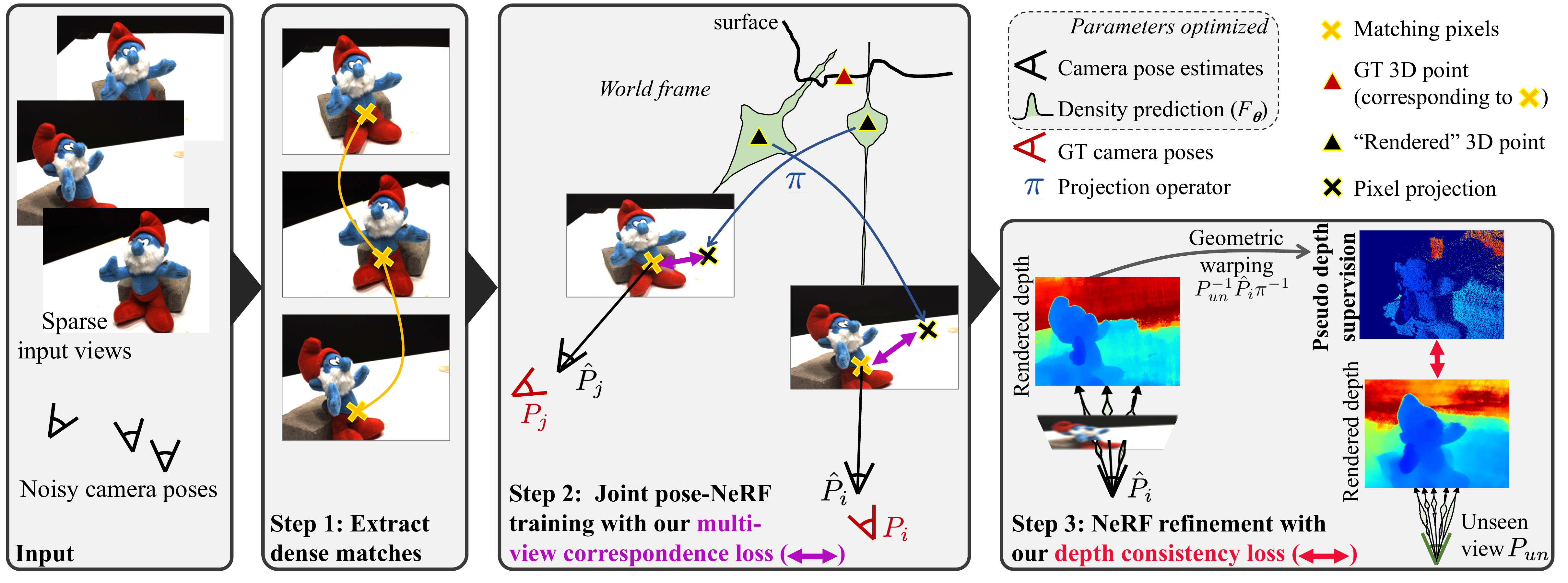}
\vspace{-3mm}
\caption{Our approach \textbf{\ours} for joint pose-NeRF training given only \emph{few input images} with \emph{noisy camera pose} estimates. We first rely on a pre-trained dense correspondence network~\cite{pdcnet} to extract matches between the training views. Our multi-view correspondence loss (Sec.~\ref{subsec:mutli-viewcons}) minimizes the re-projection error between matches, \ie it enforces each pixel of a particular training view to project to its matching pixel in another training view. We use the rendered NeRF depth~\eqref{eq:rendered-depth} and the current pose estimates $\hat{\mathcal{P}}$ to backproject each pixel in 3D space. This constraint hence encourages the learned scene and pose estimates to converge to a global and accurate geometric solution, consistent across all training views. Our depth consistency loss (Sec.~\ref{subsec:depth-cons}) further uses the rendered depths from the training viewpoints to create pseudo-depth supervision for unseen viewpoints, thereby encouraging the reconstructed scene to be consistent from any direction. }\vspace{-4mm}
\label{fig:corr-fig}
\end{figure*}

\section{Method}\label{method}

This work addresses the challenge of novel view synthesis based on neural implicit representations, in the sparse-view regime.
In particular, we assume access to only \emph{sparse input views} with \emph{noisy camera pose estimates}. The training image collection contains few images (as low as 3) and they present large viewpoint variations. 

This leads to two major challenges: 
(i) given only few input images, the NeRF model~\cite{Nerf} instantly overfits to the training views without learning a meaningful 3D geometry, even with perfect input camera poses~\cite{Regnerf, infonerf, dietnerf}. As shown in Fig.~\ref{fig:intro}, this leads to degenerate novel view renderings, including for similar train/test viewing directions. 
The problem becomes amplified when considering noisy input camera poses. 
(ii) Previous pose-NeRF refinement approaches~\cite{barf, nerfmm, sinerf, GARF, samurai} were designed considering a dense coverage of the 3D space, \ie many input views. They apply their training objectives, \eg the photometric loss~\eqref{eq:photo-pose-theta}, on each training image \emph{independently}. However, in the sparse-view regime, \ie where the 3D space is under-constrained, such supervision is often too weak for the pose/NeRF system to converge to a \emph{globally consistent} geometric solution. Failure to correctly register the training poses also leads to poor novel view rendering quality (see Fig.~\ref{fig:intro},~\ref{fig:qual}).

We propose \ours, a simple, yet effective training strategy to jointly learn the scene representation and refine the initial training poses, tailored for the sparse-view scenario. 
As the prominent source of inspiration, we draw from well-established principles of multi-view geometry~\cite{multiview}, which we adapt to the NeRF framework.
In Sec.~\ref{subsec:mutli-viewcons}, we introduce our \emph{multi-view correspondence objective} as the main driving signal for the joint pose-NeRF training. By relying on pixel correspondences between the training views, the loss enforces convergence to a global and accurate geometric solution consistent across all training views, thereby solving both challenges (i) and (ii). 
Moreover, in Sec.~\ref{subsec:depth-cons} we propose an additional term, \ie the \emph{depth consistency loss}, which encourages the learned scene geometry to be consistent across \emph{all viewpoints}, including those for which no RGB supervision is available. In doing so, it boosts novel-view rendering quality, further tackling the overfitting problem (i). 
We present our final training strategy in Sec.~\ref{subsec:prog-training} and visualize our approach in Fig.~\ref{fig:corr-fig}.

\subsection{Multi-View Correspondence Loss }
\label{subsec:mutli-viewcons}

Directly overfitting on the training images leads to a corrupted neural radiance field collapsing towards the provided views, even when assuming perfect camera poses~\cite{dietnerf, Regnerf, DSNerf}.  With noisy input poses, the problem becomes amplified, making it impossible to use the photometric loss~\eqref{eq:photo-pose-theta} as the main signal for the joint pose-NeRF training. We propose a training objective, the multi-view correspondence loss, to enforce learning a \emph{globally consistent 3D solution} over the optimized scene geometry and camera poses.

\parsection{Multi-view geometry constraint} We draw inspiration from principles of multi-view geometry~\cite{multiview}. We assume that given an image pair $\left( I_i, I_j \right)$, we can obtain pairs of matching pixels $\mathbf{p} \in I_i$ and $\mathbf{q} \in I_j$. We then compute estimates of the depth at both pixels $\hat{z}_{i, \text{p}} = \hat{z}(\mathbf{p}; \theta, \hat{P}_i)$ and $\hat{z}_{j, \text{q}} = \hat{z}(\mathbf{q}; \theta, \hat{P}_j)$ according to eq.~\eqref{eq:rendered-depth}. 
Principles of multi-view geometry dictate that both pixels must backproject to the same 3D point in the world coordinate system. This is formulated as $ \hat{P}_j \pi^{-1}(\mathbf{q}, \hat{z}_{j, \text{q}} ) = \hat{P}_i \pi^{-1}(\mathbf{p}, \hat{z}_{i, \text{p}} )$.   
However, when translating this constraint into a training objective, the magnitude of the loss is subject to large variations depending on the scene scale and the initial camera poses, requiring a tedious tuning of the loss weighting.

\parsection{Training objective} We instead project the 3D points back to image space, therefore minimizing the distance between pixels rather than directly in 3D space. We illustrate this objective in Fig.~\ref{fig:corr-fig} (steps 1-2).
For a randomly sampled training image pair $\left( I_i, I_j \right)$, our multi-view correspondence objective is formulated as $\mathcal{L}_{\text{MVCorr}}(\theta, \hat{\mathcal{P}}) = \sum_{\text{p} \in \mathcal{V}} \mathcal{L}_{\text{p}} $, where
\begin{equation}
\label{eq:mutli-cons}
\mathcal{L}_{\text{p}}  =  w_{\text{p}} \rho \left(  \, \mathbf{q} -  \pi \left(  \hat{P}_j^{-1} \hat{P}_i  \,\,\,  \pi^{-1}   (  \mathbf{p}, \hat{z}(\mathbf{p}; \theta, \hat{P}_i) \, )    \right)  \right) \,.
\end{equation}
Here $\rho$ denotes the Huber loss function~\cite{hastie01statisticallearning} and $w_{\text{p}} \in \left[0, 1\right]$ is the confidence associated with the correspondence $(\mathbf{p}, \mathbf{q})$, which we obtain as detailed below. We additionally define the set $\mathcal{V} = \left\{ \mathbf{p} : w_{\text{p}} \ge \kappa \right\}$, where $\kappa = 0.95$. 
The homogenization operations were omitted for clarity.

Our loss serves two purposes. By connecting the training images through correspondences, our multi-view correspondence objective enforces the learned geometry and camera poses to converge to a solution geometrically consistent across all training images. 
This is unlike the photometric loss~\eqref{eq:photo-pose-theta} which applies supervision on each training image independently. 
Moreover, the underlying constraint is only satisfied if the learned 3D points converge to the true reconstructed scene (up to a similarity). 
As such, the objective~\eqref{eq:mutli-cons} provides direct supervision on the rendered depth~\eqref{eq:rendered-depth}, implicitly enforcing it to be close to the surface.

\parsection{Correspondence prediction} Any classical~\cite{sift,Rublee2011} or learned~\cite{Dusmanu2019CVPR, SarlinDMR20, GLUNet, pdcnetplus, glampoints} matching approach could be used to obtain the matches relating pairs of training views. We rely on a pre-trained dense correspondence regression network, in particular PDC-Net~\cite{pdcnet}. It predicts a match $\mathbf{q}$ for each pixel $\mathbf{p}$, along with a confidence $w_{\text{p}}$. We found the high number of accurate matches to be beneficial for our joint pose-NeRF refinement. Similar conclusions were derived in the context of dense versus sparse depth supervision~\cite{DSNerf, densedepth}. The dense correspondence map also implicitly imposes a smoothness prior to the rendered depth. In suppl., we present results using a sparse matcher~\cite{superpoint, SarlinDMR20} instead.  

\subsection{Improving Geometry at Unobserved Views}
\label{subsec:depth-cons}

The multi-view correspondence loss favors a global and geometrically accurate solution, consistent across all training images. Nevertheless, the reconstructed scene often still suffers from inconsistencies when seen from novel viewpoints. For those, no RGB supervision is available during training. We propose an additional training objective, the depth consistency loss, which encourages the learned geometry to be consistent from any viewing direction. 

\parsection{Depth consistency loss} The main idea is to use the depth maps rendered from the training viewpoints to create pseudo-depth supervision for novel, unseen, viewpoints (Fig.~\ref{fig:corr-fig}, step 3). We sample a virtual pose $P_{un}$, in practice obtained as an interpolation between the poses of two close-by training  views. 
For a pixel $\mathbf{p}$ in a sampled training image $I_i$,   $\mathbf{r}^{un}_{\text{p}} = P_{un}^{-1} \hat{P}_i  \,\,\,  \pi^{-1}   (  \mathbf{p}, \hat{z}(\mathbf{p}; \theta, \hat{P}_i))$ is the corresponding 3D point in the coordinate system of the unseen view $P_{un}$. $\mathbf{y} \in \reals^2$ denotes its pixel projection in view $P_{un}$ as $\mathbf{y} = \pi \left ( \mathbf{r}^{un}_{\text{p}}  \, \right) $, and $z_{\text{y}}$ is its projected depth in $P_{un}$, \ie $z_{\text{y}}= \left [ \mathbf{r}^{un}_{\text{p}} \, \right]_3$, where $[\cdot]_3$ refers to taking the third coordinate of the vector. We formulate our depth consistency loss as,
\begin{equation}
\label{eq:depth-cons}
    \mathcal{L}_{\text{DCons}}(\theta) =  \sum_{\text{p}}  \gamma_{\text{y}} \, \rho \left(  z_{\text{y}} -  \hat{z}(\mathbf{y}; \theta, P_{un})   \right) \,.
\end{equation}
To account for occlusion and out-of-view projections in which~\eqref{eq:depth-cons} is invalid, we have included a visibility mask $\gamma_{\text{y}} \in \left[0, 1\right]$. We explain its definition in the section below. 

Since the pseudo-depth supervision $z_{\text{y}}$ is created from renderings, it is subject to errors. For this reason, we find it important to backpropagate through the pseudo-supervision $\mathbf{y}$ and $z_{\text{y}}$. Note that we however do not backpropagate through the pose estimate $\hat{P}_i$. Moreover, as verified experimentally in Tab.~\ref{tab:dtu-pixelnerf-ablation-fixed-pose}, our depth consistency objective~\eqref{eq:depth-cons} is complementary to our multi-view correspondence loss~\eqref{eq:mutli-cons}, the latter enforcing an \emph{accurate} reconstructed geometry while the former ensures it is \emph{consistent} from any viewpoint.

\parsection{Visibility mask $\gamma_{\text{y}}$} We first exclude points if their pixel projections $\mathbf{y}$ is outside of the virtual view, by setting the 
mask as $\gamma_{\text{y}} = 0$.  The depth consistency loss is also invalid for pixels that are occluded by the reconstructed scene in the virtual view. To identify these occluded pixels, we follow the strategy of~\cite{neuralwarp}. In particular, we check whether there are occupied regions on the ray between the camera center $\mathbf{o}_{un}$ of $P_{un}$ and the 3D point $\mathbf{r}_{un, \text{y}}(z_{\text{y}})$ at depth $z_{\text{y}}$. We compute how occluded a 3D point is with its transmittance~\eqref{eq:transmittance} in the unseen view, as $\gamma_{\text{y}} = T_{un, z_{\text{y}}}$. Intuitively, $\gamma_{\text{y}}$ is close to 1 if there is no point with a large density between the camera center $\mathbf{o}_{un}$ and $\mathbf{r}_{un, \text{y}}(z_{\text{y}})$, otherwise it is close to 0. Next, we present our overall training framework.

\subsection{Training Framework}
\label{subsec:prog-training}

\parsection{Staged training} Our final training objective is formulated as $\mathcal{L}(\theta, \hat{\mathcal{P}}) = \mathcal{L}_{\text{photo}}(\theta, \hat{\mathcal{P}}) + \lambda_{\text{c}} \mathcal{L}_{\text{MVCorr}}(\theta, \hat{\mathcal{P}}) + \lambda_{\text{d}} \mathcal{L}_{\text{DCons}}(\theta)$, where $\lambda_{\text{c}}$ and $\lambda_{\text{d}}$ are predefined weighting factors. The training is split into two stages. In the first part, the pose estimates are trained jointly with the coarse MLP $F^c_{\theta}$. However, due to the exploration of the pose space at the early stages of training, the learned scene tends to showcase blurry surfaces. As a result, in the second training stage, we freeze the pose estimates and train both the coarse and fine networks $F^c_{\theta}$ and $F^f_{\theta}$. This ensures that the fine network learns a sharp geometry, benefiting from the pre-trained coarse network. From a practical perspective, our training objectives can be integrated at a low computational cost, since the RGB or depth pixel renderings~\eqref{eq:volume_rendering}-\eqref{eq:rendered-depth} can be shared between the three loss terms.

\parsection{Coarse-to-fine positional encoding} In BARF~\cite{barf}, Lin~\etal propose to gradually activate the high-frequency components of the positional encodings~\eqref{eq:mlp} over the course of the optimization.  
We refer the reader to~\cite{barf} for the exact formulation. 
While originally proposed in the context of pose refinement, we found that this strategy is also extremely beneficial in the sparse-view setting, even when the poses are fixed. It prevents the network from immediately overfitting to the training images, thereby avoiding the worst degenerate geometries. We therefore adopt this coarse-to-fine positional encoding approach as default. 

%% file: body/04_exp.tex
\section{Experimental Results}\label{exp}

We evaluate the proposed \ours for novel-view rendering in the few-view setting, in particular when only three input views are available. Results with different numbers of views are provided in suppl. We extensively analyze our method and compare it to earlier approaches, setting a new state of the art on multiple datasets. 
Further results, visualizations, and implementation details are provided in suppl. 

\subsection{Experimental Settings}
\label{sec:exp-settings}

\parsection{Datasets and metrics} We report results on the \textbf{DTU}~\cite{dtu}, \textbf{LLFF}~\cite{llff} and \textbf{Replica}~\cite{replica} datasets, for the challenging scenario of 3 input views. DTU is composed of complex object-level scenes with wide-baseline views spanning a half hemisphere. We adhere to the protocol of~\cite{pixelnerf} and evaluate on their reported test split of 15 scenes. Following~\cite{Regnerf}, we additionally evaluate all methods with the object masks applied to the rendered images, to avoid penalizing methods for incorrect background predictions. 
On LLFF, we follow community standards~\cite{Nerf} and use every 8$^{\text{th}}$ image as the test set. We sample the training views evenly from the remaining images.  For the Replica dataset, which depicts videos of room-scale indoor scenes, we subsample every $k^{th}$ frame, from which we randomly select a triplet of consecutive training images.  
As metrics, we report the average rotation and translation errors for pose registration after global alignement of the optimized poses with the ground-truth ones.  For view synthesis, we report PSNR, SSIM~\cite{ssim} and LPIPS~\cite{lpips} (with AlexNet~\cite{alexnet}). On the DTU and Replica datasets, we additionally compare the rendered depth with the available ground-truth depth and compute the mean depth absolute error (DE).  More details about the pose alignment and metrics computations are provided in the supplementary, Sec.~\ref{suppl-metrics}. 

\parsection{Implementation details} 
We train our approach for 100K iterations, which takes about 10 hours on a single A100 GPU. As pose parametrization, we adopt the continuous 6-vector representation~\cite{ZhouBLYL19} for the rotation and directly optimize the translation vector. We provide all training hyperparameters in the supplementary.

\begin{table}[t]
\centering
\resizebox{0.49\textwidth}{!}{
\begin{tabular}{@{~}l@{~}l|c|c|c|c}
\toprule
& Method & PSNR $\uparrow$ & SSIM $\uparrow$  & LPIPS $\downarrow$ & DE $\downarrow $ \\ \toprule
I & Full PE~\cite{Nerf}  &  8.41 (9.34) &   0.31 (0.63) &  0.71 (0.36) &  0.87  \\
II & Smaller MLP model & 9.03 (10.06) & 0.34 (0.65) & 0.68 (0.34) & 0.79 \\
III & No PE & 16.11 (18.40) & 0.68 (0.80) & 0.37 (0.24) & \textbf{0.30} \\
IV & CF PE~\cite{barf} (Sec.~\ref{subsec:prog-training}) & \textbf{16.27} (\textbf{18.41}) & \textbf{0.69} (\textbf{0.81}) & \textbf{0.29} (\textbf{0.14}) & 0.39 \\
\bottomrule
\end{tabular}%
}\vspace{-2mm}
\caption{Comparison of different positional encoding strategies applied to NeRF~\cite{Nerf} on DTU (3 views), using ground-truth poses. Results in ($\cdot$) are computed by masking the background. }
\vspace{-4mm}
\label{tab:dtu-pixelnerf-overfitting-fixed-pose}
\end{table}

\newcommand{\green}[1]{\textcolor{Green}{\textbf{#1}}}
\newcommand{\red}[1]{\textcolor{BrickRed}{\textbf{#1}}}
\begin{table}[b]
\vspace{-3mm}
\centering
\resizebox{0.48\textwidth}{!}{
\begin{tabular}{@{~}cc@{~}|c|c|c|c}
\toprule
MV-Corr~\eqref{eq:mutli-cons} & DCons~\eqref{eq:depth-cons} & PSNR $\uparrow$ & SSIM $\uparrow$  & LPIPS $\downarrow$ & DE $\downarrow $ \\ \toprule
\red{\xmark} & \red{\xmark} & 16.27 (18.41) & 0.69 (0.81) & 0.29 (0.14) & 0.39  \\
 \red{\xmark} & \green{\cmark}  & 15.86 (18.91) & 0.71 (0.82) & 0.28 (0.14) & 0.20 \\ 
\green{\cmark} & \red{\xmark} & 18.13 (20.81) & 0.77 (\textbf{0.87}) & 0.22 (\textbf{0.10}) & 0.10 \\ 
\green{\cmark} & \green{\cmark} & \textbf{18.30} (\textbf{21.01}) & \textbf{0.78} (\textbf{0.87}) & \textbf{0.21} (\textbf{0.10}) & \textbf{0.08} \\ 
\bottomrule
\end{tabular}%
}\vspace{-2mm}
\caption{Ablation study on the DTU dataset (3 views), with fixed ground-truth poses. Results in ($\cdot$) are computed by masking the background. All networks use the coarse-to-fine PE~\cite{barf}. }
\label{tab:dtu-pixelnerf-ablation-fixed-pose}
\end{table}

\subsection{Method Analysis}

We first perform a comprehensive analysis of our approach, on DTU~\cite{dtu}, considering only 3 input views. 

\parsection{Impact of positional encoding} Training on sparse input views using the standard NeRF~\cite{Nerf} immediately overfits to the provided images, even with perfect poses. We noticed that the overfitting is largely due to the high-frequency positional encodings (PE), and thus experimented with different PE strategies. We present the results in Tab.~\ref{tab:dtu-pixelnerf-overfitting-fixed-pose}. 
The standard NeRF (I) with high-frequency PE~\cite{Nerf} leads to degenerate geometry and novel view renderings. In (II), using a simplified MLP makes little difference. While training without PE (III) largely prevents overfitting, the coarse-to-fine PE strategy~\cite{barf} leads to the best result, as shown in (IV). 

\parsection{Ablation study}  In Tab.~\ref{tab:dtu-pixelnerf-ablation-fixed-pose}, we ablate the key components of our approach, here assuming fixed ground-truth poses and starting from NeRF with coarse-to-fine PE. 
Adding our multi-view correspondence loss~\eqref{eq:mutli-cons} results in drastically better performance on all metrics. Including our depth-consistency module~\eqref{eq:depth-cons} further leads to a small improvement, achieving the best performance overall.
Also note that our depth-consistency module~\eqref{eq:depth-cons} works best in collaboration with our multi-view correspondences loss~\eqref{eq:mutli-cons} since the latter is needed to learn an accurate geometry.

\begin{figure}[t]

\centering%
\includegraphics[width=0.48\textwidth, trim={0 0 0 14}, clip]{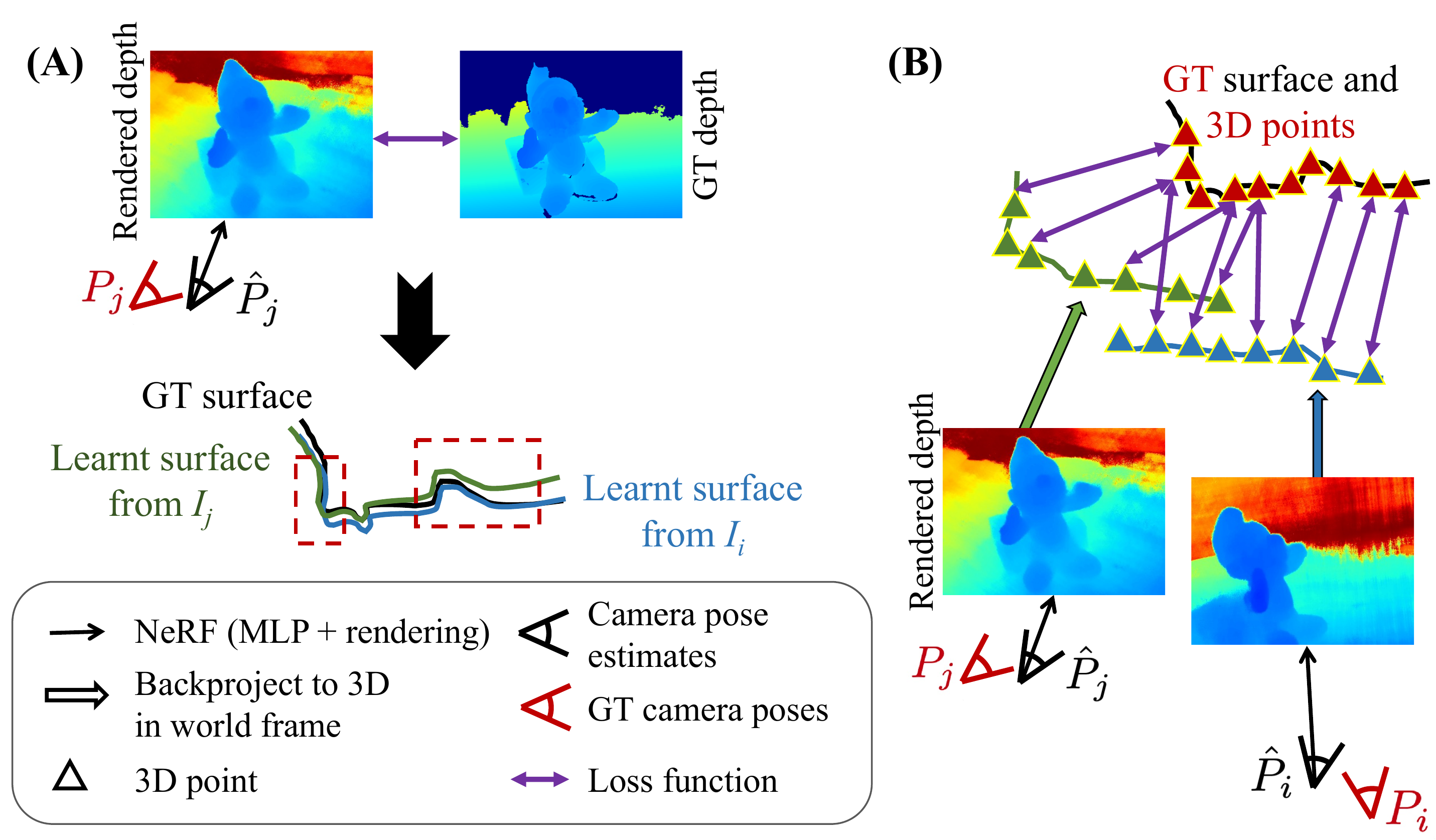}
\vspace{-8mm}
\caption{We compare two training objectives using \emph{ground-truth depth} for pose-NeRF training in the sparse-view regime. In (A), a loss comparing for each training image the rendered depth~\eqref{eq:rendered-depth} with the ground truth one, can learn locally perfect geometry (as highlighted by the dashed red rectangles). 
However, the NeRF/poses do not converge to a global solution, because the optimized poses and geometry of the different images are disjoint.
Instead, supervising the learned 3D points of each training image to be equal to the ground-truth 3D points in (B) solves this issue, by enforcing the system to converge to a global (unique) geometric solution. }
\vspace{-4mm}
\label{fig:gtdepth-loss}
\end{figure}

\parsection{Intuition on pose-NeRF training losses} We first want to build an intuition on what loss might be suitable for joint pose-NeRF training in the \emph{sparse regime}. To do so, we use ground-truth depth or 3D data in two alternative training losses, which we compare here. We illustrate this experiment in Fig.~\ref{fig:gtdepth-loss} and present results in Tab.~\ref{tab:dtu-pixelnerf-ablation-pose}, top part. As in previous work~\cite{barf}, for each scene of DTU~\cite{dtu}, we synthetically perturb the ground-truth camera poses with 15\% of additive gaussian noise.  In (I), we train with an L1 loss comparing the rendered depth~\eqref{eq:rendered-depth} with the ground-truth depth (Fig.~\ref{fig:gtdepth-loss}A). Surprisingly, this loss struggles to refine the poses. Instead, in (II) we minimize the distance between the \emph{learned 3D points} (rendered depth~\eqref{eq:rendered-depth} backprojected to world frame) and the ground-truth 3D points, as illustrated in Fig.~\ref{fig:gtdepth-loss}B. This training loss successfully registers the poses, resulting in drastically better novel-view rendering quality. As the main insight from this experiment, we hypothesize that, in the sparse-view regime, it is crucial to enforce an explicit geometric connection between the different training images and their underlying scene geometry. 
This is not the case in (I), where the depth loss favors per-image locally accurate geometry, but the NeRF/poses can converge to disconnected solutions for each training image.

\parsection{Comparison of losses for pose-NeRF training} In Tab.~\ref{tab:dtu-pixelnerf-ablation-pose} bottom part, we then compare our loss~\eqref{eq:mutli-cons} to objectives commonly used for joint pose-NeRF training. The photometric loss~\eqref{eq:photo-pose-theta} (III), even associated with a mask/silhouette loss~\cite{neroic, samurai, ners} in (IV), completely fails to register the poses, thus leading to poor novel-view synthesis performance. This is in line with our hypothesis that it is important to explicitly exploit the \emph{geometric relation} between the training views for successful registration. Moreover, because the 3D space is under-constrained in the sparse-view regime, multiple neighboring poses can lead to similar mask losses. 
While our multi-view correspondence loss~\eqref{eq:mutli-cons} alone (V) already drastically outperforms the photometric loss (III) in terms of pose and learned geometry (depth error), combining the two in (VI) leads to the best performance. This is because, through the correspondences, our approach favors a NeRF/pose solution consistent across all training images. Note that this version neither includes our depth consistency loss~\eqref{eq:depth-cons} (Sec.~\ref{subsec:depth-cons}) nor our staged training (Sec.~\ref{subsec:prog-training}).

\newcommand{\best}[1]{\textcolor{BrickRed}{\textbf{#1}}}
\newcommand{\second}[1]{\textcolor{NavyBlue}{\textit{#1}}}

\begin{table}[t]
\centering
\resizebox{0.49\textwidth}{!}{
\begin{tabular}{@{~}l@{~}l@{~}|c@{~~}c@{~}|@{~}c@{~~}c@{~~}c@{~~}c@{~~}}
\toprule
& Losses & Rot. $\downarrow $& Trans. $\downarrow $ & PSNR $\uparrow$ & SSIM $\uparrow$ & LPIPS $\downarrow $& DE$\downarrow $ \\ \toprule
I & Photo. + L1 \textbf{GT} depth & 7.3 & 28.9 & 13.8 (14.0) & 0.54 (0.70) & 0.46 (0.27) & 0.17 \\
II & Photo. + L1 \textbf{GT} 3D points & \textbf{0.4} & \textbf{1.5} & \textbf{18.8} (\textbf{20.3}) & \textbf{0.75} (\textbf{0.84}) & \textbf{0.21} (\textbf{0.11}) & \textbf{0.07} \\

\midrule
III & Photo.~\eqref{eq:photo-pose-theta} & 10.3 & 51.5 & 10.7 (9.8) & 0.43 (0.62) & 0.59 (0.36) & 1.9 \\
IV & Photo. + mask loss~\cite{neroic, ners} & 13.2 & 57.7 & - & - & - & -\\ 
V & MVCorr~\eqref{eq:mutli-cons} &  1.98 & 6.6 & - & - & - & 0.19  \\
VI & \textbf{Photo. + MVCorr} (\eqref{eq:photo-pose-theta}-\eqref{eq:mutli-cons}) & \textbf{1.85} & \textbf{5.5} & \textbf{16.0} (\textbf{17.8}) & \textbf{0.68} (\textbf{0.81}) & \textbf{0.28} (\textbf{0.14}) & \textbf{0.13} \\ 
\bottomrule
\end{tabular}%
}\vspace{-2mm}
\caption{Comparison of training objectives for joint pose-NeRF refinement on DTU~\cite{dtu} with initial noisy poses (3 views). Rotation errors are in degree and translation errors are multiplied by 100. Results in ($\cdot$) are computed by masking the background. } \vspace{-4mm}
\label{tab:dtu-pixelnerf-ablation-pose}
\end{table}

\begin{table}[b]
\centering
\vspace{-2mm}
\setlength{\tabcolsep}{4pt}
\resizebox{0.47\textwidth}{!}{%
\begin{tabular}{l|cc|cccc}
\toprule
Method & Rot. $\downarrow$ & Trans. $\downarrow $ & PSNR $\uparrow$ & SSIM $\uparrow$ & LPIPS $\downarrow $& DE $\downarrow $  \\
\toprule

BARF~\cite{barf} & 10.33 & 51.5 & 10.71 (9.76) & 0.43 (0.62) & 0.59 (0.36) & 1.90 \\ 
RegBARF~\cite{barf, Regnerf}  & 11.20 & 52.8 & 10.38 (9.20) & 0.45 (0.62) & 0.61 (0.38) & 2.33\\
DistBARF~\cite{barf, mipnerf360} & 11.69 & 55.7 & 9.50 (9.15) & 0.34 (0.76) & 0.67 (0.36) & 1.90\\
SCNeRF~\cite{SCNeRF} & 3.44 & 16.4 & 12.04 (11.71) & 0.45 (0.66) & 0.52 (0.30) & 0.85 \\
\textbf{\ours (Ours)} & \textbf{1.81} & \textbf{5.0} & \textbf{17.74} (\textbf{18.92}) & \textbf{0.71} (\textbf{0.83}) & \textbf{0.26} (\textbf{0.13}) & \textbf{0.12} \\
\bottomrule
\end{tabular}%
}\vspace{-2mm}
\caption{Evaluation on DTU~\cite{dtu} (3 views) with noisy initial poses. Rotation errors are in $\degree$ and translation errors are multiplied by 100. Results in ($\cdot$) are computed by masking the background. }
\label{tab:dtu-pixelnerf-pose}
\end{table}

\subsection{Comparison to SOTA with Noisy Poses}

Here, we evaluate \ours, our joint pose and NeRF training approach. Results with different pose initialization schemes are presented in the supplementary.

\newcommand{\pink}[1]{\textcolor{pink}{#1}}
\newcommand{\blue}[1]{\textcolor{blue}{#1}}

\begin{figure}[t]
\centering%
\includegraphics[width=0.48\textwidth, trim={0 0 0 14}, clip]{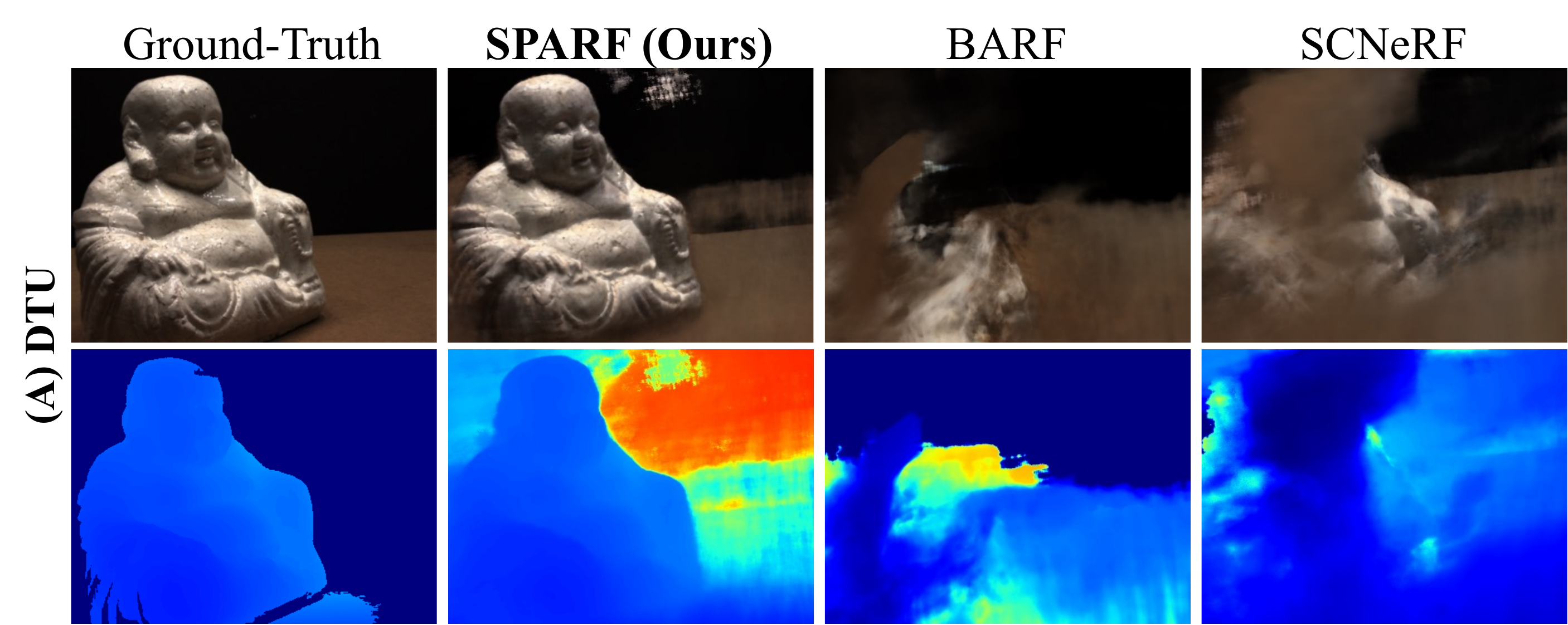} \\
\vspace{1mm}
\includegraphics[width=0.48\textwidth]{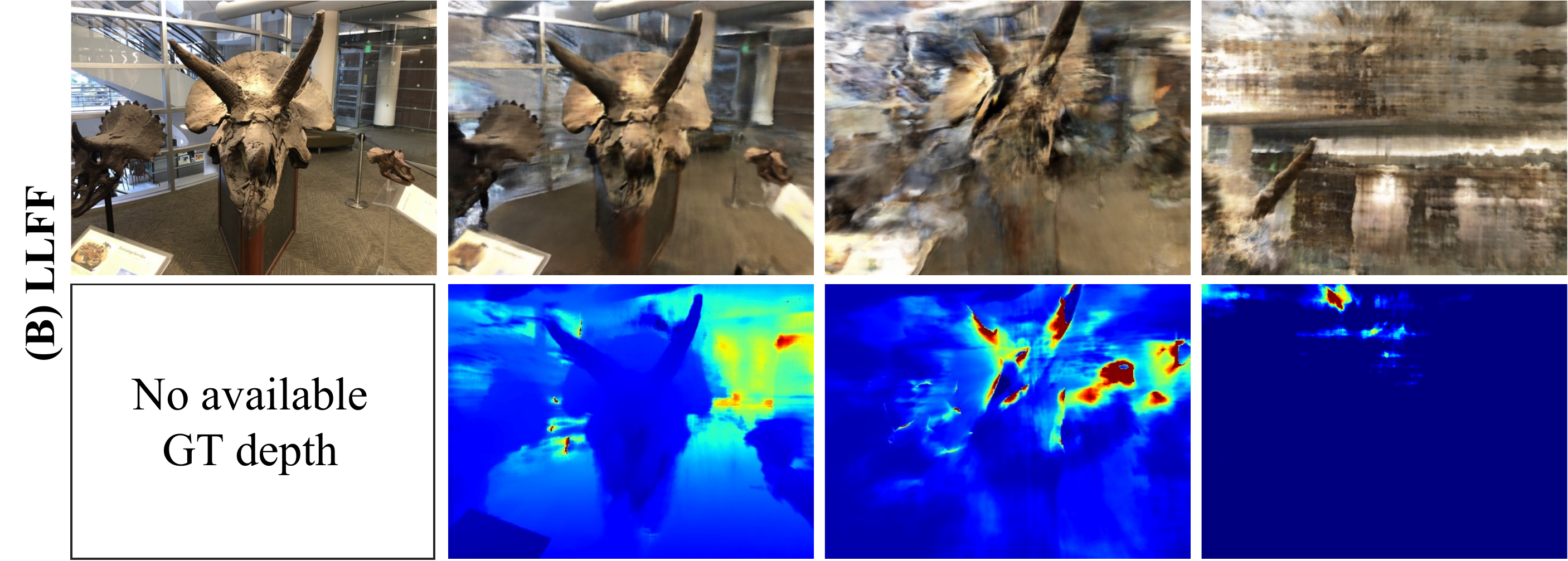} \\
\vspace{1mm}
\includegraphics[width=0.48\textwidth]{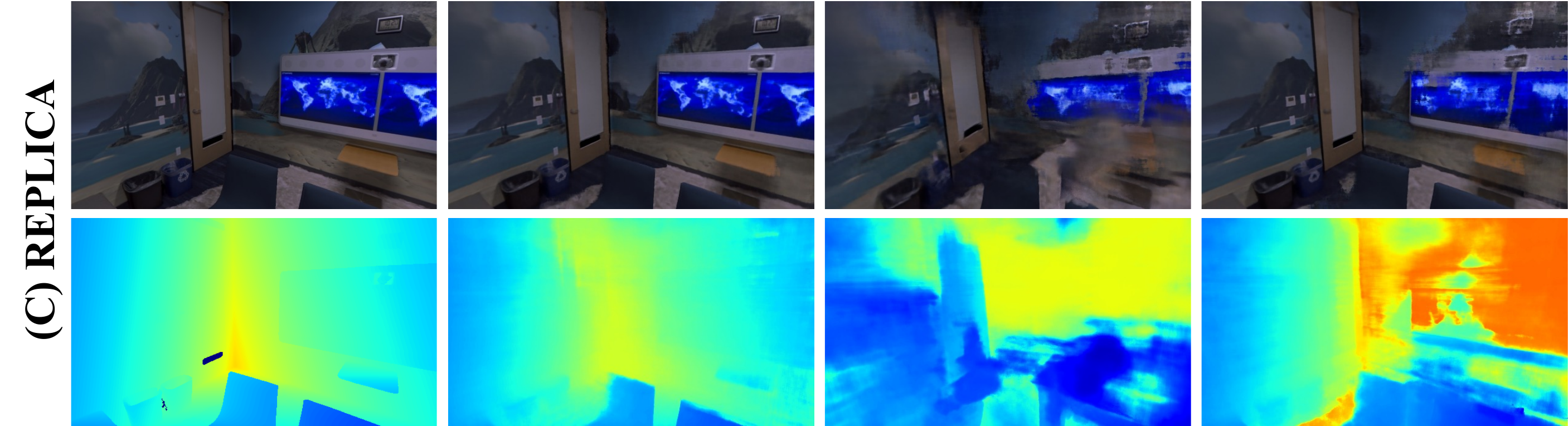}\\
\vspace{-2mm}
\caption{Novel-view rendering (RGB and depth). The input (not shown here) contains 3 images with initial noisy camera poses.
}
\vspace{-4mm}
\label{fig:qual}
\end{figure}

\begin{figure}[b]
\centering%
\vspace{-4mm}
\includegraphics[width=0.48\textwidth]{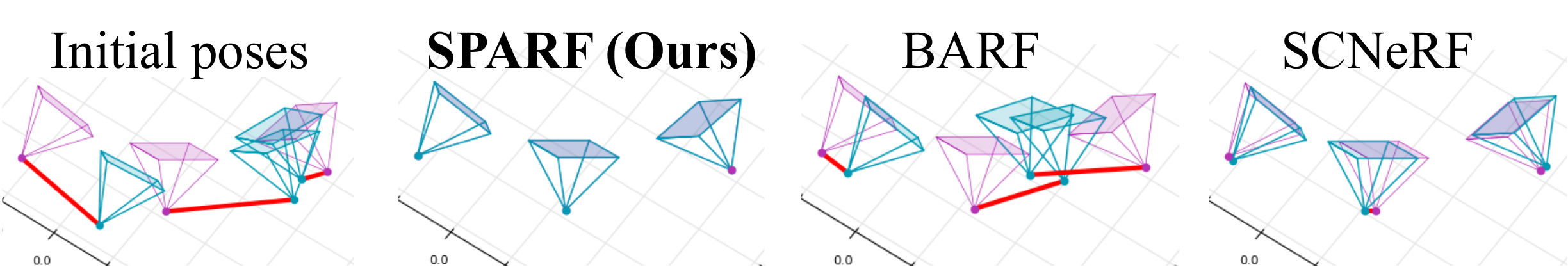} \\
\vspace{-2mm}
\caption{Optimized poses on DTU with 3 input views. We compare the ground-truth poses (in \pink{pink}) with the optimized ones (in \blue{blue}). In the first column, the initial noisy poses are in \blue{blue}. }
\label{fig:qual-pose}
\end{figure}

\parsection{Baselines} 
We compare to BARF~\cite{barf}, the state-of-the-art in pose-NeRF refinement when assuming dense input views. It is representative of a line of approaches~\cite{barf, sinerf, GARF, GNerF, nerfmm} using the photometric loss~\eqref{eq:photo-pose-theta} as the main signal. We also experiment with adding the depth regularization loss of~\cite{Regnerf} or the ray sparsity loss of~\cite{mipnerf360} to BARF, which we denote as RegBARF and DistBARF respectively. We additionally compare to SCNeRF~\cite{SCNeRF}, which uses a geometric loss based on correspondences, minimizing the rays' intersection re-projection error. For a fair comparison, we integrate coarse-to-fine PE~\cite{barf} (Sec.~\ref{subsec:prog-training}) in all methods.

\parsection{Results on DTU} Following~\cite{barf, sinerf, GARF}, for each scene, we synthetically perturb the ground-truth camera poses with 15\% of additive gaussian noise. The initial poses thus have an average rotation and translation error of 15$\degree$ and 70 respectively. We show initial and optimized poses in Fig.~\ref{fig:qual-pose}. From the results in Tab.~\ref{tab:dtu-pixelnerf-pose} and Fig.~\ref{fig:qual}A, we observe that BARF, RegBARF, and DistBARF completely fail to register the poses, leading to poor view-synthesis quality. SCNeRF's geometric loss performs better at registering the poses but the learned scene still suffers from many inconsistencies. This is because SCNeRF's loss~\cite{SCNeRF} does not influence the learned radiance field function, and thus, cannot prevent the NeRF model from overfitting to the sparse input views. Since our multi-view correspondence loss~\eqref{eq:mutli-cons} acts on \emph{both} the camera pose estimates and the learned neural field by enforcing them to fit the correspondence constraint, it leads to an accurate reconstructed scene. Our approach \ours hence significantly outperforms all others both in novel-view rendering quality and pose registration.

\parsection{Results on LLFF} The LLFF dataset consists of 8 complex forward-facing scenes. Following~\cite{barf}, we initialize all camera poses with the \emph{identity} transformation and present results in Tab.~\ref{tab:llff-pose}. In~\cite{barf}, Lin~\etal show that BARF almost perfectly registers the camera poses given \emph{dense input views}. However, we show here that it struggles in the sparse-view setting, thereby severely impacting the accuracy of novel view synthesis. While adding the depth smoothness loss (RegBARF) improves results, our approach \ours outperforms all previous works. A qualitative comparison is shown in Fig.~\ref{fig:qual}B.

\begin{table}[t]
\centering
\setlength{\tabcolsep}{4pt}
\resizebox{0.47\textwidth}{!}{%
\begin{tabular}{l|cc|ccc}
\toprule

  & Rot. ($\degree$) $\downarrow$ & Trans. ($\times$100) $\downarrow$ & PSNR $\uparrow$ & SSIM $\uparrow$ & LPIPS $\downarrow$\\
 \toprule
BARF~\cite{barf} & 2.04 & 11.6 & 17.47 & 0.48 & 0.37 \\ 
RegBARF~\cite{Regnerf, barf}  & 1.52 & 5.0 & 18.57 & 0.52 & 0.36 \\
DistBARF~\cite{barf, mipnerf360} & 5.59 & 26.5 & 14.69 & 0.34 & 0.49 \\ 
SCNeRF~\cite{SCNeRF} & 1.93 & 11.4 & 17.10 & 0.45 & 0.40 \\
\textbf{\ours (Ours)} & \textbf{0.53} & \textbf{2.8} & \textbf{19.58} & \textbf{0.61} & \textbf{0.31} \\ 
\bottomrule
\end{tabular}%
}
\vspace{-2mm}
\caption{Evaluation on the forward-facing dataset LLFF~\cite{llff} (3 views) starting from initial identity poses.}
\label{tab:llff-pose}
\end{table}

\begin{table}[t]
\centering
\setlength{\tabcolsep}{4pt}
\resizebox{0.48\textwidth}{!}{%
\begin{tabular}{@{~}@{~}l@{~}l@{~}|cc|cccc}
\toprule
& Method & Rot ($\degree$) $\downarrow$ & Trans ($\times$100) $\downarrow$ & PSNR $\uparrow$ & SSIM $\uparrow$ & LPIPS $\downarrow$  & DE $\downarrow$ \\
\toprule
G & \textbf{\ours (Ours)} & \multicolumn{2}{c|}{Fixed GT poses} & \second{26.43} & \best{0.88} & \best{0.13} & \second{0.39} \\
\midrule

F & NeRF~\cite{Nerf} &    \multicolumn{2}{c|}{Fixed poses obtained }   & 20.99 & 0.73 & 0.32 & 1.33 \\
& DS-NeRF~\cite{DSNerf} &  \multicolumn{2}{c|}{from COLMAP (run w. } & 23.52 & 0.81 & 0.20 & 0.99\\
& \textbf{\ours (Ours)} & \multicolumn{2}{c|}{PDC-Net~\cite{pdcnet} matches)} & 25.03 & \second{0.84} & \second{0.15} & 0.66 \\

\midrule
R & BARF~\cite{barf} & 3.35 & 16.96 & 20.73 & 0.72 & 0.30 & 0.84  \\
& RegBARF~\cite{Regnerf, barf}  & 3.66 & 20.87 & 20.00 & 0.70 & 0.32 & 1.00\\
& DistBARF~\cite{mipnerf360, barf} & 2.36 & 7.73 & 22.46 & 0.77 & 0.23 & 0.47 \\
& SCNeRF~\cite{SCNeRF} & \second{0.65} & \second{4.12} & 22.54 & 0.79 & 0.24 & 0.73 \\
& DS-NeRF~\cite{DSNerf} & 1.30 & 5.04 & 24.75 & 0.83 & 0.20 & 0.69 \\ 
& \textbf{\ours (Ours)} & \best{0.15} & \best{0.76} & \best{26.98}& \best{0.88} & \best{0.13} & \best{0.36} \\  
\bottomrule

\end{tabular}%
}\vspace{-2mm}
\caption{Evaluation on Replica~\cite{replica} (3 views) with initial poses obtained by COLMAP~\cite{colmap, pdcnet}. 
The initial rotation and translation errors are 0.39$\degree$ and 3.01 respectively. In the middle part (F), these initial poses are fixed and used as "pseudo-gt". In the bottom part (R), the poses are refined along with training the NeRF. For comparison, in the top part (G), we use fixed ground-truth poses. The best and second-best results are in red and blue respectively. 
}
\vspace{-4mm}\label{tab:replica-colmap-poses}
\end{table}

\parsection{Results on Replica} To demonstrate that our approach is also applicable to non-forward-facing indoor scenes, we evaluate on the Replica dataset in Tab.~\ref{tab:replica-colmap-poses} and Fig.~\ref{fig:qual}C. As pose initialization, we use COLMAP~\cite{colmap} with improved matches, \ie using PDC-Net~\cite{pdcnet}. The initial pose estimates thus have an average rotation and translation error of respectively 0.39$\degree$ and 3.01. Comparing the top (G) and middle part (F) of Tab.~\ref{tab:replica-colmap-poses}, we show that even such a low initial error impacts the novel-view rendering quality when using fixed poses. In the bottom part~(R), our pose-NeRF training strategy leads to the best results, matching the accuracy obtained by our approach with perfect poses (top row, G).

\subsection{Comparison to SOTA with Ground-Truth Poses}
\label{subsec:sota-gt-poses}

Finally, we show that our approach brings significant improvement in novel view rendering quality even when considering \emph{fixed ground-truth} poses. 

\parsection{Baselines} We compare to works specifically designed to tackle per-scene few-shot novel view rendering, namely DietNeRF~\cite{dietnerf}, DS-NeRF~\cite{DSNerf}, InfoNeRF~\cite{infonerf} and RegNeRF~\cite{Regnerf}, along with the standard NeRF~\cite{Nerf} and MipNeRF~\cite{mipnerf}. For completeness, we also compare against a state-of-the-art conditional model,  PixelNeRF~\cite{pixelnerf}, trained on DTU~\cite{dtu} and further finetuned per-scene on LLFF~\cite{llff}.

\parsection{Results} We present results on DTU and LLFF in Tab.~\ref{tab:fixedposes}. 
Compared to previous per-scene approaches~\cite{dietnerf, Regnerf, infonerf} that only apply different regularization to the learned scene, our multi-view correspondence loss~\eqref{eq:mutli-cons} provides a strong supervision on the rendered depth, implicitly encouraging it to be close to the true surface. Our depth consistency objective~\eqref{eq:depth-cons} further boosts the performance, by directly enforcing the learned scene to be consistent from any viewpoint. 
As a result, our approach \ours performs best compared to all baselines on both datasets and for all metrics. The only exception is PSNR on the whole image compared to conditional model PixelNeRF~\cite{pixelnerf}. This is because DTU has black backgrounds, where a wrong color prediction (like in Fig.~\ref{fig:qual}A for \ours) has a large impact on the PSNR value. For conditional models which rely on feature projections, it is easier to predict a correct background color. However, most real-world applications are more interested in accurately reconstructing the object of interest than the background. When evaluated only in the object region, our \ours obtains 3.24dB higher PSNR than PixelNeRF.

\begin{table}[t]
\centering
\setlength{\tabcolsep}{4pt}
\resizebox{0.48\textwidth}{!}{
\begin{tabular}{@{~}l@{~}||c@{~~}|c@{~~}|c@{~~}||@{~}c@{~}c@{~}c}
\toprule
&  \multicolumn{3}{c}{\textbf{DTU}} & \multicolumn{3}{c}{\textbf{LLFF}} \\
Method & PSNR $\uparrow
$ & SSIM $\uparrow
$  & LPIPS $\downarrow
$ & PSNR $\uparrow$ & SSIM $\uparrow$  & LPIPS $\downarrow$ \\ \toprule
 PixelNeRF~\cite{pixelnerf} & \best{19.36} (18.00) & \second{0.70} (\second{0.77}) & \second{0.32} (0.23)  & 7.93 & 0.27 & 0.68 \\
 PixelNeRF-ft~\cite{pixelnerf} & - & - & - & 16.17 & 0.44 & 0.51 \\
 \midrule  
 MipNeRF~\cite{mipnerf} & 7.64 (8.68) &  0.23 (0.57) & 0.66 (0.35)  & 14.62 & 0.35 &  0.50 \\
 NeRF~\cite{Nerf} & 8.41 (9.34) &   0.31 (0.63) &  0.71 (0.36)    & 13.61 & 0.28 & 0.56 \\
 DietNeRF~\cite{dietnerf} & 10.01 (11.85) & 0.35  (0.63) & 0.57 (0.31) & 14.94 & 0.37 & 0.5 \\
 InfoNeRF~\cite{infonerf} & 11.23 (-) & 0.44 (-) & 0.54 (-) & - & - & -  \\  
 RegNeRF~\cite{Regnerf} & 15.33 (\second{18.89}) &  0.62 (0.75) &  0.34 (\second{0.19})  & \second{19.08} &  \second{0.59} & 0.34  \\
DS-NeRF~\cite{DSNerf}& 16.52 (-) & 0.54 (-)  & 0.48 (-)  & 18.00 & 0.55 & \second{0.27}  \\
\midrule  
\textbf{\ours (Ours)} & \second{18.30} (\best{21.01}) & \best{0.78} (\best{0.87}) & \best{0.21} (\best{0.10})  & \best{20.20} & \best{0.63} & \best{0.24}\\ 
\bottomrule
\end{tabular}%
}
\vspace{-2mm}
\caption{Evaluation on DTU~\cite{dtu} and LLFF~\cite{llff} (3 views), with fixed ground-truth poses. Results in ($\cdot$) are computed by masking the background. 
Results of~\cite{mipnerf, dietnerf,Regnerf} are taken from~\cite{Regnerf}. Results of~\cite{DSNerf} for DTU are reported from their paper, while we run DS-NeRF for LLFF since the authors use a different training/test split. We compute results of~\cite{pixelnerf} using their code and pre-trained model. The best and second-best results are in red and blue respectively. }
\vspace{-4mm}
\label{tab:fixedposes}
\end{table}

%% file: body/05_conclusion.tex
\section{Conclusion}\label{conclusion}

We propose \ours, a joint pose-NeRF training strategy capable of producing realistic novel-view renderings given few wide-baseline input images with noisy camera pose estimates. By integrating two novel objectives inspired by multi-view geometry principles, we set a new state of the art on three challenging datasets.

\parsection{Limitations and future work}
Our approach is only applicable to input image collections where each image has covisible regions with at least one other.  
Moreover, the performance of our method depends on the quality of the matching network. Filtering strategies or per-scene online refinement of the correspondence network thus appear as promising future directions. 
An interesting direction is also to refine the camera intrinsics and distortion parameters along with the extrinsics. Finally, using voxel grids to encode the radiance field~\cite{mueller2022instant} instead of an MLP could lead to faster convergence, and potentially even better results.

%% file: supplementary_material.tex
\appendix

In this supplementary material, we provide additional details about our approach, experiment settings, and results.
In Sec.~\ref{sec-sup:impl}, we first give implementation details, both in terms of network architecture and training hyper-parameters. We then follow by extensively detailing the evaluation datasets and setup in Sec.~\ref{sec:data-metrics}. 

In Sec.~\ref{sec:method-analysis}, we provide additional analysis on the proposed \ours. In particular, we analyze the robustness of our joint pose-NeRF training approach to the camera pose initialization. We also present additional ablative experiments and give insights into failure cases.  Importantly, we also look at the impact of using different correspondence predictors and the influence of the quality of the predicted matches. 

In Sec.~\ref{sec-sup:results-noisy}, we present more detailed quantitative and qualitative results for our joint pose-NeRF refinement approach \ours. Notably, we start from different camera pose initialization schemes than in the main paper and train with different numbers of input views. For completeness, we also provide comparisons of our approach \ours to BARF with noisy input poses, but when \textit{all} training views are available, \ie in the many-view regime. 

Finally, we provide additional quantitative results when considering fixed ground-truth poses in Sec.~\ref{sec-sup:results-fixed}. In particular, we experiment with more input views, \ie 6 and 9 images instead of 3.

\section{Training and Implementation Details}
\label{sec-sup:impl}

In this section, we first describe the architecture of the proposed \ours. We additionally share all training details and hyper-parameters. For completeness, we also give details about the architectures and/or experimental setups used when training or evaluating baseline works.

\subsection{NeRF architecture} 

We adopt the network architecture of the original NeRF~\cite{Nerf} and its hierarchical sampling strategy with some modifications. The coarse and fine MLPs both have 128 hidden units in each layer. The numbers of sampled points of both coarse sampling and importance sampling are set to 128, and we use the softplus activation on the volume density output $\sigma$ for improved stability. 

Moreover, we found that for joint pose-NeRF optimization on LLFF, the same results are achieved with or without hierarchical sampling, \ie with a single coarse or a coarse and fine MLPs. For these experiments, we therefore only use a single MLP, since it decreases the training time. 

\parsection{Depth parametrization} On the DTU and Replica datasets, we sample the 3D points along the ray linearly in metric space, between the pre-defined near and far plane $[t_n, t_f]$. On LLFF however, we follow~\cite{barf} and sample points along each ray linearly in the inverse depth (disparity) space, where the lower and upper bounds
are $1/t_n$ = 1 and $1/t_f$ = 0.05 respectively.

\subsection{Correspondence prediction}  

\begin{figure}[t]
\centering
\includegraphics[width=0.99\columnwidth]{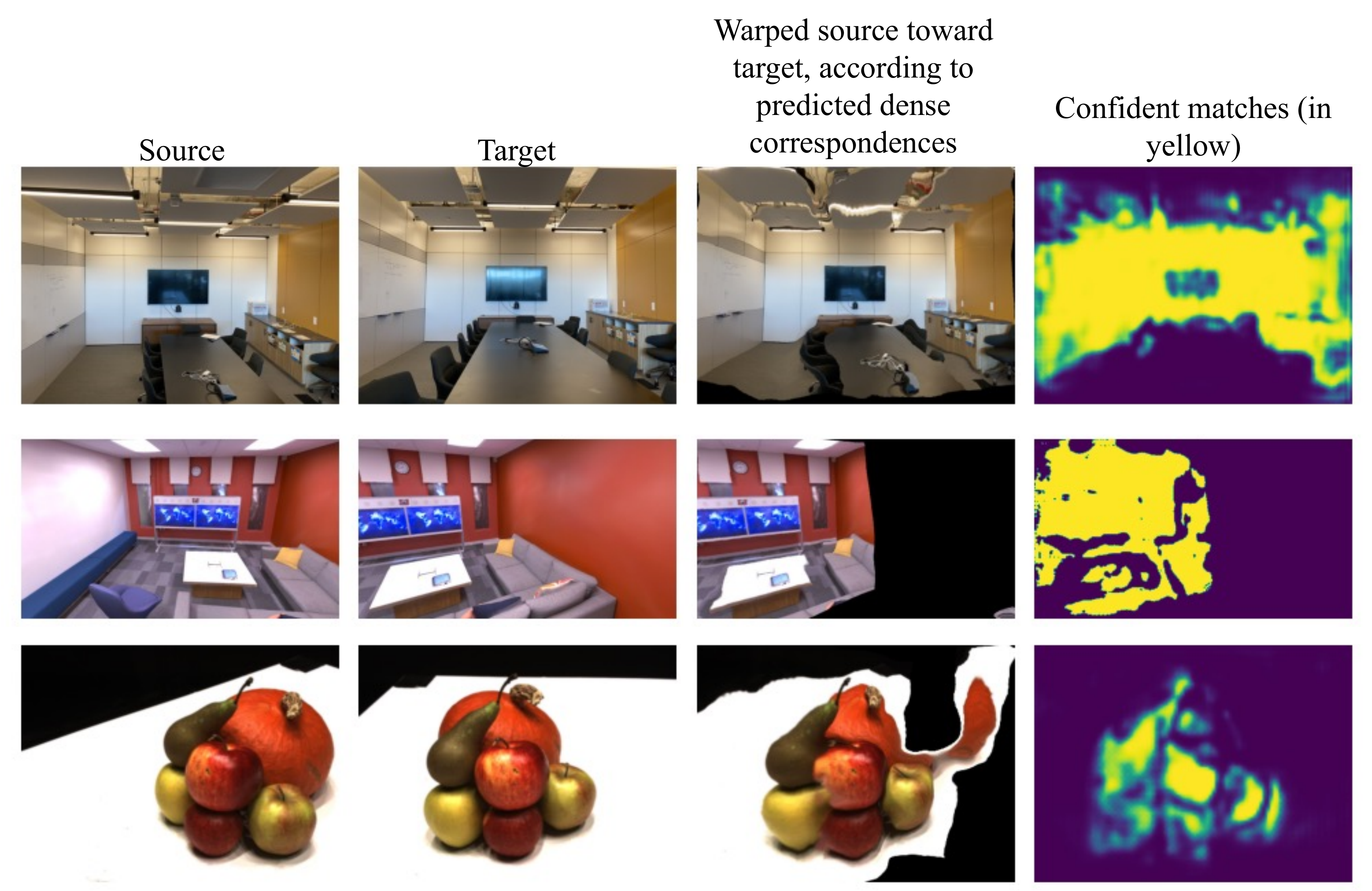}
\caption{Dense matches and associated confidence predicted by PDC-Net~\cite{pdcnet} on pair examples of the LLFF, DTU, and Replica datasets. PDC-Net predicts the dense correspondences relating the target to the source. In the 3rd column, we show the source (1$^{st}$ column) warped towards the target (2$^{nd}$ column), according to those predicted correspondences. The warped source (3$^{rd}$ column) should resemble the target (2$^{nd}$ column). The correspondences deemed reliable by PDC-Net are highlighted in yellow in the last column. 
} 
\label{fig:matches-ex}
\end{figure}

To predict the matches relating the input image pairs, we use a recent state-of-the-art dense correspondence regression network, in particular PDC-Net~\cite{pdcnet}. It predicts for each pixel the conditional probability density of the flow vector given the input image pair. In practice, this translates to predicting the mean flow vector for each pixel, which corresponds to the match, and a confidence value. As the confidence value, we use the $P_R$ operator~\cite{pdcnet}, which represents the probability that the predicted flow vector is within a certain radius of the true match. For more details, we refer the reader to the PDC-Net publication~\cite{pdcnet}. We show examples of dense matches estimated by PDC-Net in Fig.~\ref{fig:matches-ex}. 

\parsection{Matches selection} We only apply the multi-view correspondence loss (Sec.~\ref{subsec:mutli-viewcons}) on correspondences which are predicted confidently, \ie for which $P_R$ is above a certain threshold $P_R > \gamma$. In practice, we choose $\gamma = 0.95$. We optionally also further filter the correspondences by keeping only the ones that are mutually consistent, \ie for which the cyclic consistency is below 1.5 pixels.

\subsection{Training details}

Here, we describe the training hyper-parameters used in our experiments. 

\parsection{Staged training} As explained in Sec.~\ref{subsec:prog-training} of the main paper, our joint pose-NeRF training is split into two stages. In the first one, the pose estimates are jointly trained with the coarse MLP, while in the second one, the pose estimates are frozen and both coarse and fine MLPs are trained. The first training stage accounts for 30\% of the total training iterations.

We compute the matches between all-to-all views at the beginning of the training. At each iteration, the following procedure takes place. 
We sample $x$ random pixels from all the training images, on which we apply the photometric loss~\eqref{eq:photo-pose-theta}. We also sample an image pair and apply the multi-view correspondence loss (Sec.~\ref{subsec:mutli-viewcons}) on a random subset of 1024 matches. For the depth consistency loss (Sec.~\ref{subsec:mutli-viewcons}), we sample a training view $I_i$ associated with camera $\hat{P}_i$, find the closest other training view (according to current pose estimates), and compute an "unknown" camera pose $P_{un}$ as an interpolation of the two. We then randomly sample 1024 pixels in the training view $I_i$, for which we compute the depth consistency loss. 

\parsection{Coarse-to-fine positional encoding} For all datasets, we use the following scheme for the coarse-to-fine positional encoding of~\cite{barf} (Sec.~\ref{subsec:prog-training}). 
When jointly refining the poses and training the NeRF, we linearly adjust the frequency width of the positional encoding from 40\% to 70\% of the training iterations. This means that for 40\% of the training, there are no positional encodings applied to the 3D points and the ray directions. This mostly corresponds to when the camera poses are optimized. 

When the input poses are fixed, we instead adjust the positional encoding from 10\% to 50\% of the training iterations. This is because the goal of the coarse-to-fine positional encoding is in that case to prevent overfitting at the early stages of training. 

\parsection{Training schedule with 3 input views} When the poses are fixed, we train for 50K iterations on DTU and Replica, and for 70K iterations on LLFF. For the joint pose-NeRF refinement, we instead train for 100K iterations on all datasets. 

Training for longer (\ie 100K iterations) with fixed ground-truth poses leads to similar or worse results than 50 or 70K iterations since the network starts to heavily overfit to the provided few (3) training images. 

\parsection{Training schedule with 6 input views} When the poses are fixed, we train for 100K iterations on DTU and Replica, and for 140K iterations on LLFF. For the joint pose-NeRF refinement, we instead train for 150K iterations on DTU and Replica, and 170K on LLFF. 

\parsection{Training schedule with 9 input views} When the poses are fixed, we train for 150K iterations on DTU and Replica, and for 200K on LLFF. For the joint pose-NeRF refinement, we instead train for 200K iterations on DTU and Replica, and 220K on LLFF.

\parsection{Depth range} Each dataset provides a depth range $[t_n, t_f]$ within which the discrete depth values are sampled. When the initial poses are noisy, however, the provided range might not be sufficient to cover the scene. This is for example the case when we add 15\% of noise to the ground-truth poses. For the joint pose-NeRF training, we therefore use a modified depth range $[(1-\epsilon) t_n, (1+\epsilon) t_f $, where $\epsilon = 0.3$.  

\parsection{Loss weighting} Our final loss formulation is provided in Sec.~\ref{subsec:prog-training}. We set the weights $\lambda_{\text{c}}$ and $\lambda_{\text{d}}$ associated with respectively the multi-view correspondence loss (Sec.~\ref{subsec:mutli-viewcons}) and the depth consistency loss (Sec.~\ref{subsec:depth-cons}) to $\lambda_{\text{c}} = 10^{-3}$ and $\lambda_{\text{d}}=10^{-3}$.  

The intuition behind the weight  $\lambda_{\text{c}}$ is that the multi-view correspondence loss should have a magnitude in the same range as the photometric loss~\eqref{eq:photo-pose-theta} since it is the main driving force of the pose optimization at the early stages of training. 
The correspondence prediction is nevertheless prone to errors. After the poses have converged, it can lead to errors in the learned geometry. In particular, if the weight of the multi-view correspondence loss is too high, the NeRF model can learn a wrong geometry, which is consistent with the wrong correspondences, even when it violates the photometric loss. To account for that, we gradually halve the weights $\lambda_{\text{c}}$ every 10K iterations, after the poses are frozen. This enables the photometric signal to gradually gain more and more importance, thus correcting possible errors in the learned geometry. When the poses are fixed to ground truth, we also halve the weights $\lambda_{\text{c}}$ every 10K iterations, starting from the beginning of the training.

As for the weight $\lambda_{\text{d}}$, the idea is that the depth consistency loss should account for less than the multi-view correspondence loss. The reason is that the latter ensures the model learns an \emph{accurate} geometry while the former makes sure it is \emph{consistent} from any viewing directions. 

Only on DTU with fixed ground-truth poses, we find it beneficial to set $\lambda_{\text{c}} = 10^{-4}$ and $\lambda_{\text{d}}=10^{-3}$ instead. We believe that a larger weight can have a negative impact as it amplifies possible errors in the correspondence predictions, which are reverberated on the learned scene geometry.

\newcommand{\ba}{\mathbf{a}}
\newcommand{\bb}{\mathbf{b}}

\parsection{Pose parametrization}  As in BARF~\cite{barf}, we optimize the world-to-camera transformation matrices. For the camera position, we simply adopt a 3D embedding vector in Euclidean space, denoted as $\mathbf{t} \in \mathbb{R}^3$, which we can directly update throughout the optimization. However, directly learning the rotation offset for each element of a rotation matrix would break the orthogonality of the rotation matrix.

The widely-used representations such as quaternions and Euler angles are discontinuous. Following~\cite{SCNeRF}, we adopt the 6-vector representation~\cite{ZhouBLYL19}. In particular, we use and optimize a continuous embedding vector $\mathbf{r} \in \mathbb{R}^6$ to represent 3D rotations, which is more suitable for learning. Concretely, given a rotation matrix $R = [\mathbf{a_1} \,\, \mathbf{a_2} \,\,  \mathbf{a_3}] \in \mathbb{R}^{3 \times 3}$, we compute the rotation vector $\mathbf{r}$ by dropping the last column of the rotation matrix.

From the 6D pose embedding vector $\mathbf{r}$, we can then recover the original rotation matrix $R$ using a Gram-Schmidt-like process, in which the last column is computed by a generalization of the cross product to three dimension~\cite{ZhouBLYL19}. It is formulated as a function $f$, which takes as input $\mathbf{r} = [\ba_1^T, \ba_2^T]$ and enables to recover the full rotation matrix, as follows, 
\begin{equation}
R = f\left(
\begin{bmatrix}
|  \\
\mathbf{r} \\
| 
\end{bmatrix}\right) = 
\begin{bmatrix}
| & | & | \\
\bb_1 & \bb_2 & \bb_3 \\
| & | & |
\end{bmatrix},
\label{eq:6d-representation}
\end{equation}
where $\bb_1, \bb_2, \bb_3 \in \mathbb{R}^3$ are $\bb_1 = N(\ba_1)$, $\bb_2 = N(\ba_2 - (\bb_1 \cdot \ba_2) \bb_1)$, and $\bb_3 = \bb_1 \times \bb_2$, and $N(\cdot)$ denotes L2 norm. At every iteration, the estimates of the rotation and translation parameters $\hat{R}^{w2c}$ and $\mathbf{\hat{t}}^{w2c}$ are updated as, 
$$
\hat{R}^{w2c} = f(\hat{\mathbf{r}}^{w2c}_0 
+ \Delta \mathbf{r}), \;\; \mathbf{\hat{t}}^{w2c} = \mathbf{t}^{w2c}_0 + \Delta \mathbf{t} \,.
$$
Here, $\hat{\mathbf{r}}^{w2c}_0$ and $\mathbf{t}^{w2c}_0$ denote the initial (noisy) camera rotation and translation parameters. 

\parsection{Hyper-parameters used for DTU} We use the Adam optimizer to optimize the network weights and the camera poses.  For the network, we use an initial learning rate of $5\times 10^{-4}$, which is exponentially decreased to  $1\times 10^{-4}$ throughout the training. For the camera poses, we instead use an initial learning rate of $1 \times 10^{-3}$ decaying to $1\times 10^{-4}$. We resize the images to $300 \times 400$ and randomly sample 1024 pixel rays at each optimization step for the photometric loss~\eqref{eq:photo-pose-theta}. 

\parsection{Hyper-parameters used for LLFF} We use the Adam optimizer with an initial learning rate of $1\times 10^{-3}$ exponentially decreased to  $1\times 10^{-4}$ throughout the training, for the network. For the camera poses, we instead use an initial learning rate of $3 \times 10^{-3}$ decaying to $1\times 10^{-5}$. We resize the images to $378 \times 504$ and randomly sample 2048 pixel rays at each optimization step for the photometric loss~\eqref{eq:photo-pose-theta}. 

For joint pose-NeRF optimization on LLFF, we found in beneficial to only add the multi-view correspondence loss and the depth consistency loss after 1K iterations of training. This means that for the first 1K iterations, only the photometric signal~\eqref{eq:photo-pose-theta} is used. This is because for some scenes, applying the multi-view correspondence loss from the beginning can lead to the background being in front of the foreground. Applying only the photometric loss at the very beginning of the training avoids this artifact. Our additional losses can then drive the poses and the geometry correctly. Moreover, we found that for joint pose-NeRF optimization on LLFF, the same results are achieved with or without hierarchical sampling. For these experiments, we therefore only use a single MLP, since it decreases the training time.

\parsection{Hyper-parameters used for Replica} We use the same training hyper-parameters as for DTU. That is, for the network we use the Adam optimizer with an initial learning rate of $5\times 10^{-4}$ which is exponentially decreased to  $1\times 10^{-4}$ throughout the training. For the camera poses, we instead use an initial learning rate of $1 \times 10^{-3}$ decaying to $1\times 10^{-4}$. We resize the images to $360 \times 600$ and randomly sample 1024 pixel rays at each optimization step for the photometric loss~\eqref{eq:photo-pose-theta}.

\parsection{COLMAP} We run COLMAP~\cite{colmap} using the default parameters, with some exceptions. As recommended in the official documentation to better handle few images with a wide baseline, we reduce the minimum triangulation angle. We also enable the triangulation of two-view tracks,  which can in rare cases improve the stability of sparse image collections by providing additional constraints in the bundle adjustment. To increase the number of matches, we use the more discriminative DSP-SIFT features instead of plain SIFT and also estimate the affine feature shape. Finally, we enable guided feature matching. We experiment with different pixel projection thresholds for the PnP pose estimation (default is 12) but see little impact on the initial pose registration results.  

Since COLMAP often fails in the sparse-view scenario, we replace the feature matching of the standard COLMAP with better and more recent matching approaches. As reference implementation, we use the HLOC toolbox~\cite{HLOC}, which we modify to fit our needs. We try to use SuperPoint~\cite{superpoint} and SuperGlue~\cite{SarlinDMR20}, which has become the de-facto state-of-the-art sparse matching approach. As an alternative, we also use PDC-Net matches~\cite{pdcnet}, a state-of-the-art dense matching approach. Note that we also use the latter to establish the correspondences between the training images in our approach \ours. 
When using SuperPoint and SuperGlue, we set all default parameters. For the SuperGlue model, we use the indoor weights since both Replica and DTU scenes are taken indoors. We also set the default settings for PDC-Net.

\parsection{Additional runtime} The multi-view correspondence loss and depth consistency objective add a factor of around 1.5 to the optimization time, regardless of the number of views. To predict the matches, PDC-Net~\cite{pdcnet} runs at 10fps on $300 \times 400$ images.

\subsection{Baselines}

\parsection{BARF} We use the published code base for experiments using BARF. 

\parsection{SCNeRF} We use the official code base to obtain the implementation of the ray distance re-projection loss (named projected ray distance in~\cite{SCNeRF}), which we integrate into our code. In the original paper, the projected ray distance is scaled with a weight $\lambda=10^{-4}$. We kept this weighting for the LLFF experiments. However, we found that increasing this weight to $\lambda=10^{-1}$ leads to much improved results on the DTU and Replica datasets. The projected ray distance loss relies on extracted correspondences between the views. For fairness, we use PDC-Net~\cite{pdcnet} correspondences, \ie the same matches that we rely on in our multi-view correspondence loss (Sec.~\ref{subsec:mutli-viewcons}).

\parsection{PixelNeRF} For evaluation results, we run the provided pre-trained model on the official code base. 

\parsection{DS-NeRF} We use the official code base to obtain the implementation of the depth loss, which we integrate into our code base. For the results on DTU with fixed ground-truth poses, we report the results from the publication. Nevertheless, we were unable to reproduce them using the official code base, where the configuration files for DTU are not released. We suspect that the authors used a 'trick' in the NeRF architecture to prevent heavy overfitting, \eg for example reducing the positional encoding frequency. 
The results provided in the original publication for LLFF are computed using a different train/test split. We therefore re-train on our train/test splits using the released configuration files.

\section{More Details on Datasets and Metrics}
\label{sec:data-metrics}

In this section, we provide details about the evaluation datasets and metrics.

\subsection{Datasets}

We evaluate on the LLFF, DTU and Replica dataset. In general, we found it important that the scene/object 3D points are more or less centred at the origin, and rescale the initial poses such that this condition is met. 

\parsection{LLFF} As image resolution, we resize the images to $1/8^{th}$ of their original size, resulting in images of size $378 \times 504$. As stated in the main paper (Sec.~\ref{sec:exp-settings}), we follow community standards~\cite{Nerf} and use every 8$^{\text{th}}$ image as the test set. We sample the training views evenly from the remaining images.

\parsection{DTU} Following previous works~\cite{Regnerf, DSNerf}, we adhere to the evaluation protocol from PixelNerf and use the following 15 scan IDs as the test set: 8, 21, 30, 31, 34, 38, 40, 41, 45, 55, 63, 82, 103, 110, 114. The following image IDs (starting with “0”): 25, 22, 28, 40, 44, 48, 0, 8, and 13 are used as input. For the 3 and 6 input scenarios, we use the first 3/6 image IDs, respectively. For evaluation, the remaining images are used with the exception of the following image IDs due to wrong exposure: 3, 4, 5, 6, 7, 16, 17, 18, 19, 20, 21, 36, 37, 38, 39. We directly use the rescaled images provided on the PixelNeRF github, where the images have a resolution of $300 \times 400$. The poses are in the DVR's format, such that the object is centred at the origin and in the unit cube. Note that we found it important that the object is more or less centred at the origin. 
Following~\cite{Regnerf}, we additionally evaluate all methods with the object masks applied to the rendered images. The object masks are obtained from~\cite{Regnerf, idr}. This is because, in most applications, it is more important to render the object of interest with high quality, rather than the background. Applying the foreground mask to the rendered images thus avoids penalizing methods for incorrect background predictions, regardless of the quality of the rendered object of interest. 

\parsection{Replica} We use the following 7 scenes as the test set: room0, room1, room2, office0, office1, office2, and office3. Each scene features a video of an indoor room, with between 1500 to 3000 frames. To create a realistic sparse-view scenario, where only few wide-baseline images per scene are available, we sub-sample every $k^{th}$ frame, from which we randomly select a triplet of consecutive training images. Because each scene has a different frame rate, we adapt the sampling rate $k$ to each scene individually. It is chosen such as each sampled image has a minimum of $20\%$ covisible regions with another selected view. The exact sampling parameters will be included in the released code. We also adjust the poses such as the scene is centered around the origin for the selected training images. Indeed, we found it very beneficial for the pose refinement to have the scene/object 3D points centred around the origin. Without it, the range of 3D points seen by the training views can be very far off the origin. 
We use an image resolution of $340 \times 600$.

\subsection{Metrics} 
\label{suppl-metrics}

\parsection{Alignment} When refining the camera poses, we evaluate the quality of registration by globally pre-aligning the optimized poses to the ground truth ones. This is necessary because both the scene geometry and camera poses are variable up to a 3D similarity transformation. The standard procedure~\cite{nerfmm, barf} is to align the two sets of pose trajectories (optimized and ground-truth) globally with a Sim(3) transformation using Umeyama algorithm~\cite{Umeyama91} in an ATE toolbox~\cite{ATE}.
Nevertheless, we found this strategy to give very unstable and unreliable results when the trajectory contains very few views (\ie less than 9), which is the scenario we are focusing in this paper. 

As a result, we perform the alignment in a RANSAC-inspired process. We sample every possible pair of cameras in one set, and compute the Sim(3) transformation (scale/rotation/translation) relating it to the same camera pair in the other trajectory. This gives us a set of possible Sim(3) transformations relating the optimized to the ground-truth trajectories. 
We then keep as global Sim(3) transformation the one leading to the lowest average camera alignment error. This process is done for the alignment when less than nine input views are available. Otherwise, we use the standard Umeyama algorithm~\cite{Umeyama91}. 

\parsection{Pose registration} After the optimized poses are aligned with the ground-truth ones, we can compute pose registration metrics. In particular, we report the average rotation and translation errors. The rotation error $\left | R_{err}  \right |$ is computed as the absolute value of the rotation angle needed to align ground-truth rotation matrix $R$ with estimated rotation matrix $\hat{R}$, such as
\begin{equation}
    R_{err} = cos^{-1}\frac{Tr(R^{-1}\hat{R}) -1}{2} \;,
\end{equation} 
where operator $Tr$ denotes the trace of a matrix. The translation error $T_{err} $ is measured as the Euclidean distance $ \left \|\hat{T} - T \right \|$ between the estimated $\hat{T}$ and the ground-truth position $T$. Note that on all datasets, the positions of the poses are not in metric space, such that the translation error has no units. 

\parsection{Novel-view rendering} To evaluate the quality of novel view synthesis while being minimally affected by camera misalignment, we transform the test views to the coordinate system of the optimized poses by applying the scale/rotation/translation from the alignment analysis. 
To evaluate view synthesis in that case, we follow previous works~\cite{barf, nerfmm, yen2020inerf} and run an additional step of test-time photometric optimization on the trained models to factor out the pose error from the view synthesis quality. In essence, it is a more fine-grained gradient-driven camera pose alignment which minimises the photometric error on the synthesised image, while keeping the NeRF model fixed. This test-time photometric optimization is run in experiments where the poses are refined. For fairness, we also use it in experiments where we fix the initial noisy poses, \eg obtained by COLMAP~\cite{colmap}, to differentiate the novel-view rendering quality from the initial pose error. 

To evaluate the view-synthesis performance, we report the mean Peak Signal-to-Noise Ratio (PSNR), Structural Similarity Index (SSIM)~\cite{ssim}, and the Learned Perceptual Image Patch similarity (LPIPs) metric~\cite{lpips}, which estimates the distance between an image pair in a learned feature space. We use
the AlexNet network version of LPIPS following BARF~\cite{barf}.

For the depth evaluation, we first multiply the predicted depth with the scale from the alignment (since the optimized scene is variable up to a 3D similarity), such that it is in the same range than the ground-truth depth. We then compute the absolute difference between the predicted and ground-truth depths, averaged over the valid ground-truth depth areas.

\section{Additional Method Analysis}
\label{sec:method-analysis}

In this section, we present additional analyses of the proposed approach \ours. We first look at the degradation faced by COLMAP~\cite{colmap} when reducing the number of input views. We also analyze the robustness of our approach \ours to different pose initialization, and provide insights into failure cases. Additionally, we look at the impact of using different correspondence predictors and the influence of the quality of the predicted matches. Finally, we present additional ablation studies. 

\begin{figure}[t]
\centering%
\includegraphics[width=0.99\columnwidth]{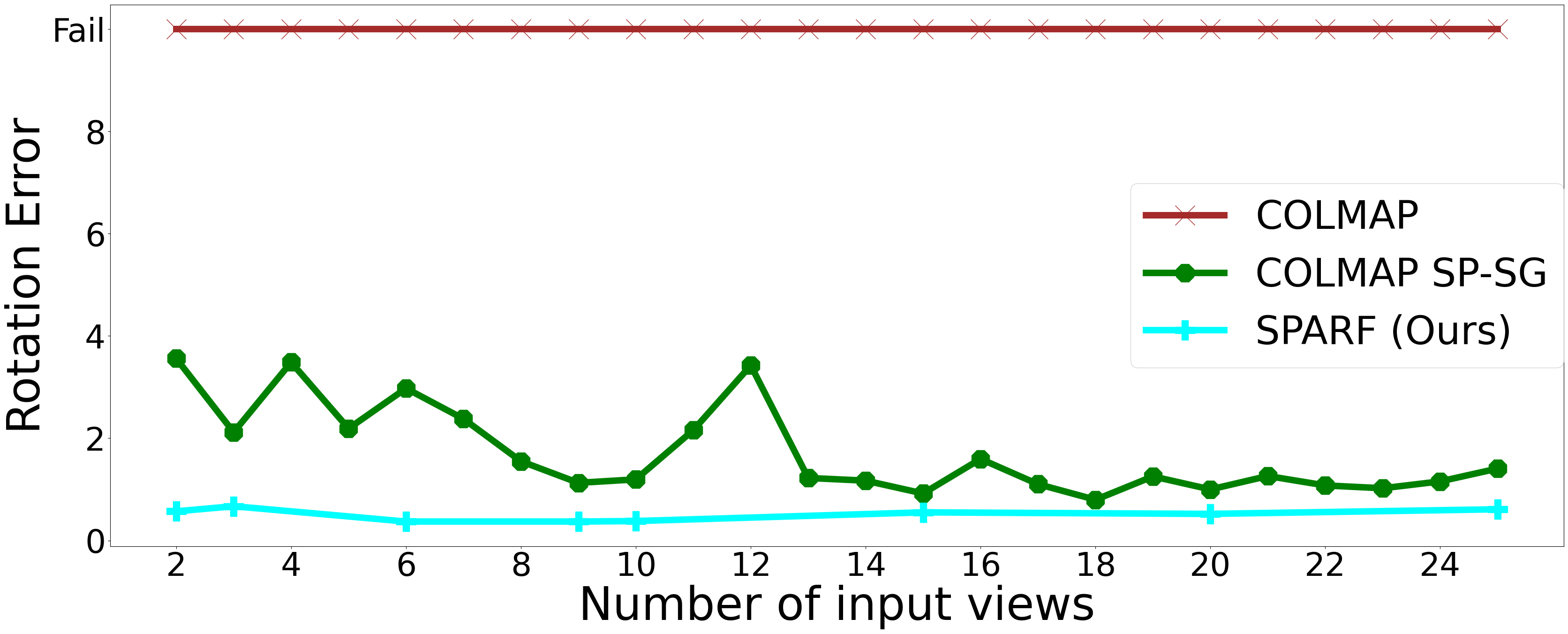} \\
\includegraphics[width=0.99\columnwidth]{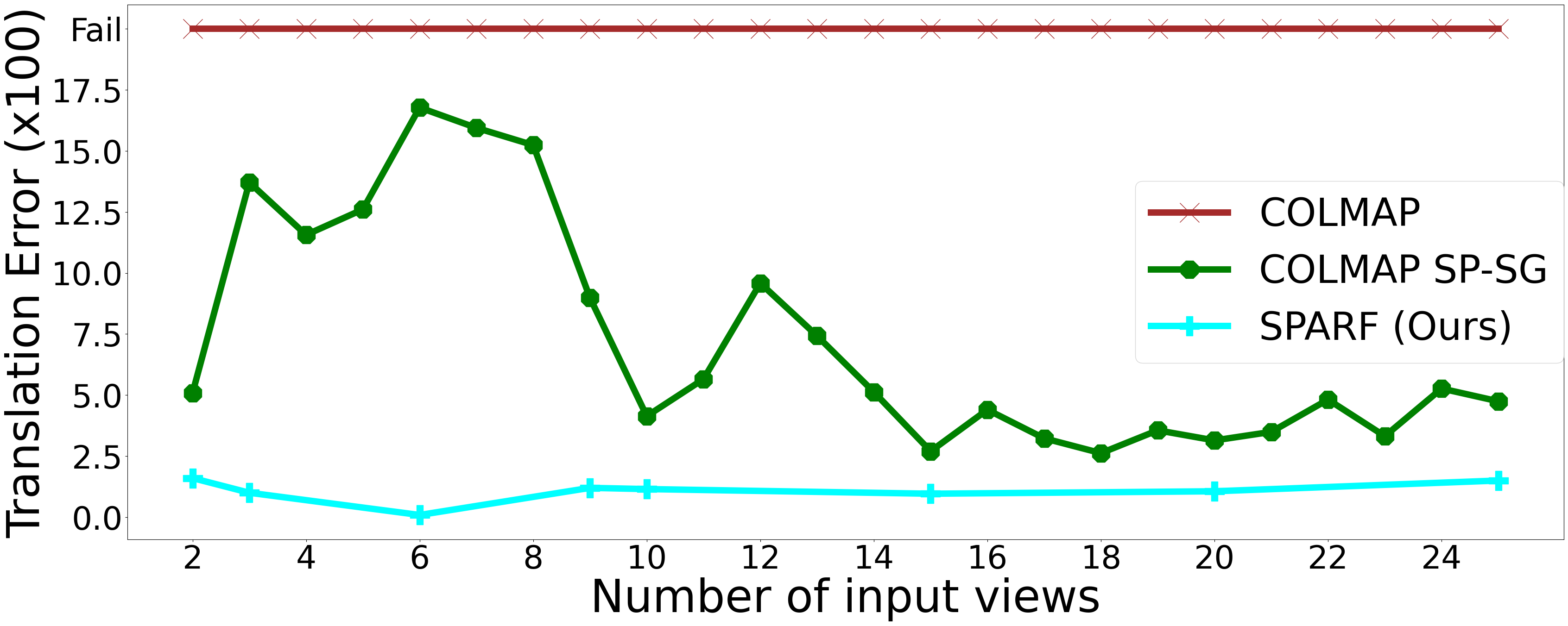} \\

\vspace{-2mm}
\caption{Rotation and translation errors versus the number of input views on a scene of the DTU dataset~\cite{dtu}. The standard COLMAP~\cite{colmap} fails to estimate initial poses for each number of input views, including relatively high numbers ($> 20$). Failing in that case means that COLMAP does not find a pose for at least one image of the set. COLMAP with better sparse matches (SuperPoint and SuperGlue~\cite{superpoint, SarlinDMR20}) performs a lot better. Nevertheless, for very few images ($<9$), the estimated poses are noisy, which can drastically impact the quality of the trained NeRF. Our approach \ours can successfully refine those poses in the sparse-view regime, and consequently, train a better-performing NeRF. Also, note that the quality of our pose refinement approach \ours stays constant when increasing the number of input images ($>9$). It consistently outperforms COLMAP-SP-SG in that regime as well. 
}
\label{fig:colmap-graphs}
\end{figure}

\begin{figure*}[t]
\centering
\includegraphics[width=0.32\textwidth]{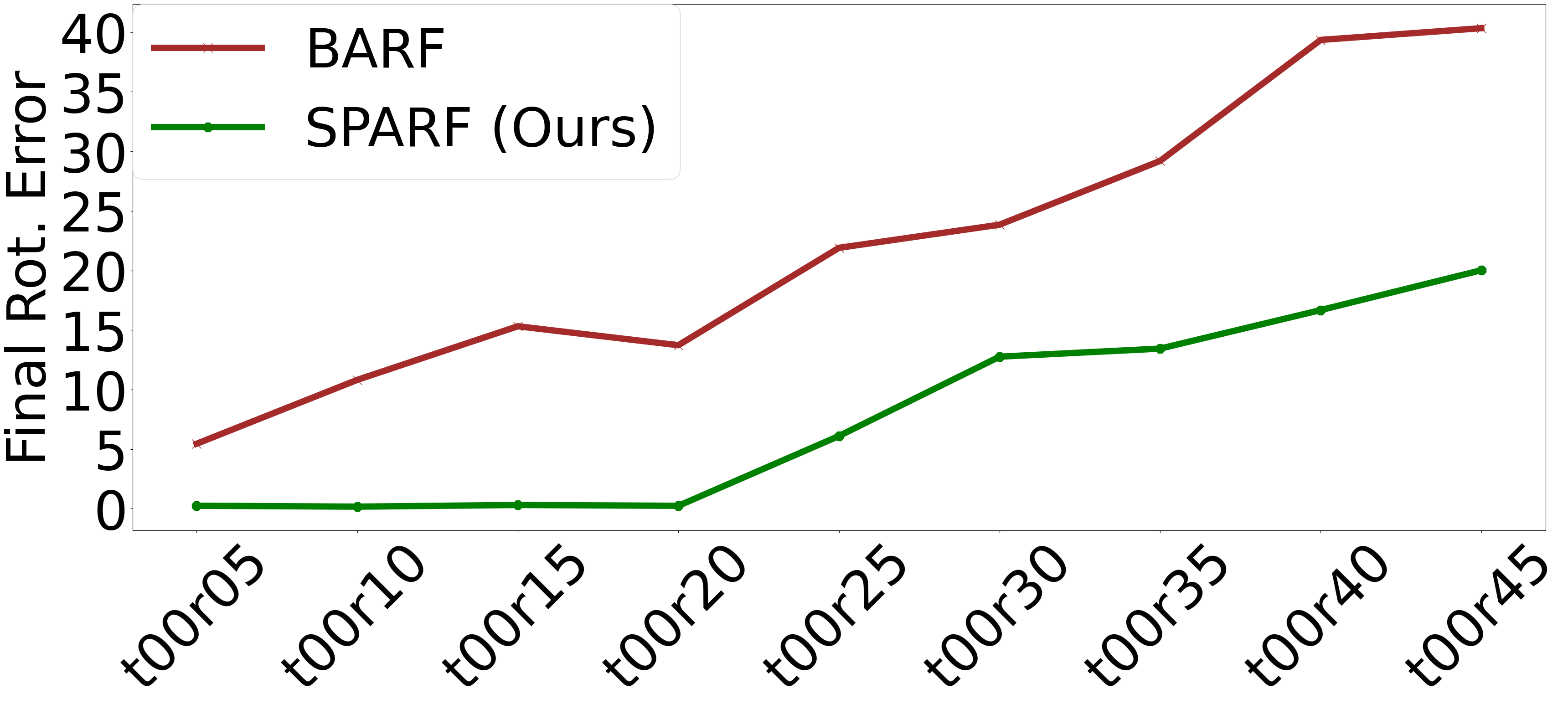}
\includegraphics[width=0.32\textwidth]{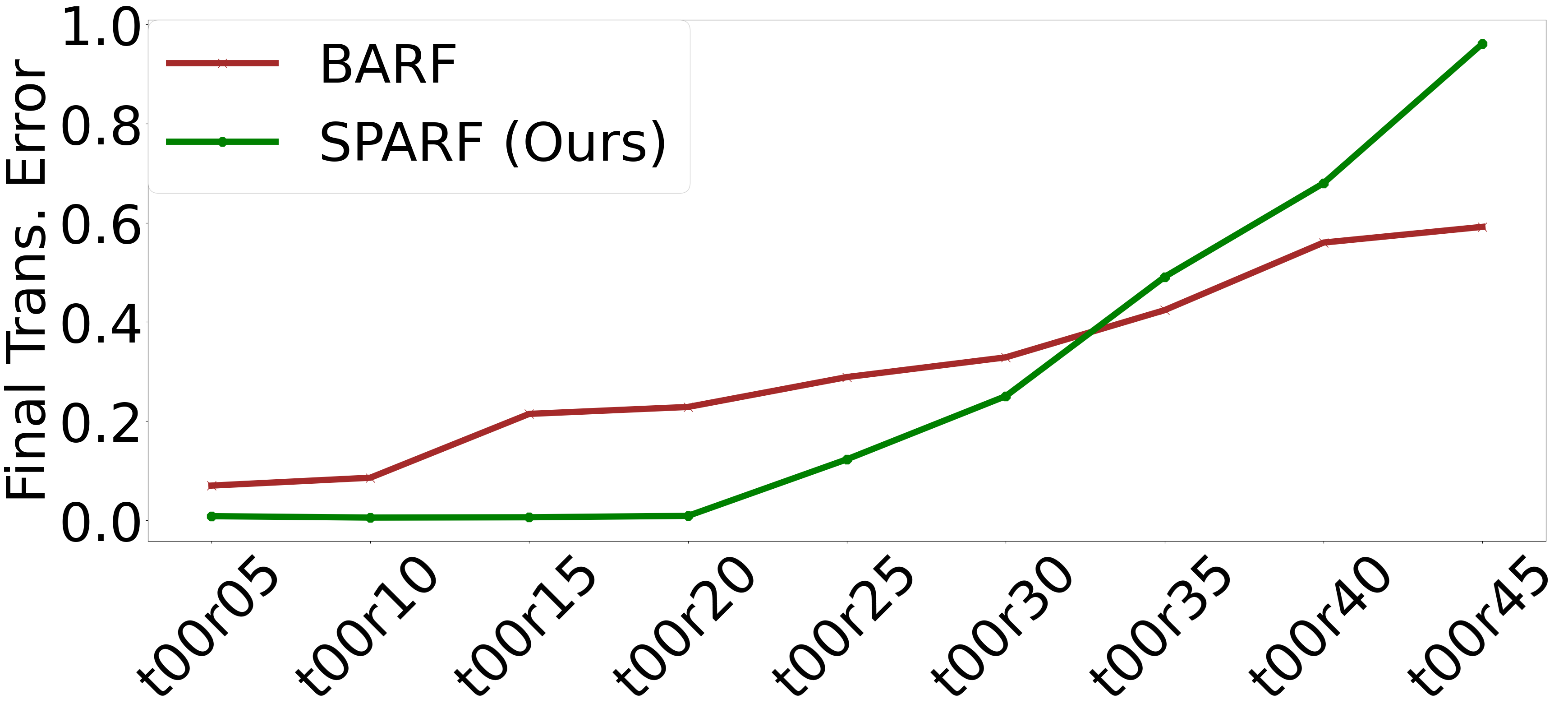}
\includegraphics[width=0.32\textwidth]{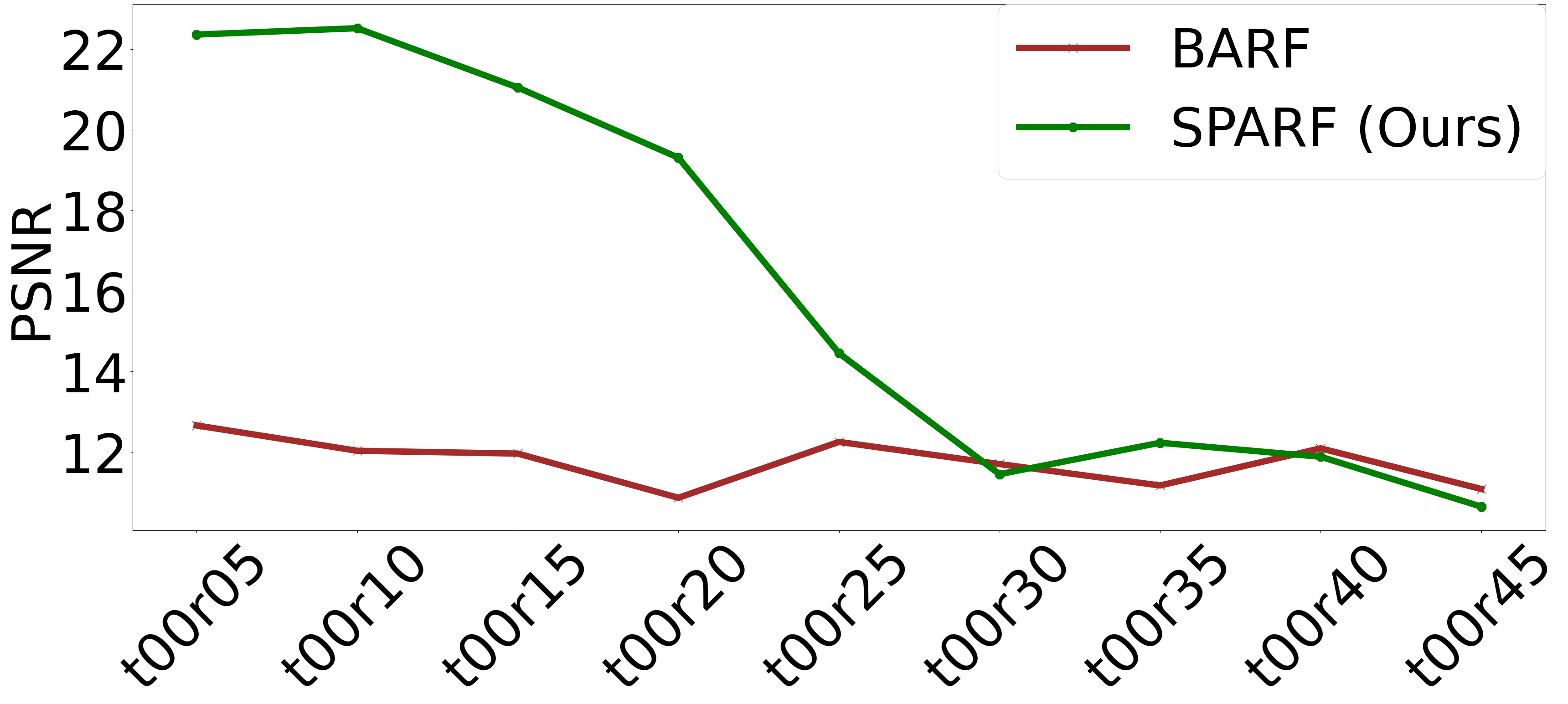} \\
(A) Varying noise in rotation \\ 
\vspace{3mm}
\includegraphics[width=0.32\textwidth]{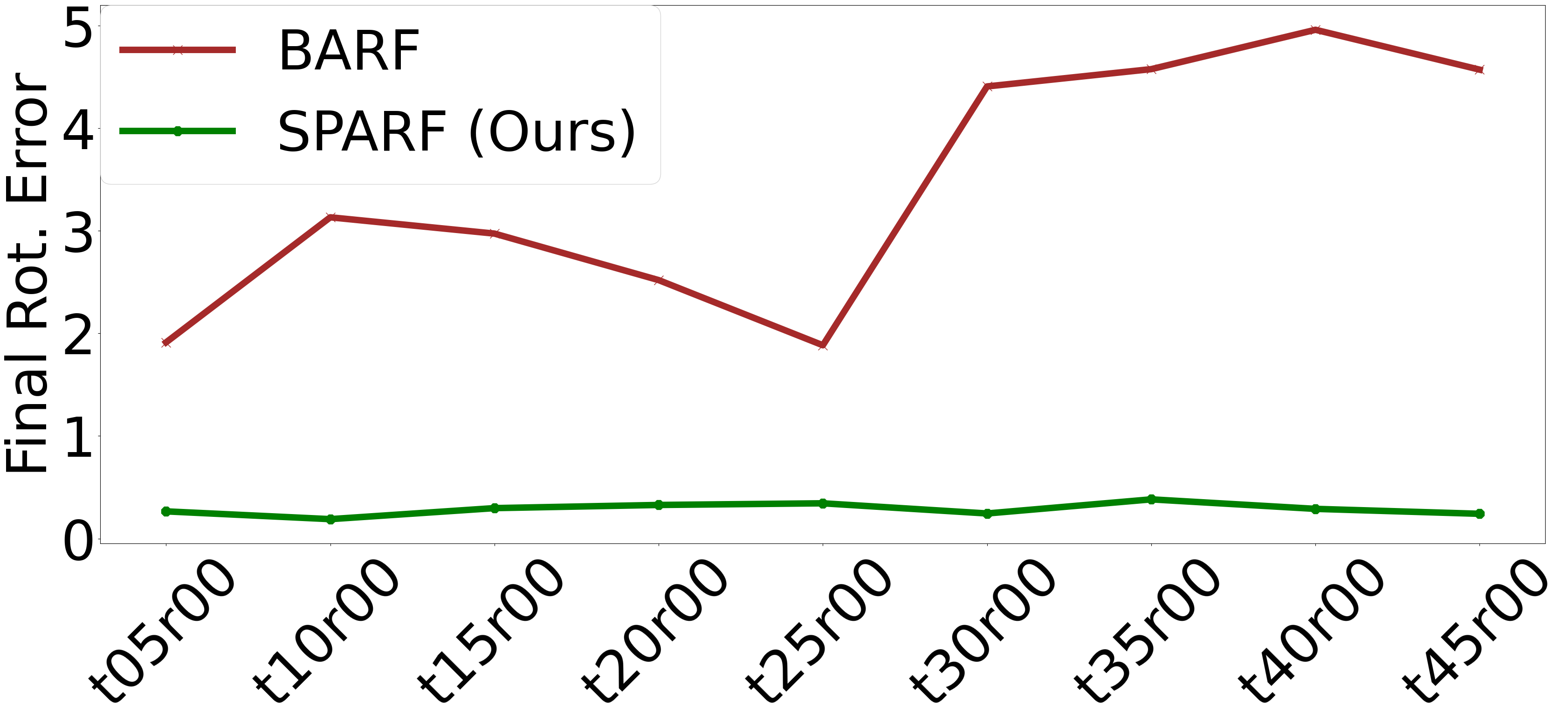}
\includegraphics[width=0.32\textwidth]{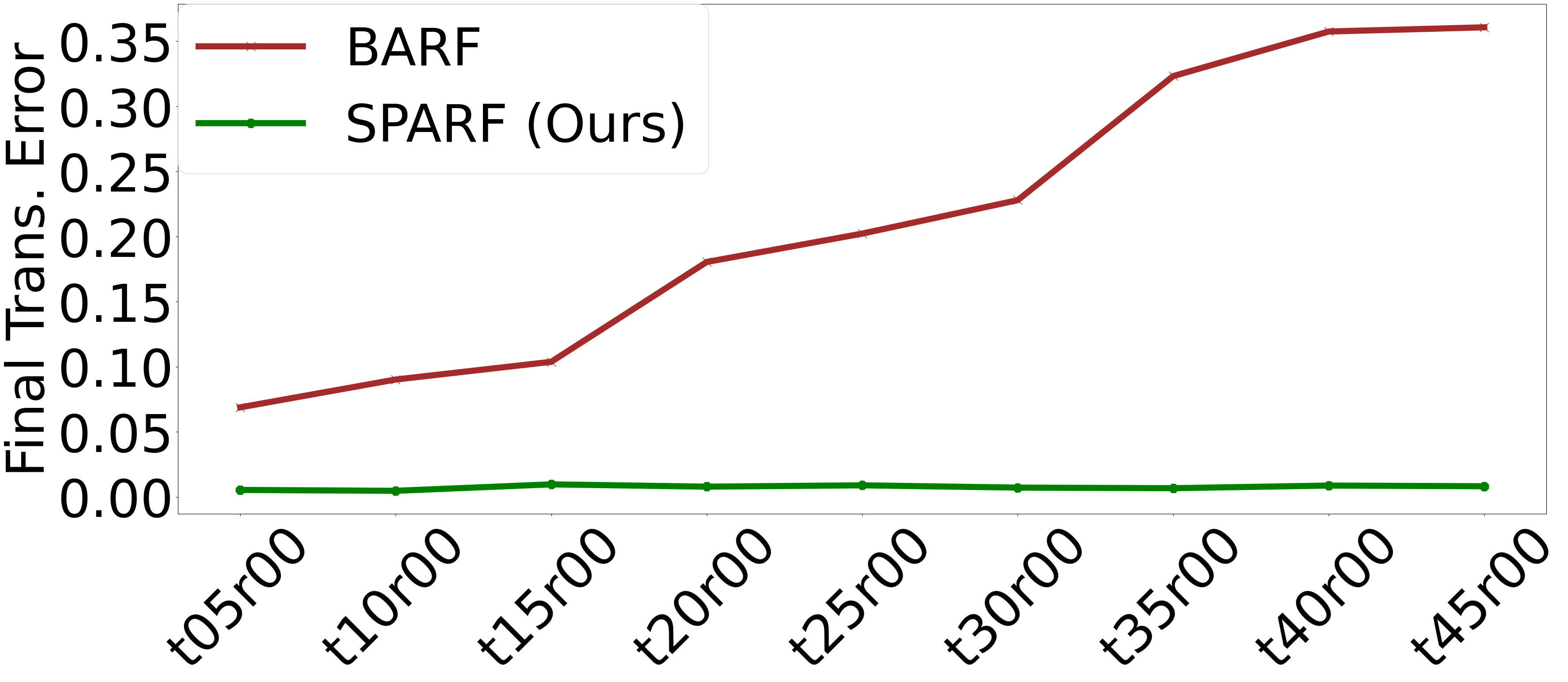}
\includegraphics[width=0.32\textwidth]{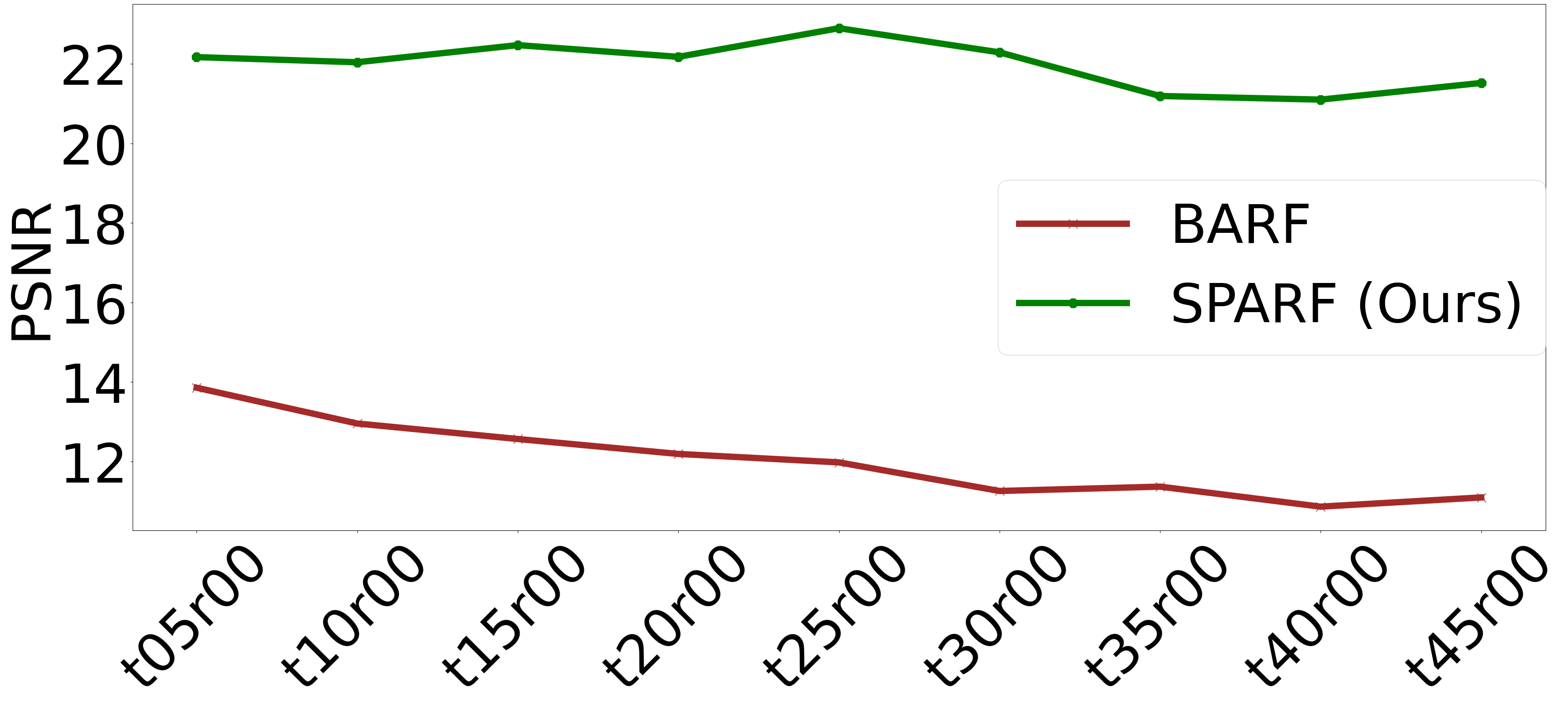} \\
(B) Varying noise in translation \\ 
\vspace{3mm}
\includegraphics[width=0.32\textwidth]{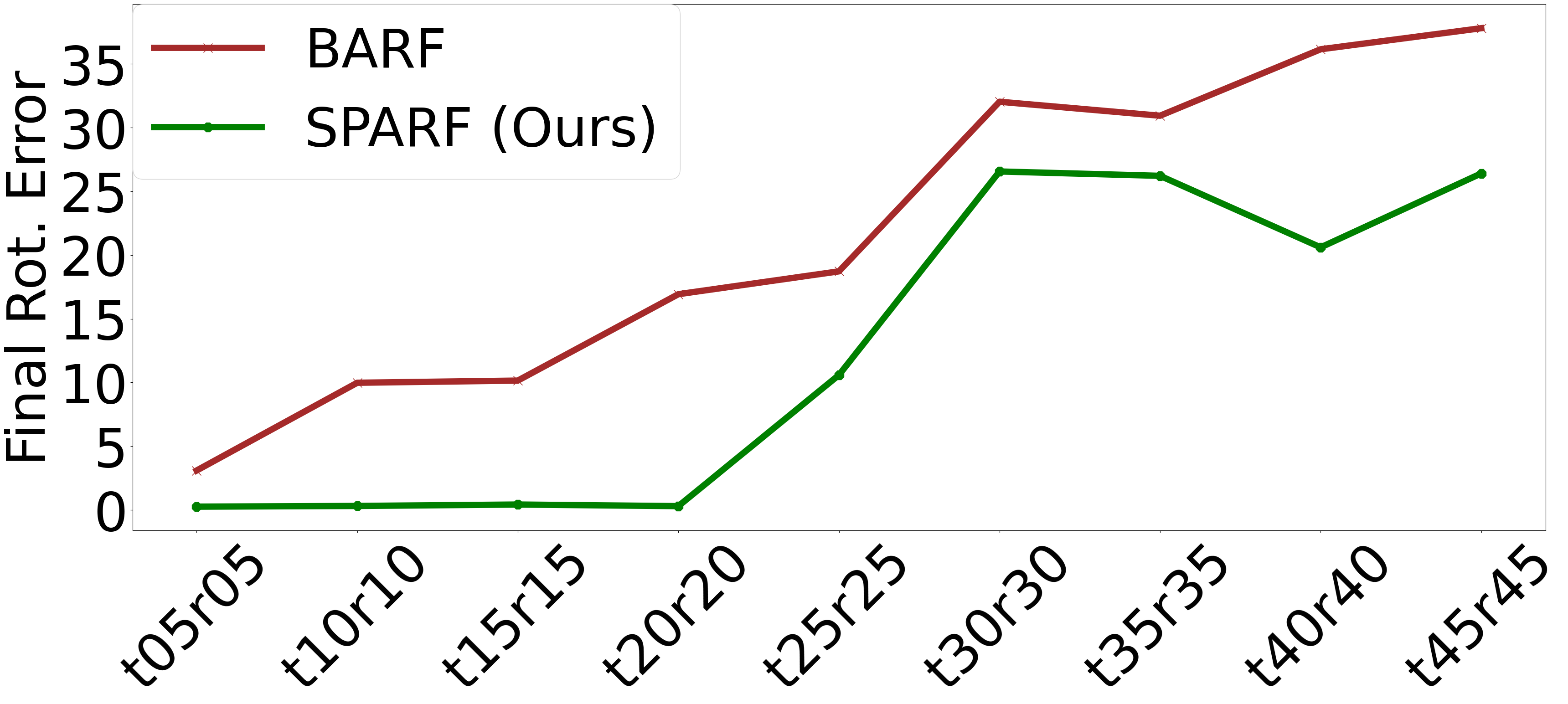}
\includegraphics[width=0.32\textwidth]{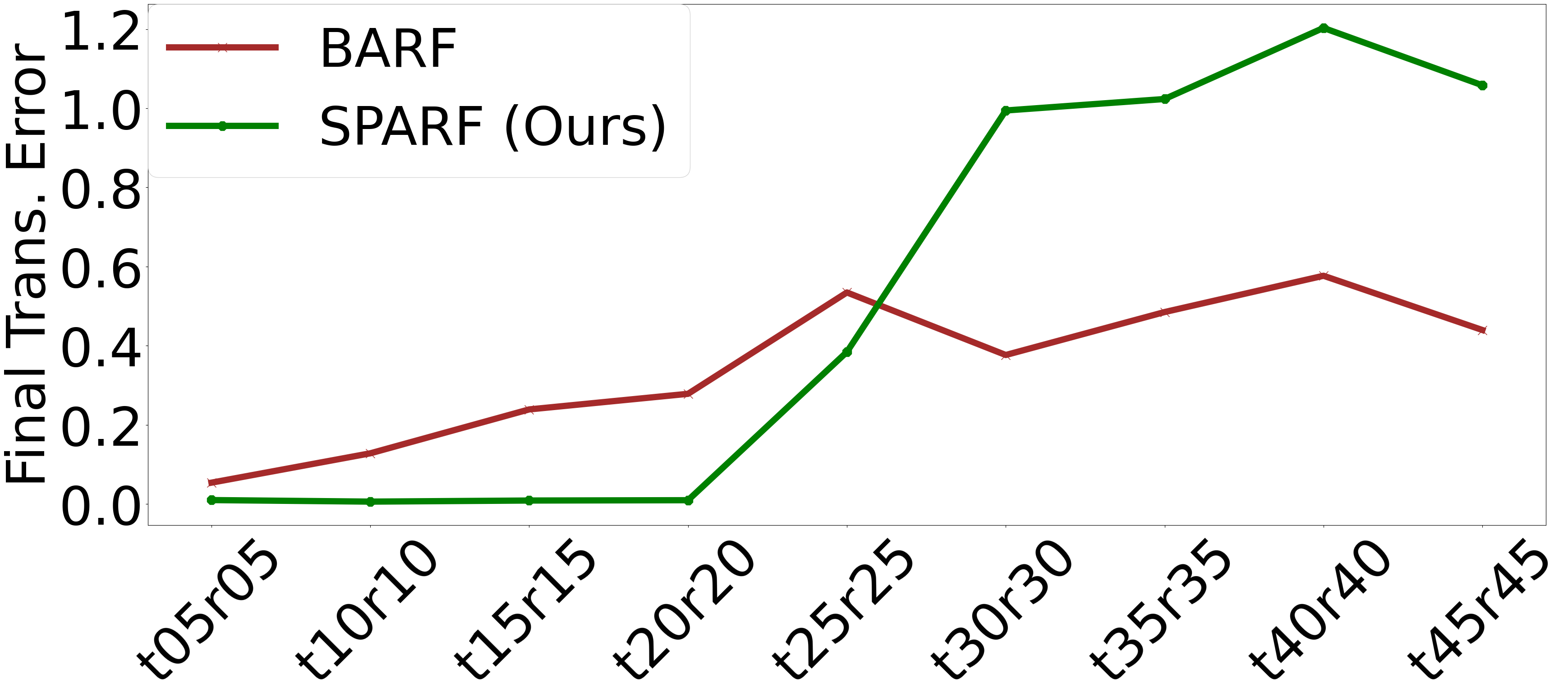}
\includegraphics[width=0.32\textwidth]{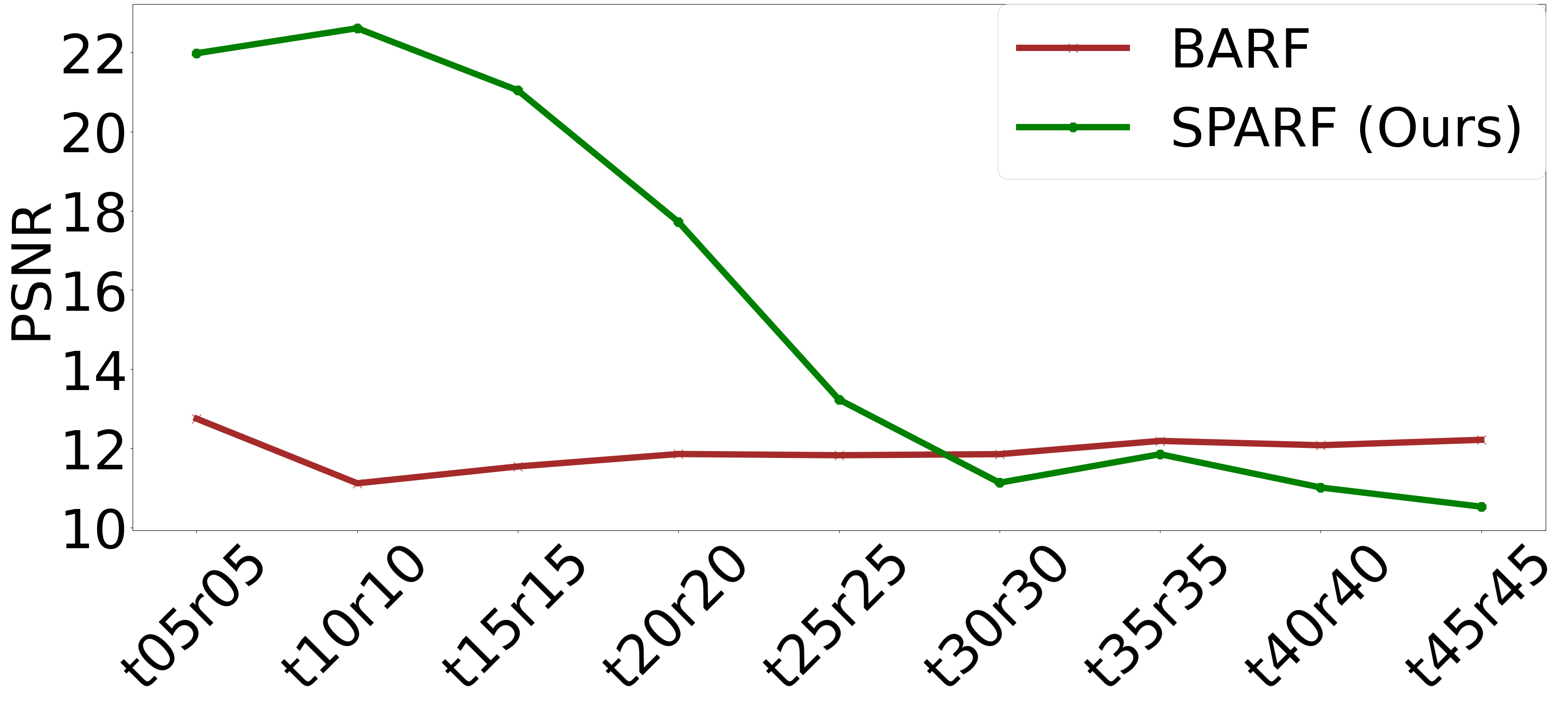}
(C) Varying noise in rotation and translation \\
\caption{Pose registration error and PSNR obtained by BARF and our \ours for different levels of initial noise. This experiment is performed on one scene of the DTU dataset, considering 3 input views. Rotation errors are in degree and translation errors are multiplied by 100. Results of PSNR ($\uparrow$) are computed by masking the background.
}
\label{fig:noise-robustness}
\end{figure*}

\begin{figure}[b]
\centering%
\includegraphics[width=0.99\columnwidth]{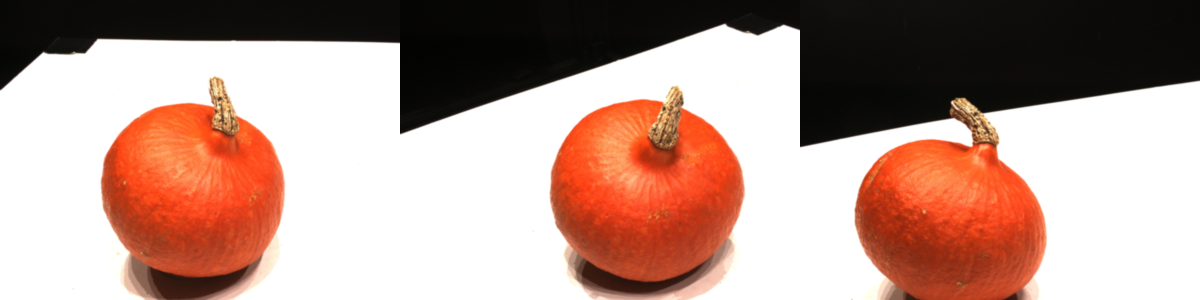}
\vspace{-2mm}
\caption{Failure case example of our approach \ours. The object, \ie the pumpkin, is almost fully symmetric with many homogeneous surfaces. The correspondence network fails to extract reliable correspondences relating the input views. As a result, our approach is unable to refine the noisy initial poses. 
}
\label{fig:failurecase}
\end{figure}

\subsection{Performance of COLMAP when reducing the number of views}

Here, we analyze the performance of COLMAP~\cite{colmap} for different numbers of input views. In Fig.~\ref{fig:colmap-graphs}, we plot the rotation and translation errors obtained by the standard COLMAP, COLMAP with SuperPoint-SuperGlue matches and our joint pose-NeRF refinement approach \ours, versus the number of input views. Even for a relatively high number of input views ($>20$), the standard COLMAP fails to estimate initial poses. This is because the images show significant viewpoint variations.
Replacing the matches with those predicted by SuperPoint and SuperGlue~\cite{superpoint, SarlinDMR20} (COLMAP SP-SG) leads to much better results. Nevertheless, for very few images ($<9$), it is very challenging to estimate high-accuracy poses. COLMAP SP-SG predicts initial poses with a rotation error between 2 and 4$\degree$, and a translation error comprised between 5.0 and 17.5. Training a NeRF with such noisy initial poses results in a drastic drop in performance compared to training with perfect input poses. Our approach \ours can successfully refine those initial poses while training the NeRF. As a result, the final optimized poses have much lower rotation and translation errors. It consequently leads to a better-performing NeRF model.

\subsection{Robustness to pose initialization and failure cases} 

\parsection{Robustness to pose initialization} We next investigate the robustness of our joint pose-NeRF refinement approach to different levels of initial noisy poses. For this experiment, our approach SPARF only uses our multi-view correspondence loss objective (Sec.~\ref{subsec:mutli-viewcons}), without our depth consistency loss (Sec.~\ref{subsec:depth-cons}) nor our staged training (Sec.~\ref{subsec:prog-training}). 
We create the noisy initial poses by synthetically perturbing the ground-truth poses with different levels of additive Gaussian noise. We present results on a randomly sampled scene of  DTU in Fig.~\ref{fig:noise-robustness}. We investigate perturbing only the rotation matrix, only the translation vector, or both in respectively (A), (B), and (C). As a reference, we also include results of BARF~\cite{barf}. Our approach \ours can handle up to 20\% of noisy rotations, which corresponds to about 20$\degree$. Interestingly, our \ours is extremely robust to translation noise, successfully registering poses with up to 45\% translation noise. When both rotation and translation noises are included, our method is robust to 20\% of noise, the rotation being the limiting factor. 

\parsection{Failure cases} Our approach \ours depends on the quality of the predicted correspondences. If only too few or inaccurate matches can be extracted between the input views, the joint pose-NeRF training will likely fail. 

It is particularly difficult to predict reliable correspondences for (almost) symmetric objects or for scenes containing many homogeneous surfaces. Such a challenging example is presented in Fig.~\ref{fig:failurecase}, which corresponds to 'scan30' of the DTU dataset. The depicted pumpkin is almost symmetric and has mostly uniform surfaces. On these images, the pre-trained correspondence network PDC-Net~\cite{pdcnet} does not predict any reliable matches. Note that the alternative matching approach SuperPoint-SuperGlue~\cite{superpoint, SarlinDMR20} is also unable to extract correspondences in that case.

\begin{table*}[t]
\centering
\resizebox{0.99\textwidth}{!}{
\begin{tabular}{@{~}l@{~}|c@{~~}c@{~}|@{~}c@{~~}c@{~~}c@{~~}c@{~~}||cc@{~~}c@{~}|@{~}c@{~~}c@{~~}c@{~~}c@{~~}}
\toprule
& \multicolumn{6}{c||}{\textbf{Over all scenes}} &   \multicolumn{7}{c}{\textbf{Over only correctly registered scenes}} \\
\toprule
& \multicolumn{2}{@{~}c@{~}|@{~}}{Pose registration} &   \multicolumn{4}{c||}{Novel-view synthesis} & \multicolumn{3}{c@{~}|@{~}}{Pose registration} &   \multicolumn{4}{c}{Novel-view synthesis} \\
& & & & & & & Nbr. corr. & & & & \\
 & Rot. $\downarrow $& Trans. $\downarrow $ & PSNR $\uparrow$ & SSIM $\uparrow$ & LPIPS $\downarrow $& DE$\downarrow $ &  sc. (/15) &  Rot. $\downarrow $& Trans. $\downarrow $ & PSNR $\uparrow$ & SSIM $\uparrow$ & LPIPS $\downarrow $& DE$\downarrow $ \\ \midrule
BARF & 10.3 & 51.5 & 10.7 (9.8) & 0.43 (0.62) & 0.59 (0.36) & 1.9 & 2 & 2.56 & 9.23 & 16.6 (17.4) & 0.66 (0.76) & 0.28 (0.18) & 0.29 \\
SCNeRF~\cite{pdcnet} & 3.44 & 16.4 & 12.0 (11.7) & 0.45 (0.66) & 0.52 (0.30) & 0.85 & 10 & 1.06 & 4.42 & 12.1 (12.6) & 0.51 (0.68) & 0.47 (0.28) & 0.80 \\
\midrule
SPARF* (PDC-Net) & \textbf{1.85} & \textbf{5.5} & \textbf{16.0} (\textbf{17.8}) & \textbf{0.68} (\textbf{0.81}) & \textbf{0.28} (\textbf{0.14}) & \textbf{0.13} & \textbf{14} &  \textbf{0.26} & \textbf{0.6} & 16.8 (\textbf{19.1}) & 0.69 (\textbf{0.81}) & 0.25 (\textbf{0.12}) & \textbf{0.08}\\ %
SPARF* (SP-SG) & 5.95 & 19.24 & 14.8 (16.1) & 0.64 (0.79) & 0.36 (0.18) & 0.19 & 11 & 0.55 & 2.05 & \textbf{17.0} (\textbf{19.1}) & \textbf{0.70} (0.80) & \textbf{0.24} (0.13) & 0.09 \\ 
\bottomrule
\end{tabular}%
}
\vspace{-2mm}
\caption{Performance of our joint pose-NeRF training, when using different pre-trained correspondence networks. The results are computed on DTU~\cite{dtu} with initial noisy poses (3 views). We simulate noisy poses by adding 15\% of random noise to the ground-truth poses. Here, SPARF* indicates that we only use the combination of the photometric loss with our multi-view correspondence objective (Sec.~\ref{subsec:mutli-viewcons}), without including our depth consistency objective (Sec.~\ref{subsec:depth-cons}) nor our staged training (Sec.~\ref{subsec:prog-training}). 
Rotation errors are in degree and translation errors are multiplied by 100. Results in ($\cdot$) are computed by masking the background. Nbr. corr. sc. designates the number of correctly registered scenes. We consider a scene to be correctly registered when the average rotation is below 10$\degree$ and the average translation is below $10$. Note that for SCNeRF~\cite{SCNeRF}, we use PDC-Net~\cite{pdcnet} correspondences. 
} 
\label{tab:diff-corres}
\end{table*}

\begin{table}[t]
\centering
\setlength{\tabcolsep}{4pt}
\resizebox{0.47\textwidth}{!}{%
\begin{tabular}{l|cc|ccc}
\toprule

  & Rot. ($\degree$) $\downarrow$ & Trans. ($\times$100) $\downarrow$ & PSNR $\uparrow$ & SSIM $\uparrow$ & LPIPS $\downarrow$\\
 \toprule
BARF~\cite{barf} & 2.04 & 11.6 & 17.47 & 0.48 & 0.37 \\ 
SCNeRF~\cite{SCNeRF} & 1.93 & 11.4 & 17.10 & 0.45 & 0.40 \\
\midrule
SPARF* (PDC-Net) &  
\textbf{0.53} & 2.8 & \textbf{19.50} & \textbf{0.61} & 0.32 \\ 

SPARF* (SP-SG) & \textbf{0.53} & \textbf{3.0} & 19.48 & 0.60 & \textbf{0.32} \\
\bottomrule
\end{tabular}%
}
\vspace{-2mm}
\caption{Performance of our joint pose-NeRF training, when using different pre-trained correspondence networks. As in Tab.~\ref{tab:diff-corres} for DTU, the evaluation is here performed on the forward-facing dataset LLFF~\cite{llff} (3 views) starting from initial identity poses. Here, SPARF* indicates that we only use the combination of the photometric loss with our multi-view correspondence objective (Sec.~\ref{subsec:mutli-viewcons}), without including our depth consistency objective (Sec.~\ref{subsec:depth-cons}) nor our staged training (Sec.~\ref{subsec:prog-training}).}
\label{tab:llff-pose-diff-corres}
\end{table}

\subsection{Impact of different correspondences} 

Our multi-view correspondence loss~\eqref{eq:mutli-cons} (Sec.~\ref{subsec:mutli-viewcons}) relies on a pre-trained correspondence network to predict matches between the training views. As stated in the main paper, while we use PDC-Net~\cite{pdcnet}, any hand-crafted or learned matching network could be used. We here compare using the dense correspondence regression network PDC-Net~\cite{pdcnet} with the state-of-the-art sparse matcher SuperGlue~\cite{SarlinDMR20}. In combination with the SuperGlue matcher, we use the SuperPoint~\cite{superpoint} detector and descriptor. 

In Tab.~\ref{tab:diff-corres}, we present results on DTU, of our joint pose-NeRF refinement approach, trained using the multi-view correspondence objective (Sec.~\ref{subsec:mutli-viewcons}) with these two alternative matching methods. As a reference, we also include results of BARF~\cite{barf} and SCNeRF~\cite{SCNeRF}. 
Sparse matchers particularly struggle to detect repeatable keypoints and predict reliable matches on images with repetitive structures and homogeneous surfaces. Dense matching approaches are more robust to these conditions. As a result, SP-SG finds an insufficient number of matches on 4 scenes out of 15, compared to 1 scene out of 15 for dense correspondence network PDC-Net. When matches are unreliable or in insufficient number, our joint pose-NeRF training is likely to fail, since our  multi-view correspondence loss~\eqref{eq:mutli-cons} relies on the predicted correspondences. As a result, when considering all scenes, SPARF* with SP-SG obtains a worse pose registration and novel-view synthesis performance than SPARF* with PDC-Net. Note nevertheless that the novel-view synthesis results are still significantly better than that of BARF and SCNeRF. 
When taking the average only over the "correctly registered scenes" instead, SPARF* with PDC-Net or SP-SG matches leads to similar pose registration and novel-view synthesis quality. 

In Tab.~\ref{tab:llff-pose-diff-corres}, we present the same comparison, on the LLFF dataset. Using PDC-Net or SP-SG matches results in a similar performance. 

\begin{figure*}[t]
\centering
\includegraphics[width=0.31\textwidth]{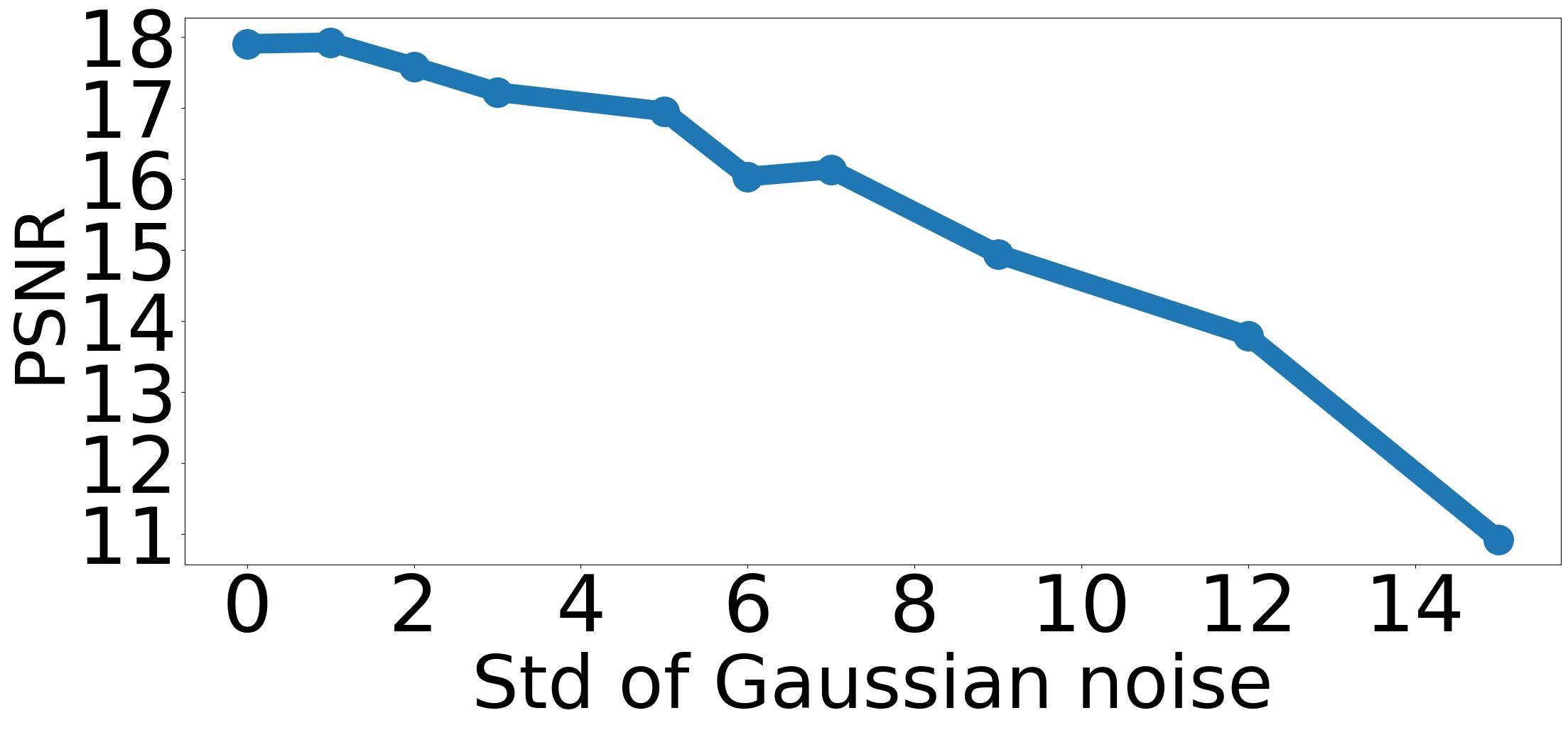}
\includegraphics[width=0.31\textwidth]{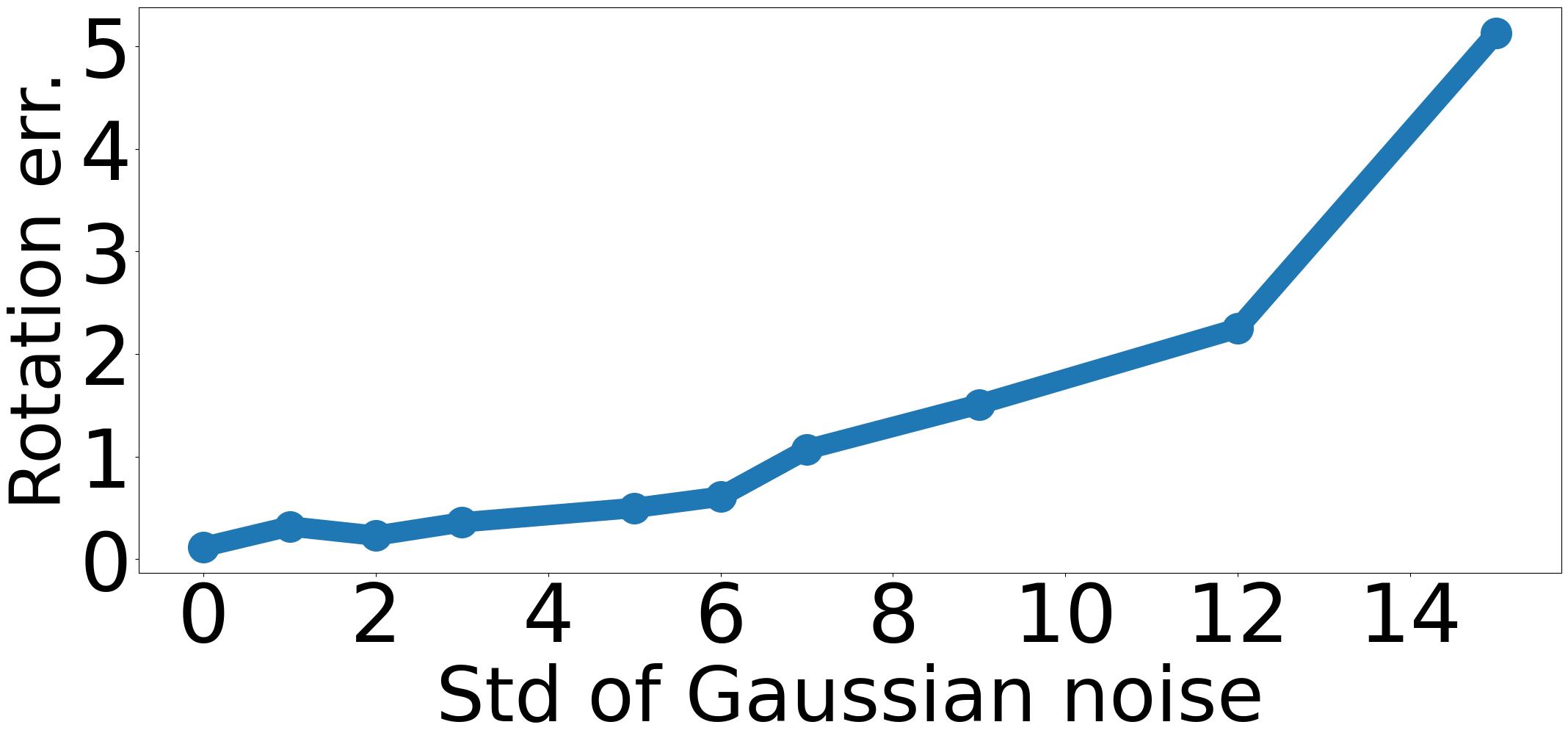}
\includegraphics[width=0.31\textwidth]{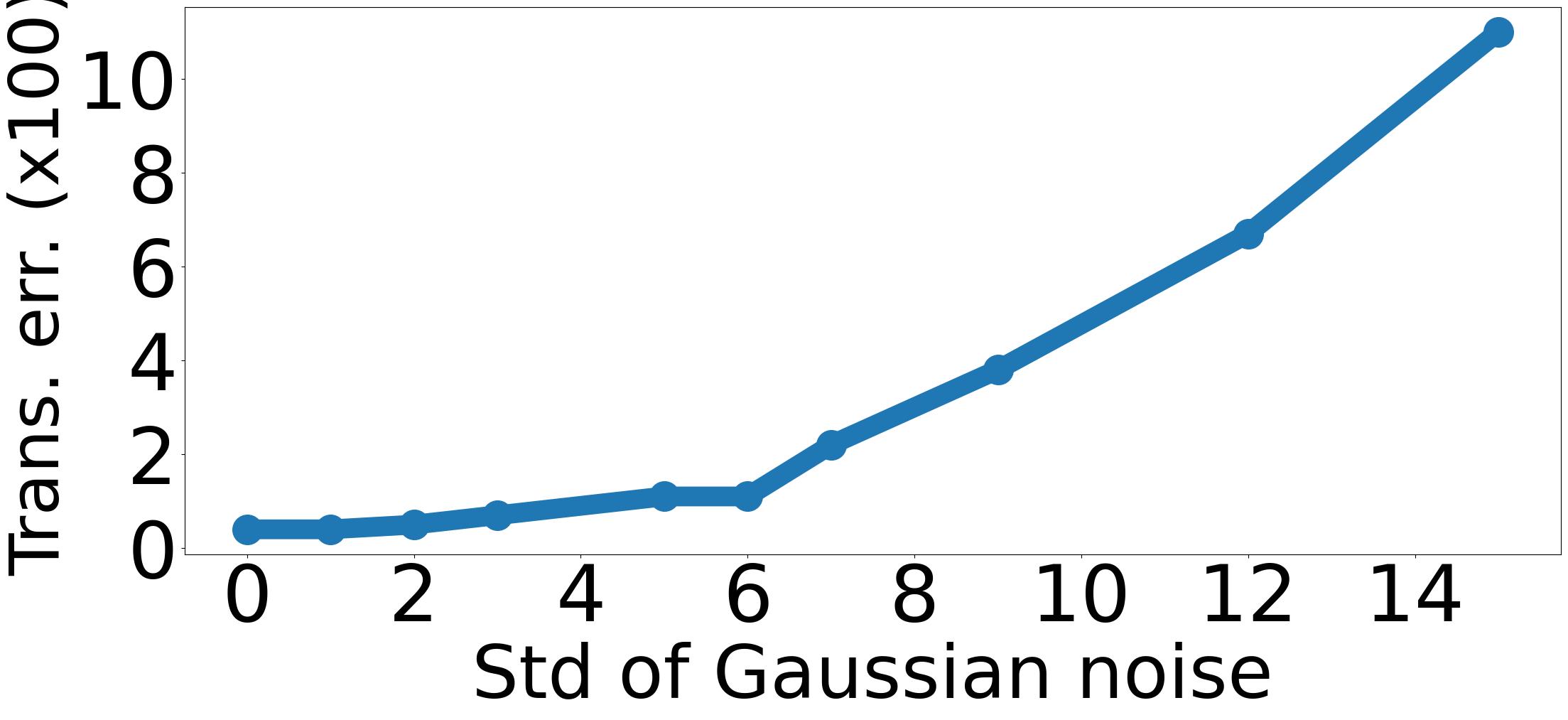}
\caption{Evaluation of SPARF* on one scene of DTU, with different levels of Gaussian noise added to ground-truth image matches. Here, SPARF* indicates that we only use the combination of the photometric loss with our multi-view correspondence objective (Sec.~\ref{subsec:mutli-viewcons}), without including our depth consistency objective (Sec.~\ref{subsec:depth-cons}) nor our staged training (Sec.~\ref{subsec:prog-training}).}
\label{fig:corres-noise}
\end{figure*}

\parsection{Impact of noisy matches} As an additional experiment, we added different levels of Gaussian noise to ground-truth matches on DTU and trained our joint pose-NeRF refinement approach \ours* using those matches. The goal is to analyze the robustness of \ours* to noisy correspondences. We conducted this experiment on one scene of DTU, with 3 input views associated with noisy poses, and present the results in Fig.~\ref{fig:corres-noise}. As previously, we consider 15\% of initial additive Gaussian noise. \ours is robust to quite noisy matches (standard-deviation up to 6 pixels) but sees its performance drop with highly erroneous correspondences.

\subsection{Additional ablation study}

\begin{table}[b]
\centering
\setlength{\tabcolsep}{4pt}
\resizebox{0.47\textwidth}{!}{%
\begin{tabular}{ll|cc|cccc}
\toprule
& & Rot. $\downarrow$ & Trans. $\downarrow $ & PSNR $\uparrow$ & SSIM $\uparrow$ & LPIPS $\downarrow $& DE $\downarrow $  \\
\toprule

I & Photo.~\eqref{eq:photo-pose-theta} & 10.3 & 51.5 & 10.7 (9.8) & 0.43 (0.62) & 0.59 (0.36) & 1.9 \\
II &+ MVCorr~\eqref{eq:mutli-cons} & 1.85 & 5.5 & 16.0 (17.8) & 0.68 (0.81) & 0.28 (0.14) & 0.13 \\ %
III &+ Staged training &  \textbf{1.81} & \textbf{5.0} &  17.58 (18.62) & \textbf{0.71} (0.82) & \textbf{0.26} (\textbf{0.13}) & 0.13 \\
IV &+ DCons~\eqref{eq:depth-cons} & \textbf{1.81} & \textbf{5.0} & \textbf{17.74} (\textbf{18.92}) & \textbf{0.71} (\textbf{0.83}) & \textbf{0.26} (\textbf{0.13}) & \textbf{0.12} \\
\midrule
II & Fully joint pose-NeRF & 1.85 & 5.5 & 16.0 (17.8) & 0.68 (0.81) & 0.28 (0.14) & 0.13 \\ %
III & Staged training (Sec.~\ref{subsec:prog-training}) &  \textbf{1.81} & \textbf{5.0} &  17.58 (18.62) & 0.71 (0.82) & 0.26 (0.13) & 0.13 \\

V & Restart NeRF & 1.84 & 5.3 & \textbf{17.80} (\textbf{19.07}) & \textbf{0.72} (\textbf{0.83}) & \textbf{0.25} (\textbf{0.12}) & \textbf{0.12}  \\ 
\bottomrule
\end{tabular}%
}\vspace{-2mm}
\caption{Ablation study on DTU~\cite{dtu} (3 views) with noisy initial poses. In the top part, from (I) to (IV), we progressively add ($+$) each component. In the bottom part, we compare multiple training schedules for the joint pose-NeRF training. The depth consistency loss (Sec.~\ref{subsec:depth-cons}) is then not included. 
Rotation errors are in degree and translation errors are multiplied by 100. Results in ($\cdot$) are computed by masking the background. }
\label{tab:dtu-ablation-refinement}
\end{table}

\parsection{Ablation study for joint pose-NeRF refinement} In Tab.~\ref{tab:dtu-pixelnerf-ablation-fixed-pose} of the main paper, we ablated key components of our approach, considering fixed ground-truth poses on the DTU dataset. Here, we ablate our approach when refining initial noisy poses along with training the NeRF model. As previously, we consider 15\% of initial additive Gaussian noise. We present results in the top part of Tab.~\ref{tab:dtu-ablation-refinement}. 
From (I) to (II), adding our multi-view correspondence loss~\eqref{eq:mutli-cons} leads to drastically better pose registration than training with only the photometric loss~\eqref{eq:photo-pose-theta} (I). The rendering quality also radically improves. This is in part due to the better pose registration, which is necessary to obtain a decent rendering quality. It is also enabled by the fact that our multi-view correspondence loss not only drives the camera poses but also applies direct supervision on the rendered depth, enforcing it to be close to the surface. As such, it enables learning an accurate scene geometry. 
In (III), we introduce our staged training (Sec.~\ref{subsec:prog-training}), which is composed of two parts. In the first stage, we refine the poses while training the coarse network $F^{c}_{\theta}$. In the second part, we freeze the pose estimates and train both the coarse and fine networks $F^{c}_{\theta}$ and $F^{f}_{\theta}$. Comparing (II) to (III), we observe that introducing this second stage leads to better PSNR and SSIM metrics. This is because the fine network can learn a sharp geometry benefiting from the frozen, registered camera poses and the pre-trained coarse network. On the other hand, when jointly training the camera poses and both coarse and fine MLP (II), the learned scene often has a slightly blurry surface due to the exploration of the pose space. 
Finally, further including our depth consistency objective (Sec.~\ref{subsec:depth-cons}) slightly improves the rendering performance, leading to the best results overall.

\parsection{Comparison of different training schedules}  In the bottom part of Tab.~\ref{tab:dtu-ablation-refinement}, we further compare different training schedules for joint pose-NeRF training. As previously explained, jointly training the poses with both the coarse and fine MLPs in (II) can lead to blurry surfaces. As demonstrated in (III), our staged training (Sec.~\ref{subsec:prog-training}) largely solves this problem, leading to better rendering quality. Nevertheless, it is worth noting that the best results are obtained with the NeRF restarting approach corresponding to (V). In (V), the NeRF is first jointly trained with the poses. Once the poses have converged, the optimized pose estimates are frozen and both coarse and fine MLPs are re-initialized. Both MLPs are then trained from scratch, considering fixed optimized poses. This approach can remove some of the artifacts learned during the pose optimization, that might still be present in our staged training (III). 
This restarting approach was also found to be the best alternative in~\cite{nerfmm}.

\parsection{Impact of visibility mask in depth consistency loss} In Sec.~\ref{subsec:depth-cons} of the main paper, we introduce our depth consistency loss. However, the proposed loss is only valid in pixels of the training views for which the projections in the virtual view are not occluded by the reconstructed scene, seen from the virtual view. We therefore use a visibility mask, following the same formulation as~\cite{neuralwarp}. We ablate the impact of this visibility mask in the depth consistency loss formulation in Tab.~\ref{tab:dtu-pixelnerf-ablation-fixed-pose-mask}. We observe that removing the visibility mask leads to a notable drop in performance in PSNR, probably because the NeRF model learns surfaces that are actually occluded, leading to artifacts in the geometry and therefore the renderings.

\begin{table}[b]
\vspace{-3mm}
\centering
\resizebox{0.48\textwidth}{!}{
\begin{tabular}{@{~}l@{~}|c|c|c|c}
\toprule
 & PSNR $\uparrow$ & SSIM $\uparrow$  & LPIPS $\downarrow$ & DE $\downarrow $ \\ \toprule
MVCorr~\eqref{eq:mutli-cons} of m.p. & 18.13 (20.81) & 0.77 (\textbf{0.87}) & 0.22 (\textbf{0.10}) & 0.10 \\ %
+ DCons~\eqref{eq:depth-cons} of m.p.& \textbf{18.30} (\textbf{21.01}) & \textbf{0.78} (\textbf{0.87}) & \textbf{0.21} (\textbf{0.10}) & \textbf{0.08} \\ 
No Vis mask (Sec.~\ref{subsec:depth-cons})& 17.89 (21.05) & 0.77 (0.86) & 0.22 (0.11) & 0.09 \\
\bottomrule
\end{tabular}%
}\vspace{-2mm}
\caption{Impact of the visibility mask for our depth consistency loss (Sec.~\ref{subsec:depth-cons} of the main paper). 
Results are computed on the DTU dataset (3 views), with fixed ground-truth poses. Results in ($\cdot$) are computed by masking the background. All networks use the coarse-to-fine PE~\cite{barf}. }
\label{tab:dtu-pixelnerf-ablation-fixed-pose-mask}
\end{table}

\begin{table*}[t]
\centering
\setlength{\tabcolsep}{4pt}
\resizebox{0.99\textwidth}{!}{%
\begin{tabular}{ll|cc|cccc||cc|cccc}
\toprule
& & \multicolumn{6}{c||}{\textbf{Initial COLMAP SP-SG}} &   \multicolumn{6}{c}{\textbf{Initial COLMAP PDCNet}} \\
& & \multicolumn{6}{c||}{\textbf{Rot: 1.34$\degree$, Trans ($\times$100): 6.84 }} &   \multicolumn{6}{c}{\textbf{Rot: 0.75$\degree$, Trans ($\times$100): 3.87 }} \\
\toprule

&  & Rot. ($\degree$) $\downarrow$ & Trans ($\times$100) $\downarrow$ & PSNR $\uparrow$ & SSIM $\uparrow$ & LPIPS $\downarrow$ & DE $\downarrow$ &  Rot. ($\degree$) $\downarrow$ & Trans ($\times$100) $\downarrow$ & PSNR $\uparrow$ & SSIM $\uparrow$ & LPIPS $\downarrow$ & DE $\downarrow$ \\
\toprule
G & \textbf{\ours} (Ours) &  \multicolumn{2}{c|}{Fixed GT poses} & \best{18.56} (\best{20.84}) & \best{0.77} (\best{0.86}) & \best{0.22} (\best{0.11}) & \best{0.08} & \multicolumn{2}{c|}{Fixed GT poses} & \best{18.56} (\best{20.84}) & \best{0.77} (\best{0.86}) & \second{0.22} (\best{0.11}) & \best{0.08} \\
\midrule
F & NeRF~\cite{Nerf} &  \multicolumn{2}{c|}{Fixed poses obtained }  & 8.95 (9.77) & 0.30 (0.60) & 0.72 (0.38) & 1.25 &  \multicolumn{2}{c|}{Fixed poses obtained } & 8.88 (9.66) & 0.31 (0.62) & 0.73 (0.37) & 1.28 \\

& DS-NeRF~\cite{DSNerf}&  \multicolumn{2}{c|}{from COLMAP (run w. } & 11.89 (13.28) & 0.46 (0.69) & 0.49 (0.25) & 0.38 & \multicolumn{2}{c|}{from COLMAP (run w. } & 11.61 (12.81) & 0.46 (0.70) & 0.51 (0.25) & 0.60 \\

& DS-NeRF w. CF PE~\cite{DSNerf, barf} &\multicolumn{2}{c|}{SP-SG~\cite{SarlinDMR20} matches)}  & 16.58 (17.58) & 0.66 (0.77) & 0.29 (0.17) & 0.21 & \multicolumn{2}{c|}{ PDC-Net~\cite{pdcnet} matches)}  & 18.10 (19.30) & 0.71 (0.80) & 0.24 (0.13) & \second{0.12} \\ 
& \textbf{\ours} (Ours) &  & & 17.34 (17.92) & 0.68 (0.78) & \second{0.26} (0.14) & 0.15 & &  & 18.42 (19.61) & 0.72 (0.82) & \second{0.22} (\second{0.12}) & \second{0.12} \\
\midrule
R & BARF~\cite{barf} & 4.90 & 12.74 & 13.14 (13.01) & 0.52 (0.69) & 0.45 (0.25) & 0.55 & 3.5 & 11.94 & 14.27 (14.59) & 0.56 (0.70) & 0.39 (0.23) & 0.54 \\
& RegBARF~\cite{Regnerf, barf}  & 4.3 & 11.0 & 14.65 (15.30) & 0.6 (0.73) & 0.38 (0.22) & 0.25 & 3.71 & 9.81 & 15.22 (15.98) & 0.60 (0.73) & 0.36 (0.22) & 0.25 \\
& SCNeRF~\cite{SCNeRF} & \second{0.97} & \second{3.08} & 15.94 (16.73) & 0.63 (0.75) & 0.32 (0.19) & 0.43 & \second{1.08} & \second{3.3} & 15.94 (16.42) & 0.63 (0.75) & 0.32 (0.18) & 0.43 \\
& DS-NeRF~\cite{DSNerf} & 3.7 & 10.0 & 13.67 (14.30) & 0.54 (0.72) & 0.40 (0.22) & 0.21 & 2.66 & 7.58 & 16.00 (16.87) & 0.63 (0.77) & 0.31 (0.17) & 0.24 \\ 
& \textbf{\ours} (Ours) &  \best{0.35} & \best{0.9} & \second{18.39} (\second{19.67}) & \second{0.73} (\second{0.82}) & \best{0.22} (\second{0.12}) & \second{0.09} & \best{0.3} & \best{0.7} & \second{18.52} (\second{20.00}) & \second{0.73} (\second{0.83}) & \best{0.21} (\best{0.11}) & \best{0.08} \\   
\bottomrule
\end{tabular}%
}\vspace{-2mm}
\caption{Evaluation on 14 scenes of the DTU dataset (3 views) with initial poses obtained by COLMAP using SP-SG~\cite{SarlinDMR20} (left) or PDCNet~\cite{pdcnet} (right) matches. Note that both approaches fail to obtain the initial poses on one of the pre-defined 15 test scenes ('scan30'), which we therefore excluded from this evaluation. In the middle part (F), the initial poses are fixed and used as "pseudo-ground-truth". In the bottom part (R), the poses are refined along with training the NeRF. For comparison, in the top part (G), we use fixed ground-truth poses. All methods in the bottom part (R), which perform joint pose-NeRF training, use the coarse-to-fine PE approach~\cite{barf} (Sec.~\ref{subsec:prog-training} of m.p.). Results in ($\cdot$) are computed by masking the background. The best and second-best results are in red and blue respectively.
}
\label{tab:suppl-dtu-pixelnerf-pose}
\end{table*}

\begin{table*}[t]
\centering
\setlength{\tabcolsep}{4pt}
\resizebox{0.99\textwidth}{!}{%
\begin{tabular}{l|cc|cccc|cc|cccc|cc|cccc}
\toprule
 & \multicolumn{6}{c|}{\textbf{3 input views}} & \multicolumn{6}{c|}{\textbf{6 input views}} & \multicolumn{6}{c}{\textbf{9 input views}} \\

 & Rot. $\downarrow$ & Trans. $\downarrow $ & PSNR $\uparrow$ & SSIM $\uparrow$ & LPIPS $\downarrow $& DE $\downarrow $  &  Rot. $\downarrow$ & Trans. $\downarrow $ & PSNR $\uparrow$ & SSIM $\uparrow$ & LPIPS $\downarrow $& DE $\downarrow $   &  Rot. $\downarrow$ & Trans. $\downarrow $ & PSNR $\uparrow$ & SSIM $\uparrow$ & LPIPS $\downarrow $& DE $\downarrow $   \\
\toprule
BARF~\cite{barf} & 10.33 & 51.5 & 10.71 (9.76) & 0.43 (0.62) & 0.59 (0.36) & 1.9 &  9.20 & 31.1 & 14.02 (14.22) & 0.54 (0.69) & 0.46 (0.27 ) & 0.49 & 8.34 & 26.72 & 16.20 (16.38) & 0.60 (0.73) & 0.38 (0.22) & 0.35\\ 
RegBARF~\cite{barf, Regnerf}  & 11.2 & 52.8 & 10.38 (9.20) & 0.45 (0.62) & 0.61 (0.38) & 2.33 & 9.19 & 26.63 & 14.59 (14.58) & 0.57 (0.70) & 0.44 (0.27) & 0.32 & 5.28 & 18.51 & 18.98 (19.08) & 0.67 (0.77) & 0.29 (0.18) &  0.23 \\
DistBARF~\cite{barf, mipnerf360} & 11.69 & 55.7 & 9.50 (9.15) & 0.34 (0.76) & 0.67 (0.36) & 1.90 & 8.96 & 28.85 & 14.31 (14.60) & 0.55 (0.70) & 0.43 (0.26) & 0.53 & 7.00 & 26.42 & 16.18 (16.27) & 0.58 (0.71) & 0.37 (0.22) & 0.29 \\
SCNeRF~\cite{SCNeRF} & 3.44 & 16.4 & 12.04 (11.71) & 0.45 (0.66) & 0.52 (0.30) & 0.85 & 4.10 & 12.80 & 17.76 (18.16) & 0.70 (0.80) & 0.31 (0.18) & 0.28 & 4.76 & 16.25 & 18.19 (18.01) & 0.69 (0.81) & 0.31 (0.17) & 0.31\\
\textbf{\ours} (Ours) & \textbf{1.81} & \textbf{5.0} & \textbf{17.74} (\textbf{18.92}) & \textbf{0.71} (\textbf{0.83}) & \textbf{0.26} (\textbf{0.13}) & \textbf{0.12} &  \textbf{1.31} & \textbf{2.7} & \textbf{21.39} (\textbf{22.01}) & \textbf{0.81} (\textbf{0.88}) & \textbf{0.18} (\textbf{0.10}) & \textbf{0.09} & \textbf{1.15} & \textbf{2.55} & \textbf{24.69} (\textbf{25.05}) & \textbf{0.88} (\textbf{0.92}) & \textbf{0.12}(\textbf{0.06}) & \textbf{0.06}
\\ %
\midrule
\textbf{\ours} - No 'scan30' & 0.36 & 0.8 & 18.13 (19.53) & 0.72 (0.82) & 0.22 (0.11) & 0.09 & 0.39 & 1.05 & 22.34 (23.16) & 0.83 (0.88) & 0.14 (0.08) & 0.05 & 0.25 & 0.8 & 25.35 (25.86) & 0.88 (0.92) & 0.10 (0.06) & 0.04 \\ 
\bottomrule
\end{tabular}%
}\vspace{-2mm}
\caption{Evaluation on DTU~\cite{dtu} with different numbers of input views (3, 6, or 9) and considering noisy initial poses. We simulate noisy poses by adding $15\%$ of Gaussian noise to the ground-truth poses. 
The results for 3 input views correspond to Tab.~\ref{tab:dtu-pixelnerf-pose} of the main paper and are repeated here for ease of comparison. Rotation errors are in $\degree$ and translation errors are multiplied by 100. Results in ($\cdot$) are computed by masking the background. }
\label{tab:dtu-pixelnerf-manyviews-pose}
\end{table*}

\section{Additional Results with Initial Noisy Poses}
\label{sec-sup:results-noisy}

In this section, we provide additional results considering initial noisy poses. In particular, we experiment with different initialization schemes. We also use different numbers of input views and present extensive qualitative results. Finally, we experiment with training considering all available training views (\ie 25), instead of a subset.

\subsection{Results on the DTU dataset}

Here, we present additional results for our joint pose-NeRF refinement approach \ours, evaluated on the DTU dataset~\cite{dtu}. In the main paper, we showed results when considering three input views and starting from initial noisy poses, created by synthetically perturbing ground-truth poses. Here, we first evaluate starting from an alternative initialization scheme, in particular initial poses obtained by COLMAP~\cite{colmap}. 
Moreover, we also evaluate for different numbers of input views, in particular 6 or 9. We also show multiple qualitative comparisons for the 3-view setting.

\begin{figure}[t]
\centering
\includegraphics[width=0.99\columnwidth]{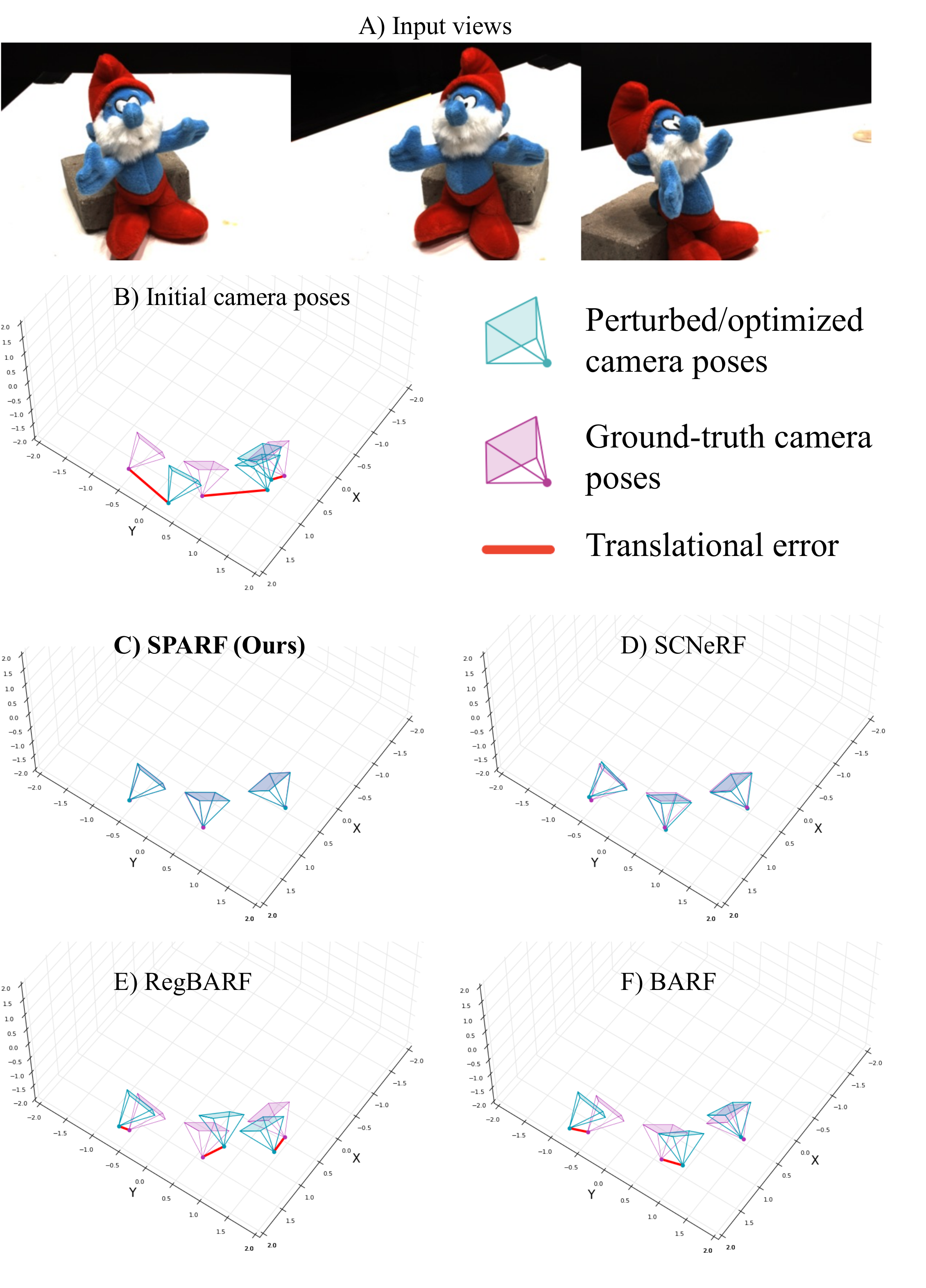}
\vspace{-3mm}
\caption{Initial and optimized poses on one scene of the DTU dataset, given 3 input views. 
}

\label{fig:noise-dtu}
\end{figure}

\parsection{Initialization with COLMAP} On the DTU input images, COLMAP~\cite{colmap} mostly fails when reducing the number of input views to 3 (see Fig.~\ref{fig:colmap-graphs}). As a result, to obtain the initial camera pose estimates, we experiment with COLMAP run with matches predicted by SuperPoint and SuperGlue~\cite{SarlinDMR20} (SP-SG) or PDC-Net~\cite{pdcnet}. Both COLMAP-SP-SG and COLMAP-PDCNet fail to obtain initial pose estimates on one out of the 15 scenes composing the test set ('scan30', see Fig.~\ref{fig:failurecase}). We thus present results on the remaining 14 scenes in Tab.~\ref{tab:suppl-dtu-pixelnerf-pose}. In the middle part of the table (F), we fix the initial poses, which we consider as "pseudo-ground-truth", and train the NeRF model. In the bottom part (R), we instead compare multiple joint pose-NeRF refinement approaches. Finally, in the top part (G), we present the results of SPARF, trained considering fixed ground-truth poses for reference. 

SP-SG sometimes struggles with homogeneous surfaces, where it is difficult to extract repeatable keypoints. It leads to an initial rotation and translation error of respectively 1.34$\degree$ and 6.84. PDC-Net, which can heavily rely on smoothness properties when predicting dense matches, performs better on homogeneous regions. It results in slightly better initial poses, \ie with an initial rotation and translation error of 0.75$\degree$ and 3.87 respectively. 

For both initialization schemes, the trend is the same. Considering the COLMAP poses as "pseudo-ground-truth" and training the NeRF with fixed poses (part F) leads to significantly worse results than when using ground-truth poses (top part, G), particularly in PSNR and SSIM. This is because the NeRF learns artifacts caused by the wrong positioning of the poses. 
Instead, using our approach to jointly refine the poses and train the NeRF (R) narrows the gap between fixed COLMAP poses (F) and the ideal case of fixed ground-truth poses (G). Note that the latter case of fixed ground-truth poses is unrealistic in practice. Notably, when refining the poses, \ours obtains similar performance in LPIPS and depth error compared to the fixed ground-truth pose version. The lower PSNR and SSIM values indicate that the NeRF model still learns artifacts during the joint refinement. Note that this issue can be partially circumvented by re-initializing the NeRF model and training from scratch with fixed poses, once the poses have converged (see Tab.~\ref{tab:dtu-ablation-refinement}).

\parsection{Results with 6 and 9 views} In Tab.~\ref{tab:dtu-pixelnerf-pose} of the main paper, we evaluate our proposed approach \ours for joint pose-NeRF training, when considering only \emph{3 input views}. For completeness, we here provide results when 6 or 9 input views are available. As in the 3-view setting, we synthetically perturb the ground-truth poses by adding 15\% of additive Gaussian noise. The results are presented in Tab.~\ref{tab:dtu-pixelnerf-manyviews-pose}. 
We included the results with 3 input views for ease of comparison. 
The trend is similar for 3, 6, or 9 input views. BARF, RegBARF, and DistBARF struggle to refine the initial noisy poses, leading to poor novel-view rendering performance. While increasing the number of views leads to better synthesis quality, it remains drastically lower than the performance obtained by our \ours. 
SCNeRF performs better at registering the poses. The rendering quality and learned geometry are nevertheless much worse than the proposed \ours. 

With 3, 6, or 9 input views, our \ours outperforms all previous works. For completeness, we also provide results of our approach when excluding one of the scenes, \ie 'scan30', on which no correspondences are found. When excluding this scene, the rotation and translation errors of the optimized scenes are below 1$\degree$ and 1 (multiplied by 100) respectively. The average novel-view rendering performance is also significantly increased.

\begin{table}[t]
\centering
\setlength{\tabcolsep}{4pt}
\resizebox{0.47\textwidth}{!}{%
\begin{tabular}{l|cc|cccc}
\toprule

  & Rot. $\downarrow$ & Trans. $\downarrow$ & PSNR $\uparrow$ & SSIM $\uparrow$ & LPIPS $\downarrow$ & DE $\downarrow$\\
 \toprule
BARF~\cite{barf} & 2.46 & 6.72 & 21.67 (21.71) & 0.77 (0.84) & 0.21 (0.13) & 0.14  \\ 

\textbf{SPARF} (Ours) & \textbf{1.0} & \textbf{1.23} & \textbf{24.77} (\textbf{24.41}) & \textbf{0.85} (\textbf{0.89}) & \textbf{0.15} (\textbf{0.10}) & \textbf{0.05} \\ 
\bottomrule
\end{tabular}%
}
\vspace{-2mm}
\caption{Evaluation on DTU, considering all available training views (25) and initial noisy poses. We simulate noisy poses by adding $15\%$ of Gaussian noise to the ground-truth poses. It leads to an initial rotation and translation error of 13.36$\degree$ and 47.87 respectively. Rotation errors are in degree and translation errors are multiplied by 100. Results in ($\cdot$) are computed by masking the background. Also note that some of the training images have inconsistent illumination, making them unsuitable for the NeRF training. 
}
\label{tab:dtu-all-views}
\end{table}

\begin{table*}[t]
\centering
\setlength{\tabcolsep}{4pt}
\resizebox{0.99\textwidth}{!}{%
\begin{tabular}{l|cc|ccc|cc|ccc|cc|ccc}
\toprule
 & \multicolumn{5}{c|}{\textbf{2 input views}} & \multicolumn{5}{c|}{\textbf{6 input views}} & \multicolumn{5}{c}{\textbf{9 input views}} \\
 & Rot. $\downarrow$ & Trans. $\downarrow $ & PSNR $\uparrow$ & SSIM $\uparrow$ & LPIPS $\downarrow $ &  Rot. $\downarrow$ & Trans. $\downarrow $ & PSNR $\uparrow$ & SSIM $\uparrow$ & LPIPS $\downarrow $  &  Rot. $\downarrow$ & Trans. $\downarrow $ & PSNR $\uparrow$ & SSIM $\uparrow$ & LPIPS $\downarrow $   \\
 
\toprule
BARF~\cite{barf} & 5.16 & 37.87 & 15.06 & 0.35 & 0.50 & \best{0.25} & \second{0.37} & \second{23.09} & \best{0.72} & \best{0.22} & \second{0.20} & \second{0.34} & \second{24.10} & \best{0.76} & \best{0.20} \\ %
RegBARF~\cite{Regnerf, barf}  & \second{3.77} & \second{28.59} & \second{15.94} & \second{0.40} & \second{0.47} & \second{0.48} & 0.55 & 22.21 & \second{0.68} & 0.26 & 0.93 & 4.2 & 22.68 & \second{0.70} & \second{0.26} \\
DistBARF~\cite{barf, mipnerf360} & 7.32 & 110.0 & 14.06 & 0.30 & 0.55  & 2.38 & 11.23 & 18.31 & 0.52 & 0.37 & 2.81 & 13.41 & 20.36 & 0.59 & 0.34   \\ 
SCNeRF~\cite{SCNeRF} & 4.88 & 44.27 & 14.43 & 0.32 & 0.51 & 2.07 & 8.11 & 21.82 & 0.66 & 0.26 & 0.47 & 3.87 & 22.72 & 0.70 & 0.24 \\

\textbf{\ours} (Ours) & \best{1.54} & \best{8.38} & \best{17.32} & \best{0.47} & \best{0.40} & \best{0.25} & \best{0.32} & \best{23.30} & \best{0.72} & \second{0.23} & \best{0.18} & \best{0.30} & \best{24.12} & \best{0.76} & \best{0.20} \\ 
\bottomrule
\end{tabular}%
}\vspace{-2mm}
\caption{Evaluation on LLFF~\cite{llff} with different numbers of input views (2, 6, or 9) and starting from initial identity poses. The results for 3 input views can be found in Tab.~\ref{tab:llff-pose} of the main paper. Rotation errors are in $\degree$ and translation errors are multiplied by 100.  The best and second-best results are in red and blue respectively. }
\vspace{-3mm}\label{tab:llff-manyviews-pose}
\end{table*}

\begin{figure*}[t]
\centering%
\vspace{-15mm}
\includegraphics[width=0.8\textwidth]{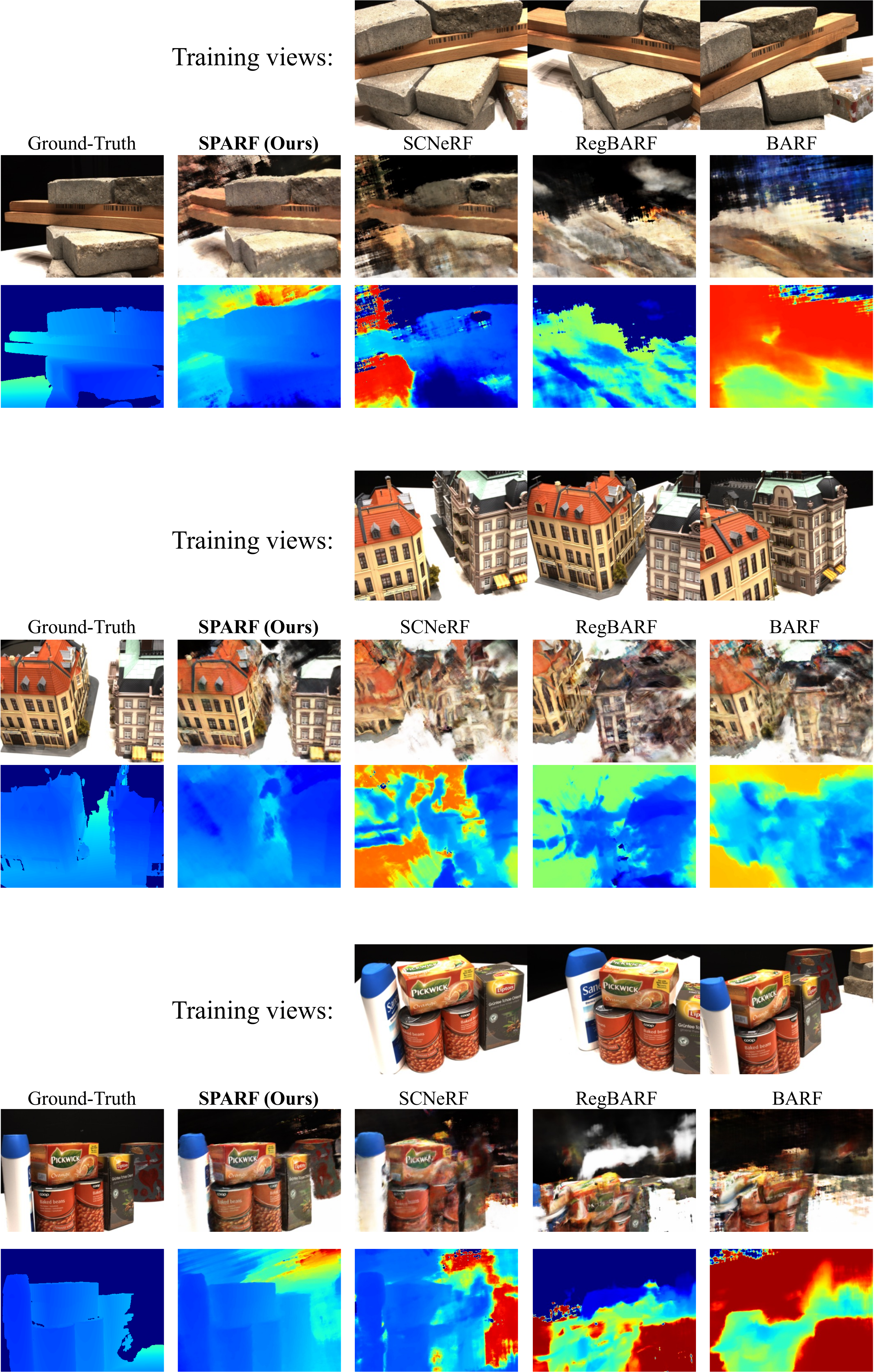}
\vspace{-2mm}
\caption{Novel-view renderings of alternative joint pose-NeRF training approaches on the DTU dataset. For each scene, we show the RGB (first row) and depth (second row) renderings from an unseen viewpoint.  We consider 3 input views with initial noisy poses. The initial camera poses are created by perturbing the ground-truth poses with 15\% of additive Gaussian noise.  
}
\vspace{-10mm}
\label{fig:qual-dtu-comp}
\end{figure*}

\parsection{Qualitative comparisons} We provide qualitative comparisons for the 3-view regime. In Fig.~\ref{fig:noise-dtu}, we show the initial and optimized poses on one scene of DTU. We visually compare the novel-view renderings (RGB and depth) of our \ours, SCNeRF, BARF, and RegBARF in Fig.~\ref{fig:qual-dtu-comp}. 

\begin{figure*}[t]
\centering%
\vspace{-15mm}
\includegraphics[width=0.99\textwidth]{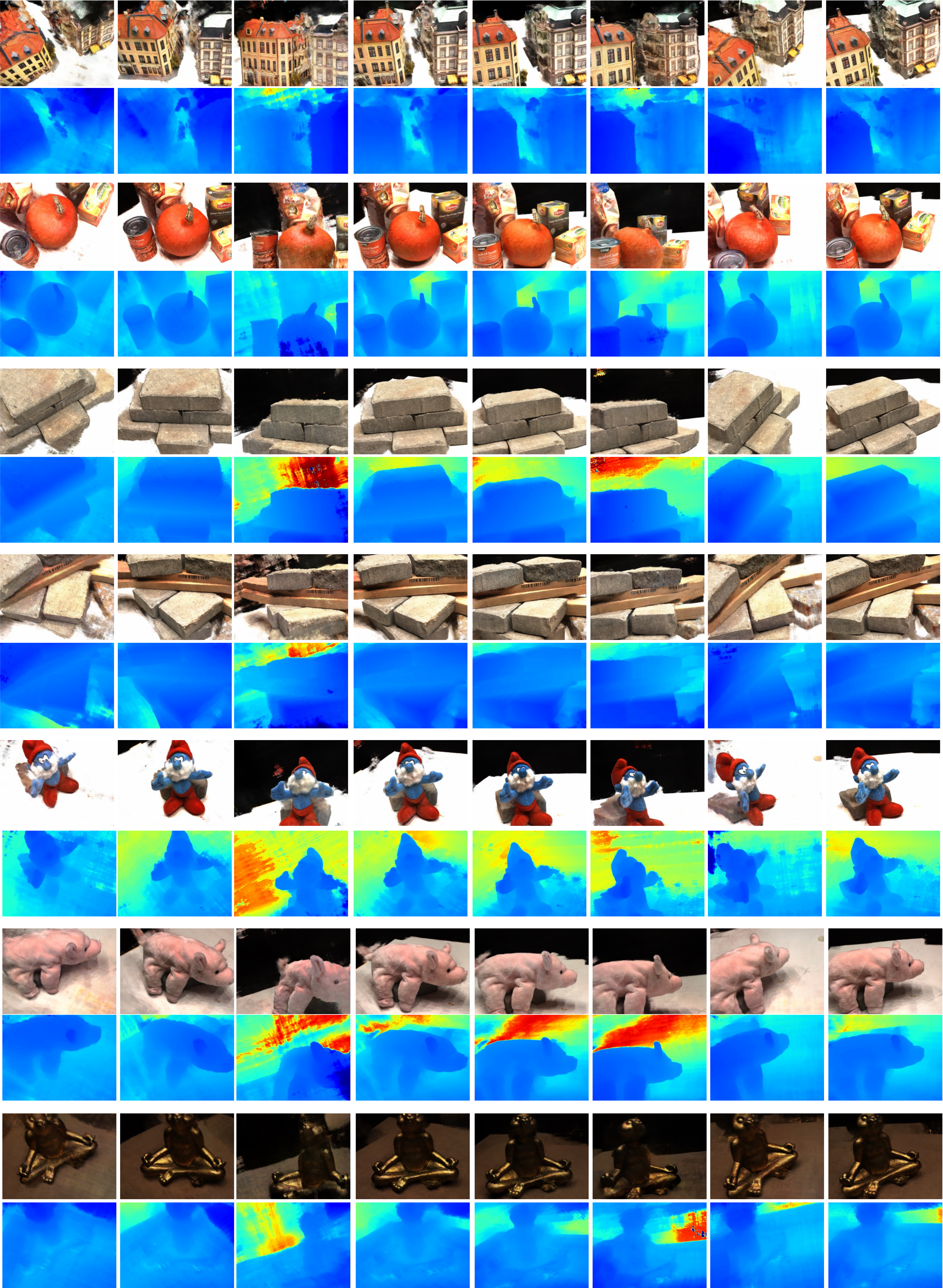}
\vspace{-2mm}
\caption{Novel-view renderings of our \ours on the DTU dataset.  For each scene, we show the RGB (first row) and depth (second row) renderings from multiple unseen viewpoints. In each scene, we consider 3 input views (not shown here) with initial noisy poses, created by perturbing the ground-truth poses with 15\% of additive Gaussian noise.  
}
\vspace{-10mm}
\label{fig:qual-dtu}
\end{figure*}

Finally, we provide extensive examples of the novel-view synthesis capabilities of our approach \ours in Fig.~\ref{fig:qual-dtu}. It produces realistic novel views with accurate geometry on a large variety of scenes and from many different viewing directions, given only 3 input views with noisy initial poses.

\parsection{Results with all views} For completeness, we evaluate our joint pose and NeRF training approach \ours, when many input views are available. While this is not the goal of this work, which was specifically designed for the sparse-view regime, we show here that it can generalize to the many-view setting. We present results on DTU in Tab.~\ref{tab:dtu-all-views}. Even in this setting, our \ours significantly outperforms baseline BARF~\cite{barf} in pose registration and novel-view synthesis performance. We note that some of the training images have inconsistent illumination, which were excluded when considering subsets. Inconsistent illumination can cause problems when training a NeRF since it relies on the photometric loss as the primary training signal. This explains why the PSNR and SSIM values obtained by \ours with all 25 input views (Tab.~\ref{tab:dtu-all-views}) are slightly worse than when trained on only a subset of 9 views (Tab.~\ref{tab:dtu-pixelnerf-manyviews-pose}).

\subsection{Results on the LLFF dataset}

Here, we present additional results for our joint pose-NeRF refinement approach \ours, evaluated on the LLFF dataset~\cite{dtu}. 

\begin{table}[b]
\centering
\setlength{\tabcolsep}{4pt}
\resizebox{0.47\textwidth}{!}{%
\begin{tabular}{l|cc|ccc}
\toprule

  & Rot. ($\degree$) $\downarrow$ & Trans. (x 100) $\downarrow$ & PSNR $\uparrow$ & SSIM $\uparrow$ & LPIPS $\downarrow$ \\
 \toprule
BARF~\cite{barf} & 0.85 & 0.26 & 25.09 & 0.77 & \textbf{0.20} \\

SPARF* & \textbf{0.77} & \textbf{0.23} & \textbf{25.18} & \textbf{0.78} & \textbf{0.20} \\
\bottomrule
\end{tabular}%
}
\vspace{-2mm}
\caption{Evaluation on LLFF~\cite{llff}, considering all available training views and initial identity poses. Here, SPARF* indicates that we only use the combination of the photometric loss with our multi-view correspondence objective (Sec.~\ref{subsec:mutli-viewcons}), without including our depth consistency objective (Sec.~\ref{subsec:depth-cons}) nor our staged training (Sec.~\ref{subsec:prog-training}).
}
\label{tab:llff-all-views}
\end{table}

\parsection{Results with 2, 6 and 9 views} As for DTU~\cite{dtu}, we here evaluate our pose-NeRF refinement approach when \emph{6 or 9} input views are available instead of only 3.  For completeness, we also include results when only 2 views are available. 

When considering 2 or 3 input views, BARF struggles to refine the poses, which impacts its novel-view synthesis performance.
Nevertheless, LLFF represents forward-facing scenes, for which a limited number of homogeneously spread views can cover the majority of the scene. As a result, for 6 input views and more, the 3D space is sufficiently constrained for BARF to successfully register the initial identity poses. In the 6 and 9 view cases, our approach \ours and BARF obtain similar performance in pose registration and novel-view rendering quality. 

Interestingly, while adding the depth regularization loss (RegBARF) to the photometric loss (BARF) helps the pose registration and novel-view rendering performance in the 2 and 3-view regimes, it is harmful with denser views (6 and 9). Our approach \ours instead does not negatively impact the performance of BARF in the 6 and 9-view scenarios. Surprisingly, SCNeRF obtains worse registration and novel-view rendering results than BARF, and consequently our approach \ours. 

\begin{figure*}[t]
\centering
\vspace{-5mm}
\includegraphics[width=0.97\columnwidth]{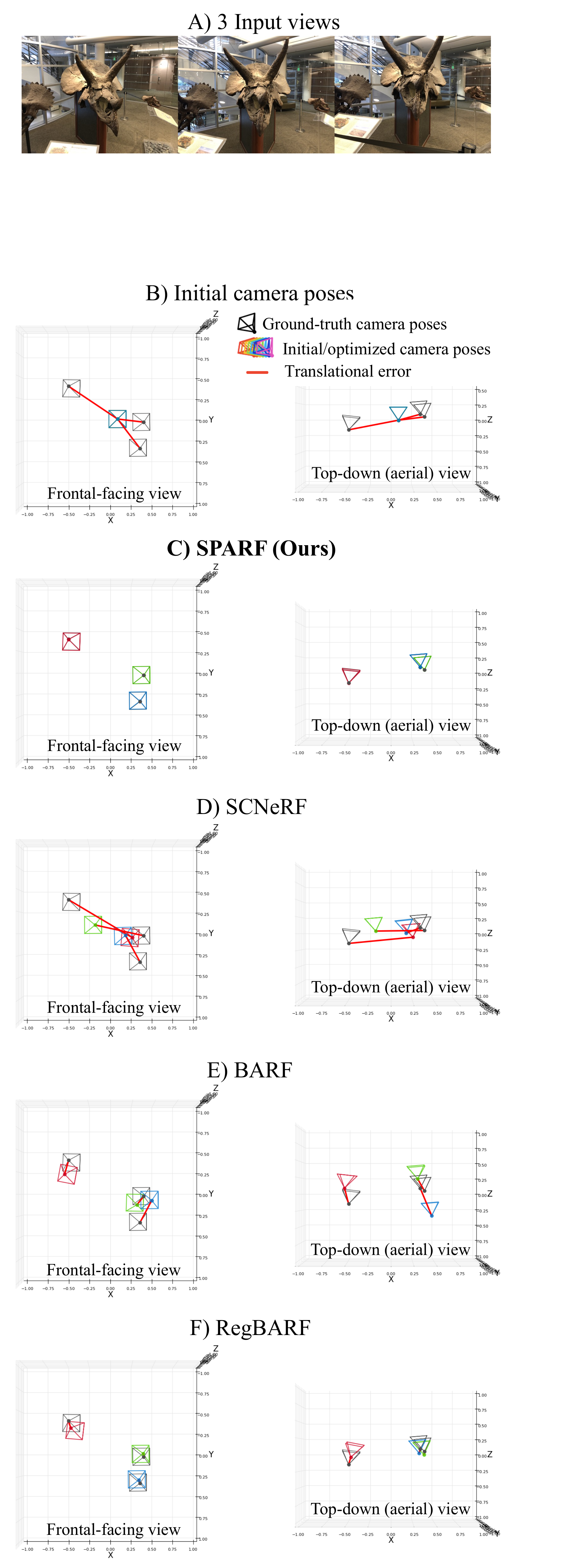} 
\includegraphics[width=0.97\columnwidth]{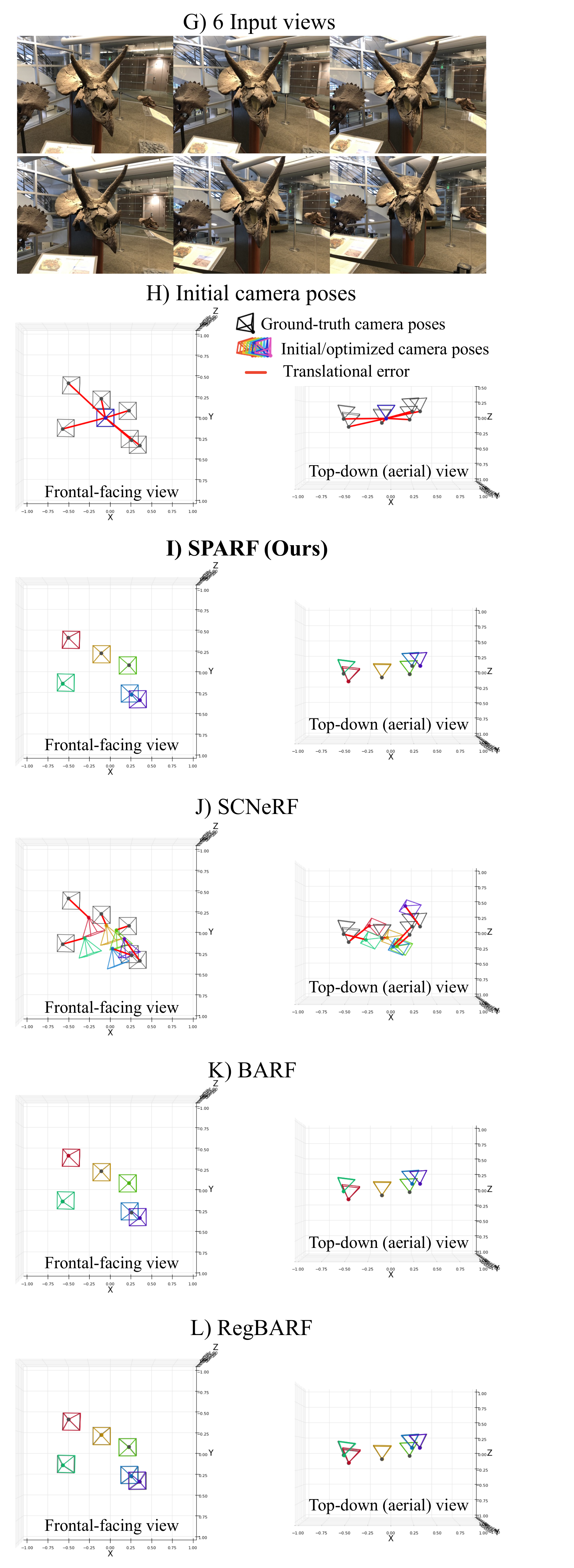}
\caption{Initial and optimized camera poses on the scene 'horns' of the LLFF dataset. We consider 3 or 6 input views with initial identity poses. 
}\vspace{-3mm} 
\label{fig:noise-llff}
\end{figure*}

\begin{figure*}[t]
\centering%
\vspace{-4mm}
\includegraphics[width=0.8\textwidth]{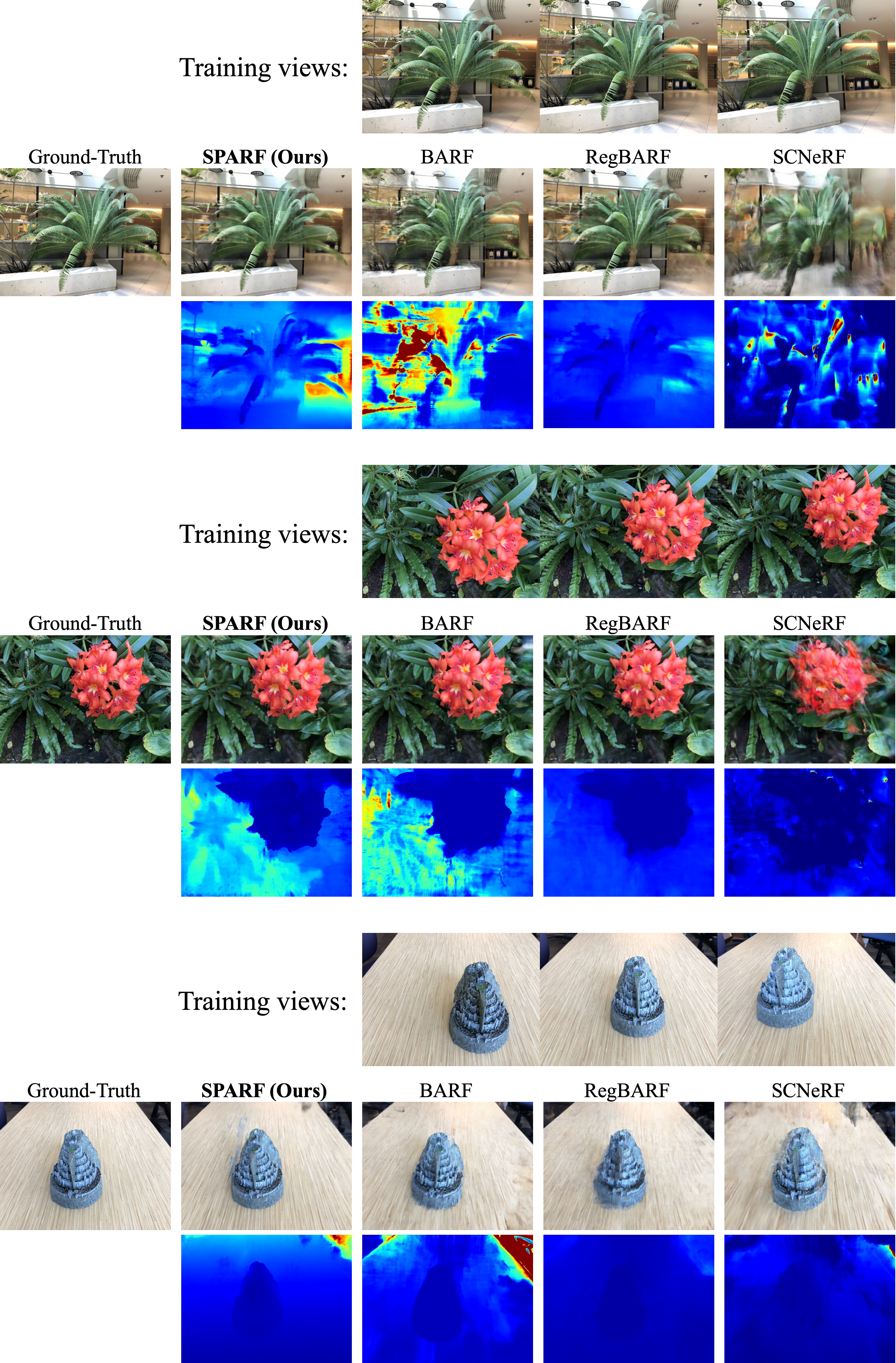}
\vspace{-2mm}
\caption{Novel-view renderings of alternative joint pose-NeRF training approaches on the LLFF dataset. For each scene, we show the RGB (first row) and depth (second row) renderings from an unseen viewpoint.  We consider 3 input views with initial identity poses. 
}
\vspace{-4mm}
\label{fig:qual-llff-compar}
\end{figure*}

\begin{figure*}[t]
\centering%
\vspace{-10mm}
\includegraphics[width=0.7\textwidth]{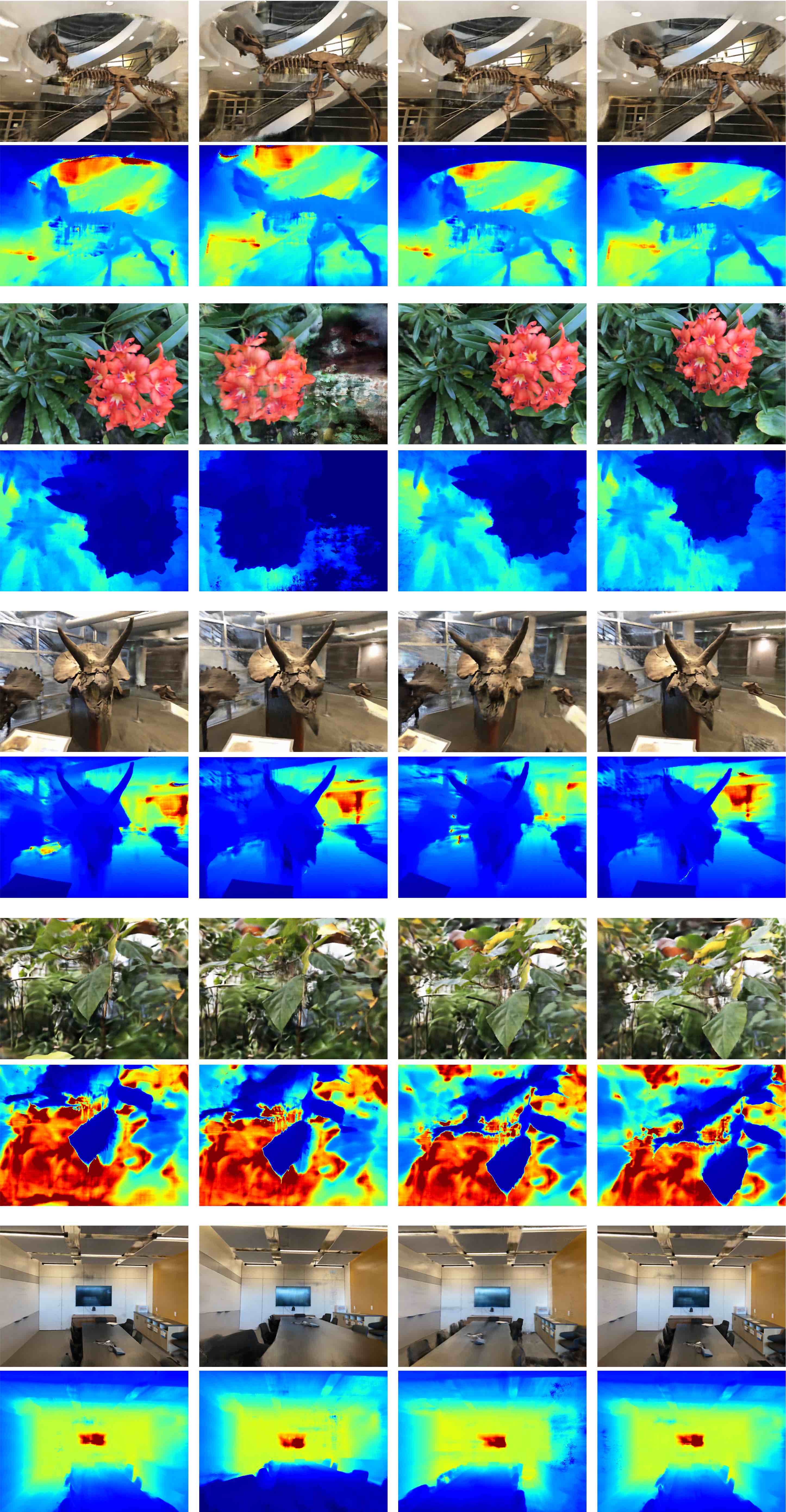}
\vspace{-2mm}
\caption{Novel-view renderings of our \ours on the LLFF dataset.  For each scene, we show the RGB (first row) and depth (second row) renderings from multiple unseen viewpoints. In each scene, we consider 3 input views (not shown here) with initial identity poses. 
}
\vspace{-10mm}
\label{fig:qual-llff}
\end{figure*}

We visualize the initial and optimized poses for one scene of LLFF in the 3 and 6 views scenario in Fig.~\ref{fig:noise-llff}. Here, it is visible that even 6 views can cover most of the scene, which is why BARF performs well even in this sparse-view regime. 
In Fig.~\ref{fig:qual-llff-compar}, we visually compare novel-view renderings of \ours, BARF, RegBARF, and SCNeRF in the 3-view setting. Our approach encodes the scene geometry more accurately. The RGB renderings also contain fewer artifacts and blurriness. 
Finally, we provide examples of the renderings produced by our approach \ours on multiple scenes of LLFF and from different viewpoints in Fig.~\ref{fig:qual-llff}. Given as few as 3 input views with initial identity poses, \ours produces realistic novel-view renderings from many different viewing directions. It also leads to a geometrically accurate scene. 

\parsection{Results with all views} For completeness, in Tab.~\ref{tab:llff-all-views} we compare joint pose-NeRF training approaches BARF and \ours, considering all available training views of LLFF, and starting from identity poses. On this forward-facing dataset, BARF and \ours reach a similar performance in the many-view regime.

\subsection{Results on the Replica dataset}

We here provide additional evaluation results on the Replica dataset, with different pose initialization schemes. We also include more qualitative examples. 

\parsection{Further analysis on Tab.~\ref{tab:replica-colmap-poses} of the main paper}
In Tab.~\ref{tab:replica-colmap-poses}, we evaluated multiple approaches on Replica, with 3 input views and initial poses obtained by COLMAP~\cite{colmap} with PDC-Net~\cite{pdcnet} matches. Those initial poses have an error of 0.39$\degree$ and 3.01 in rotation and translation respectively. 
In the bottom part of the table, we show that \ours can refine the initial poses to a final rotation and translation error of 0.15 and 0.76 respectively. While this might seem like a small improvement in terms of pose registration, the rendering quality improves a lot between \ours with fixed COLMAP poses (F) and \ours with pose refinement (R). This is because the provided initial rotation and translation errors are an \textit{average} over all the scenes. Some scenes actually have an initial translation error of up to 8, which can cause a notable drop in rendering quality. Refining the poses for those scenes is then particularly beneficial in terms of rendering quality. 
This explains the PSNR difference between SPARF in (F) or in (R).

Moreover, some of the baselines show similar rendering quality despite larger pose differences because they struggle to learn a meaningful geometry, \ie they cannot go beyond a certain PSNR. Finally, rendering scores are overall higher on Replica compared to other datasets (even for poor pose registration), because the dataset contains many homogeneous surfaces (\eg. wall). 

\parsection{COLMAP initialization w. SP-SG matches} In the main paper, we compared joint pose-NeRF refinement approaches considering initial poses obtained by COLMAP~\cite{colmap} run with PDC-Net~\cite{pdcnet} matches. For completeness, we here present the same comparison, when the initial poses are obtained with COLMAP with SuperPoint~\cite{superpoint} and SuperGlue~\cite{SarlinDMR20} matches instead. It corresponds to an initial rotation and translation errors of 2.61$\degree$ and 15.31 respectively. The results are presented in Tab.~\ref{tab:suppl-replica-colmap-poses}. 

Compared to initialization with COLMAP-PDCNet (Tab~\ref{tab:replica-colmap-poses} of main paper), the same conclusions apply. Comparing the top (G) and middle part (F) of Tab.~\ref{tab:suppl-replica-colmap-poses}, we show that even a relatively low initial error impacts the novel-view rendering quality when using fixed poses. In the bottom part~(R), our pose-NeRF training strategy \ours leads to the best results, matching the accuracy obtained by our approach with perfect poses (top row, G).

\begin{table}[t]
\centering
\setlength{\tabcolsep}{4pt}
\resizebox{0.47\textwidth}{!}{%
\begin{tabular}{ll|cc|cccc}
\toprule
& & Rot ($\degree$) $\downarrow$ & Trans ($\times$100) $\downarrow$ & PSNR $\uparrow$ & SSIM $\uparrow$ & LPIPS $\downarrow$  & DE $\downarrow$ \\
\toprule
G & \textbf{\ours} &  \multicolumn{2}{c|}{Fixed GT poses}  & 26.43 & \textbf{0.88} & \textbf{0.13} & 0.39 \\
\midrule
F & NeRF &    \multicolumn{2}{c|}{Fixed poses obtained}  & 19.50 & 0.66 & 0.41 & 1.63\\
& DS-NeRF~\cite{DSNerf} &  \multicolumn{2}{c|}{from COLMAP (run w. } & 21.55 & 0.74 & 0.26 & 0.91  \\ 
& \textbf{\ours} (Ours) & \multicolumn{2}{c|}{SP-SG~\cite{SarlinDMR20} matches)}& 22.18 & 0.74 & 0.25 & 0.93 \\

\midrule
R & BARF~\cite{barf} & 3.23 & 18.05 & 19.41 & 0.68 & 0.34 & 0.95  \\
& SCNeRF~\cite{SCNeRF} &  0.21 & 1.17 & 23.67 & 0.82 & 0.22 & 0.83 \\
& DS-NeRF & 1.01 & 3.85 & 24.68 & 0.83 & 0.18 & 0.70  \\ 
& \textbf{\ours} (Ours) & \textbf{0.16} & \textbf{0.8} & \textbf{26.80} & \textbf{0.88} & 0.14 & \textbf{0.36}  \\
\bottomrule
\end{tabular}%
}
\vspace{-2mm}
\caption{Evaluation on Replica~\cite{replica} (3 views) with initial poses obtained by COLMAP~\cite{colmap, pdcnet} with SP-SG~\cite{SarlinDMR20} matches. 
The initial rotation and translation errors are 2.61$\degree$ and 15.31 respectively. In the middle part (F), these initial poses are fixed and used as "pseudo-gt". In the bottom part (R), the poses are refined along with training the NeRF. For comparison, in the top part (G), we use fixed ground-truth poses. 
}
\label{tab:suppl-replica-colmap-poses}
\end{table}

\begin{table}[b]
\centering
\setlength{\tabcolsep}{4pt}
\resizebox{0.47\textwidth}{!}{%
\begin{tabular}{l|cc|cccc}
\toprule
&  \multicolumn{2}{c}{\textbf{Pose Registration}} & \multicolumn{4}{c}{\textbf{Novel View Synthetis}} \\
 & Rot ($\degree$) $\downarrow$ & Trans ($\times$100) $\downarrow$ & PSNR $\uparrow$ & SSIM $\uparrow$ & LPIPS $\downarrow$  & DE $\downarrow$ \\
 \toprule
BARF~\cite{barf} & 12.81 & 39.96 & 16.39 & 0.60 & 0.52 & 2.3 \\
RegBARF~\cite{Regnerf, barf} & 9.0 & 29.34 & 17.05 & 0.62 & 0.48 & 1.11 \\
DistBARF ~\cite{Regnerf, mipnerf360} & 5.28 & 20.45 & 19.82 & 0.69 & 0.36 & 0.68  \\
SCNeRF~\cite{SCNeRF} &  2.26 & 10.37 & 22.50 & 0.76 & 0.27 & 1.57 \\
\textbf{\ours} & \textbf{1.06} & \textbf{6.63} & \textbf{25.57} & \textbf{0.85} & \textbf{0.16} & \textbf{0.45}  \\ 
\bottomrule
\end{tabular}%
}\vspace{-2mm}
\caption{Evaluation on the Replica dataset (3 views) starting from noisy poses. In particular, the ground-truth poses are synthetically perturbed with 15\% of additive Gaussian noise. This initialization leads to an initial rotation and translation errors of 15.62$\degree$ and 112 (multiplied by 100) respectively. }
\vspace{-3mm}\label{tab:suppl-replica-pose}
\end{table}

\parsection{Initial noisy poses} For completeness, we also start from synthetically perturbed ground-truth poses. In particular, as previously for DTU, we synthetically perturb the ground-truth poses with $15\%$ of additive Gaussian noise. It leads to an initial rotation and translation errors of 15.62$\degree$ and 112 (scaled by 100) respectively.  This corresponds to a significantly noisier setting than starting from COLMAP poses.  Results are presented in Tab.~\ref{tab:suppl-replica-pose}. 
BARF struggles to refine the poses. RegBARF and DistBARF lead to better pose registration and novel-view synthesis. Here, it is interesting to note that both regularizations seem to help in learning a more accurate geometry (lower depth error). Indeed, SCNeRF, which better registers the poses, still obtains a higher depth error. Our approach \ours, which acts on \emph{both} the learned scene geometry and the camera poses, significantly outperforms all others.

\begin{figure*}[t]
\centering%
\vspace{-15mm}
\includegraphics[width=0.99\textwidth]{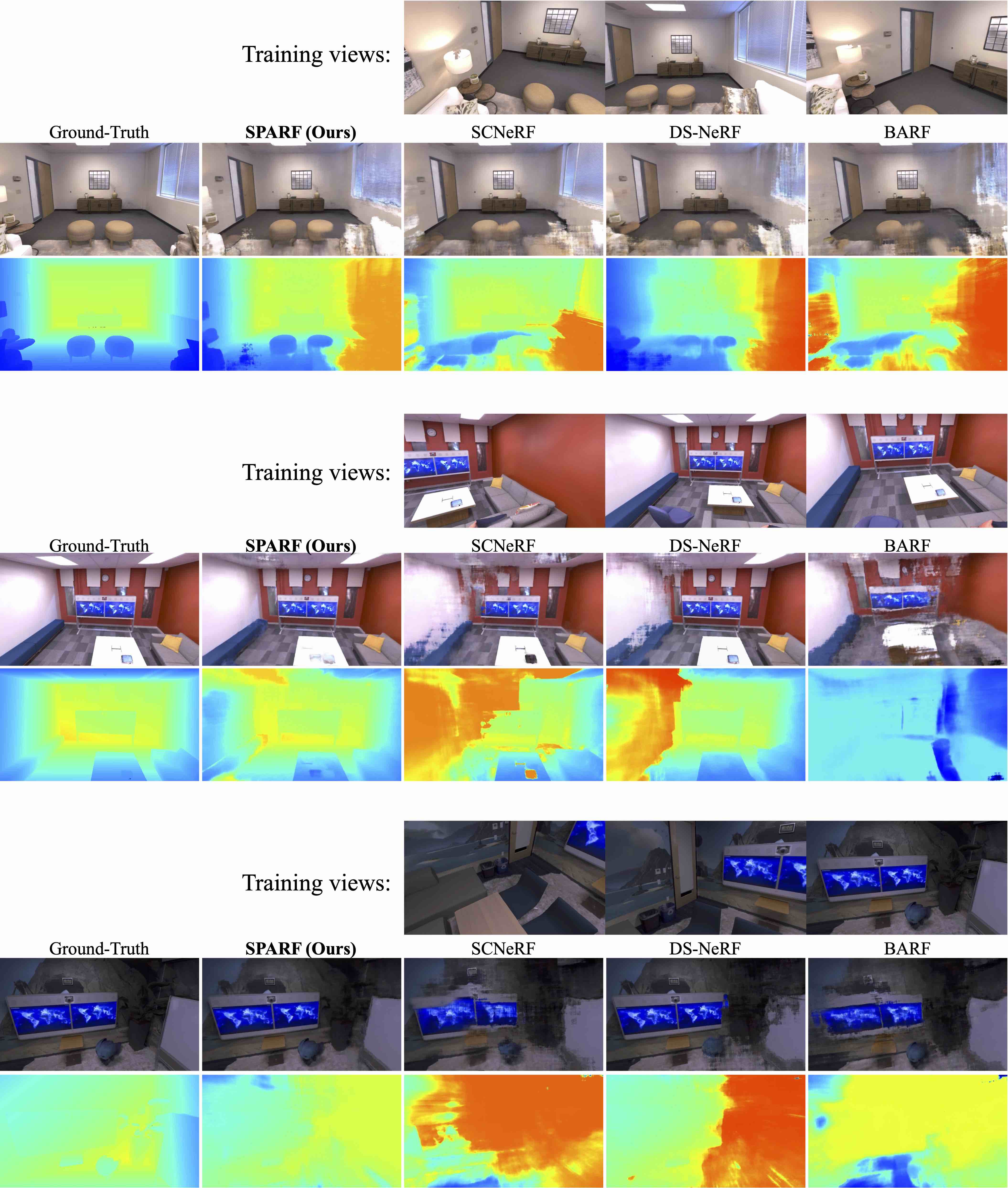}
\vspace{-2mm}
\caption{Novel-view renderings of alternative joint pose-NeRF training approaches on the Replica dataset. For each scene, we show the RGB (first row) and depth (second row) renderings from an unseen viewpoint.  On each scene, we consider 3 input views (not shown here) with initial poses obtained by COLMAP~\cite{colmap} with PDC-Net matches~\cite{pdcnet}. 
}

\vspace{-10mm}
\label{fig:qual-replica-comp}
\end{figure*}

\begin{figure*}[t]
\centering%
\vspace{-15mm}
\includegraphics[width=0.99\textwidth]{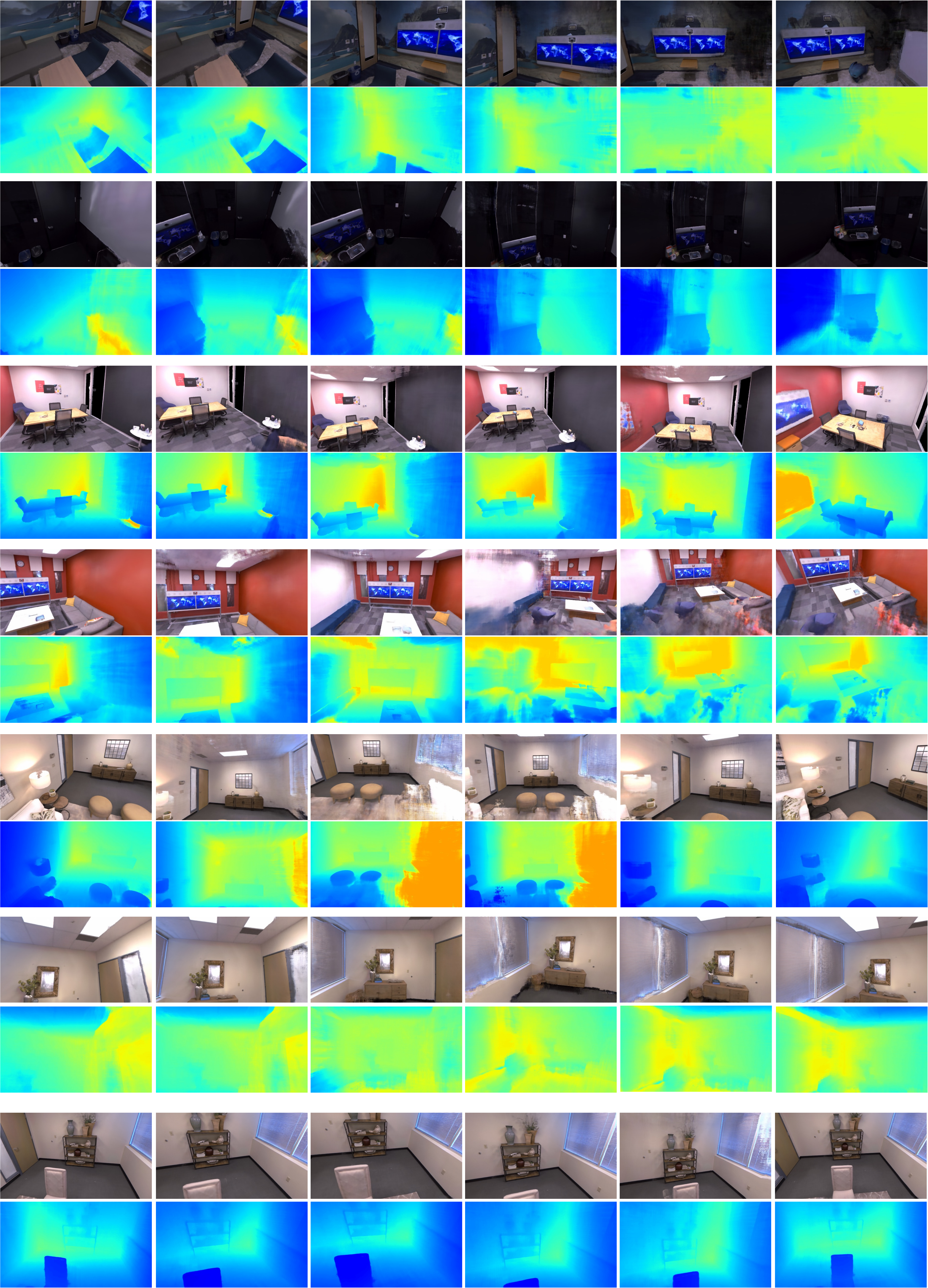}
\vspace{-2mm}
\caption{Novel-view renderings of our \ours on the Replica dataset.  For each scene, we show the RGB (first row) and depth (second row) renderings from multiple unseen viewpoints. On each scene, we consider 3 input views (not shown here) with initial poses obtained by COLMAP~\cite{colmap} with PDC-Net matches~\cite{pdcnet}. 
}
\vspace{-10mm}
\label{fig:qual-replica}
\end{figure*}

\parsection{Qualitative comparisons} In Fig.~\ref{fig:qual-replica-comp}, we qualitatively compare \ours with BARF, DS-NeRF and SCNeRF. Our approach \ours produces the best renderings, with significantly fewer floaters and blurry surfaces. The learned scene geometry is also significantly sharper and more accurate, as shown by the depth renderings. This is confirmed in Fig.~\ref{fig:qual-replica}, where we present additional renderings produced by \ours on all scenes of the Replica dataset. Note that in all those cases, our approach is only trained with 3 input views, and noisy input camera poses (obtained by COLMAP-PDCNet).

\begin{table*}[t]
\centering

\setlength{\tabcolsep}{4pt}
\resizebox{0.99\textwidth}{!}{%
\begin{tabular}{@{~}l@{~}|ccc|ccc|ccc|@{~}c@{~}c@{~}c@{~}}
\toprule
  & \multicolumn{3}{c}{\textbf{PSNR} $\uparrow
$} & \multicolumn{3}{c}{\textbf{SSIM} $\uparrow
$ } & \multicolumn{3}{c}{\textbf{LPIPS} $\downarrow
$} & \multicolumn{3}{c}{\textbf{DE} $\downarrow
$}\\ 
 & 3 & 6 & 9 & 3 & 6 & 9 & 3 &  6 & 9 & 3 &  6 & 9\\
 \toprule
 PixelNeRF~\cite{pixelnerf} & \best{19.36} (18.00) & 20.46 (19.12) & 20.91 (19.56) &  \second{0.70} (\second{0.77}) & 0.75 (0.80) & 0.76 (0.81) & \second{0.32} (0.23) & 0.30 (0.22) & 0.29 (0.21) & \second{0.12} & \second{0.12} & 0.13 \\
\midrule  
NeRF~\cite{Nerf} & 8.41 (9.34) &  17.51 (18.52) & 21.45 (23.25) &        0.31 (0.63) &  0.73 (0.83) & \second{0.85} (\second{0.91}) &  0.71 (0.36) & 0.25 (0.13) & \second{0.14} (\second{0.06}) &   0.87  & 0.21 & \second{0.08}    \\
DietNeRF~\cite{dietnerf} & 10.01 (11.85) & 18.70 (20.63) & 22.16 (23.83) &  0.35  (0.63) & 0.67 (0.78) & 0.68 (0.82) & 0.57 (0.31) & 0.35 (0.20) & 0.34 (0.17) &-& - & - \\
RegNeRF~\cite{Regnerf} & 15.33 (\second{18.89}) &  19.10 (\second{22.20}) & \second{22.30} (\second{24.93}) & 0.62 (0.75) & \second{0.76} (\second{0.84}) & 0.82 (0.88) &  0.34 (\second{0.19}) & \second{0.23} (\second{0.12}) & 0.18 (0.09)  & - & - & -   \\
DS-NeRF~\cite{DSNerf}& 16.52 (-) & \second{20.54} (-) & 22.23 (-) &  0.54 (-)  & 0.73 (-) & 0.77 (-) & 0.48 (-) & 0.31 (-) & 0.26 (-) & - & - & -\\

\textbf{\ours} (Ours) & \second{18.30} (\best{21.01}) & \best{23.24} (\best{25.76}) & \best{25.75} (\best{27.30}) & \best{0.78} (\best{0.87}) & \best{0.87} (\best{0.92}) & \best{0.91} (\best{0.94}) & \best{0.21} (\best{0.10}) & \best{0.12} (\best{0.06}) & \best{0.08} (\best{0.04})  & \best{0.083} & \best{0.049} & \best{0.043}\\  
\bottomrule
\end{tabular}%
}\vspace{-1mm}
\caption{Evaluation on the DTU dataset~\cite{dtu}, considering fixed ground-truth poses. We present novel-view synthesis results for different numbers of input views. Results in ($\cdot$) are computed by masking the background. Results of~\cite{dietnerf, mipnerf, Regnerf, mvsnerf, pixelnerf} are from~\cite{Regnerf}. The best and second-best results are in red and blue respectively.}
\vspace{-3mm}\label{tab:suppl-dtu-pixelnerf}
\end{table*}

\begin{table*}[t]
\centering
\setlength{\tabcolsep}{4pt}
\resizebox{0.60\textwidth}{!}{%
\begin{tabular}{l|ccc|ccc|ccc}
\toprule
  & \multicolumn{3}{c}{\textbf{PSNR} $\uparrow
$ } & \multicolumn{3}{c}{\textbf{SSIM} $\uparrow
$ } & \multicolumn{3}{c}{\textbf{LPIPS} $\downarrow
$} \\ 
 & 3 & 6 & 9 & 3 & 6 & 9 & 3 & 6 & 9 \\
\toprule
PixelNeRF~\cite{pixelnerf}& 7.93 & 8.74 & 8.61 & 0.27 & 0.28 & 0.27 & 0.68 & 0.68 & 0.67 \\
SRF~\cite{Chibane2021StereoRF} & 12.3&13.1 & 13.0 & 0.25 & 0.29 & 0.30 & 0.59 &0.59 & 0.61 \\
MVSNeRF~\cite{mvsnerf}& 17.25 & 19.79 & 20.47 & 0.56 & 0.66 & 0.69 & 0.36 & 0.27 & 0.24 \\
\midrule
PixelNeRF-ft  & 16.17 & 17.03 & 18.92 &  0.44 & 0.47 & 0.54 & 0.51 & 0.48 & 0.43 \\
SRF-ft & 17.07 &16.75 & 17.39 & 0.44 &  0.44 & 0.47 & 0.53 &0.52 & 0.50 \\
MVSNeRF-ft & 17.88 & 19.99 & 20.47 & 0.58 & 0.66 & 0.70 & 0.33 & 0.26 & 0.24  \\
\midrule
NeRF~\cite{Nerf} & 13.61 & 16.70 & 18.45  &  0.28 & 0.43 & 0.51 & 0.56& 0.40 &  0.31\\
MipNeRF~\cite{mipnerf}  & 14.62 & 20.87 & 24.26 & 0.35 &0.69 & \second{0.81} & 0.50 & 0.26 & \second{0.17} \\
DietNeRF*~\cite{dietnerf} & 14.94 & 21.75 & 24.3 &  0.37 &0.72 & 0.80 & 0.5 &0.25 & 0.18 \\
RegNeRF*~\cite{Regnerf} & \second{19.08} & \second{23.10} & \best{24.86} & \second{0.59} &  \best{0.76} & \best{0.82} &  0.34 &  \second{0.21} & \best{0.16}\\
DS-NeRF~\cite{DSNerf}&  18.00 & 21.60 & 22.84 & 0.55 & 0.67& 0.71 & \second{0.27} & \second{0.21} & 0.19   \\
\midrule

\textbf{\ours} (Ours) &  \best{20.20} & \best{23.35}& \second{24.40} & \best{0.63} & \second{0.74} & 0.77 & \best{0.24} & \best{0.20} & 0.18 \\ 
\bottomrule
\end{tabular}%
}\vspace{-1mm}
\caption{Evaluation on the LLFF dataset~\cite{llff}, considering fixed ground-truth poses. We present novel-view synthesis results for different numbers of input views. The top part contains conditional models trained on DTU. In the middle part, we present the same conditional models, further finetuned per scene on LLFF. Finally, in the last part, we compare per-scene NeRF-based approaches.  Approaches with $*$ use the MipNeRF~\cite{mipnerf} as their base architecture, while the others use NeRF~\cite{Nerf}. Results of~\cite{dietnerf, mipnerf, Regnerf, mvsnerf, pixelnerf} are from~\cite{Regnerf}. The best and second-best results are in red and blue respectively.
}
\label{tab:suppl-llff}
\end{table*}

\section{Additional Results with Fixed GT Poses}
\label{sec-sup:results-fixed}

In Sec.~\ref{subsec:sota-gt-poses}, we evaluated our approach when considering fixed ground-truth poses, in the three-input-views setting. For completeness, we extend this evaluation for the cases of 6 and 9 input views. This is the same setup as in~\cite{Regnerf}. 

\parsection{Results on DTU} We present results on DTU in Tab.~\ref{tab:suppl-dtu-pixelnerf}. Our approach \ours sets a new state of the art on all metrics for 3, 6, or 9 input views. The only exception is PSNR on the whole image when only 3 input views are available, which we already mentioned in the main paper.

\parsection{Results on LLFF} We present results on LLFF in Tab.~\ref{tab:suppl-llff}. The conditional models PixelNeRF, SRF, and MVSNeRF are trained on the DTU dataset. LLFF thus serves as an out-of-distribution scenario. It appears that SRF and PixelNeRF tend to overfit to the training data, leading to poor quantitative results. MVSNeRF generalizes better to novel data. All three conditional models seem to benefit from additional fine-tuning. For 3 input views, NeRF, MipNeRF, and DietNeRF perform worse than conditional models. DS-NeRF, RegNeRF, and our approach \ours nevertheless outperform the best conditional model, \ie MVSNeRF.  In the 6 and 9 view settings, all per-scene approaches except for the standard NeRF outperform MVSNeRF. 

Our approach \ours outperforms all others on all metrics in the sparsest scenario, \ie when considering 3 input views. For 6 and 9 views, it obtains a slightly lower performance than MipNeRF and RegNeRF, the latter using MipNeRF as the base architecture. Nevertheless, our \ours, which is based on the NeRF architecture, obtains drastically better results than the standard NeRF or DS-NeRF. Our approach could in theory be applied to any base network, for example, MipNeRF. As a result, we believe combining our approach with the MipNeRF base architecture could lead to even better rendering quality.

